\documentclass[runningheads]{llncs}

\usepackage[mobile]{eccv}

\usepackage{eccvabbrv}
\usepackage{pifont}  %
\usepackage{graphicx}
\usepackage{booktabs}
\usepackage{comment}

\usepackage[accsupp]{axessibility}  %
\usepackage{amssymb}
\usepackage{pifont}
\usepackage{booktabs}
\usepackage{tabularx}
\usepackage{array}
\usepackage{graphicx} %
\usepackage[export]{adjustbox} %
\usepackage{multirow}

\usepackage{hyperref}

\usepackage{orcidlink}
\usepackage{graphicx}
\usepackage{subcaption}

\begin{document}

\title{EDGE-Shield: Efficient Denoising-staGE Shield for Violative Content Filtering via Scalable Reference-Based Similarity Matching} 

\titlerunning{Abbreviated paper title}

\author{Takara Taniguchi\thanks{Work done during the internship at SB intuitions} \and
Ryohei Shimizu \and
Duc Minh Vo \and
Kota Izumi \and
Shiqi Yang \and
Teppei Suzuki
}
\authorrunning{F.~Author et al.}

\institute{SB intuitions \\
\email{hiroshi-tani@g.ecc.u-tokyo.ac.jp, \{ryohei.shimizu, minh.duc.vo, kota.izumi, shiqi.yang, teppei.suzuki\}@sbintuitions.co.jp}}

\maketitle

\begin{abstract}
The advent of Text-to-Image generative models poses significant risks of copyright violation and deepfake generation.
Since the rapid proliferation of new copyrighted works and private individuals constantly emerges, reference-based training-free content filters are essential for providing up-to-date protection without the constraints of a fixed knowledge cutoff.
However, existing reference-based approaches often lack scalability when handling numerous references and require waiting for finishing image generation.
To solve these problems, we propose EDGE-Shield, a scalable content filter during the denoising process that maintains practical latency while effectively blocking violative content. 
We leverage embedding-based matching for efficient reference comparison. 
Additionally, we introduce an \textit{$x$}-pred transformation that converts the model's noisy intermediate latent into the pseudo-estimated clean latent at the later stage, enhancing classification accuracy of violative content at earlier denoising stages. 
We conduct experiments of violative content filtering against two generative models including Z-Image-Turbo and Qwen-Image.
EDGE-Shield significantly outperforms traditional reference-based methods in terms of latency; it achieves an approximate $79\%$ reduction in processing time for Z-Image-Turbo and approximate $50\%$ reduction for Qwen-Image, maintaining the filtering accuracy across different model architectures.
\end{abstract}

\section{Introduction}

\begin{figure}
    \centering
    \includegraphics[width=1.0\linewidth]{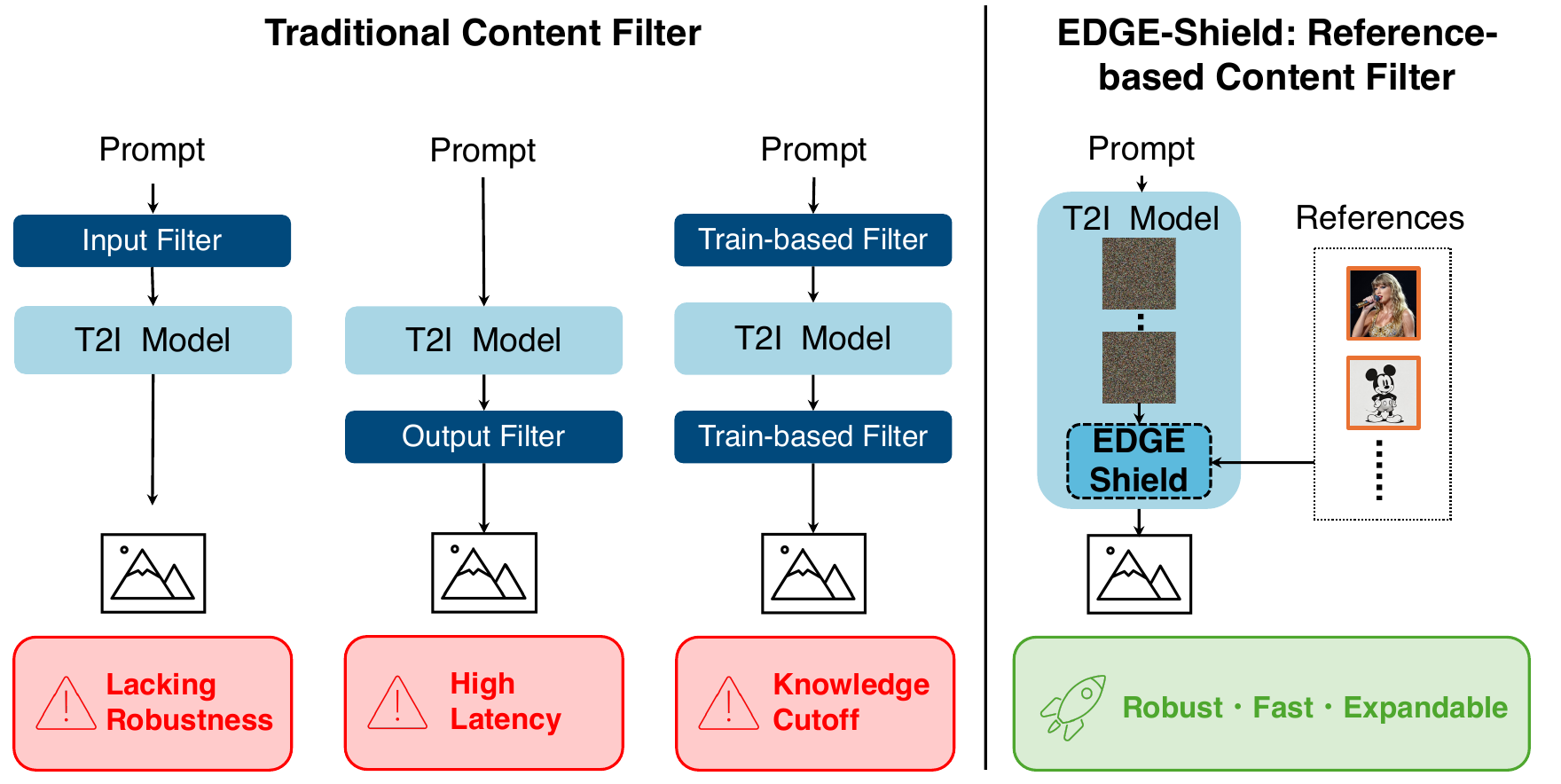}
    \caption{Comparison of our EDGE-Shield with the existing content filters.}
    \label{fig:introduction}
\end{figure}

\begin{table}[t]
\setlength{\tabcolsep}{8pt}
\centering
\caption{Comparison of content filtering paradigms. Output-based methods are crucial for prompt-agnostic filtering, while reference-based approaches are essential for protecting constantly emerging IPs and private individuals. EDGE-Shield bridges the critical gap by integrating these advantages while overcoming the scalability and latency limitations of existing methods.}\label{tbl:method_comparison}
\begin{tabular}{lccccc}
\toprule
 & Train + Inp & Train + Out & Ref + Inp & Ref + Out & \textbf{Ours} \\ 
\midrule
Latency     & \checkmark  & $\times$    & \checkmark  & $\times$    & \checkmark \\ 
Scalability & \checkmark  & \checkmark  & \checkmark  & $\times$    & \checkmark \\ 
Accuracy    & \checkmark & \checkmark  & \checkmark & \checkmark  & \checkmark \\ 
Knowledge   & $\times$    & $\times$    & \checkmark  & \checkmark  & \checkmark \\ 
Robustness  & $\times$    & \checkmark  & $\times$    & \checkmark  & \checkmark \\ 
\bottomrule
\end{tabular}
\end{table}

The rise of Text-to-Image (T2I) models~\cite{wu2025qwenimagetechnicalreport,labs2025flux1kontextflowmatching,imageteam2025zimageefficientimagegeneration,Rombach_2022_CVPR_stablediffusion,ramesh2022hierarchicaltextconditionalimagegenerationDall-E,podell2024sdxl} has brought issues of violative content including copyright infringement and deepfake to the forefront. 
A challenge of violative content generation arises from the fact that training datasets for these models can inadvertently include violative images~\cite{L2norm2023Carlini}, enabling the models to generate these images.

Approaches to prevent violative content generation can be broadly categorized into two primary paradigms: concept removal and content filters.
While concept removal~\cite{yoon2024safree,cywinski2025saeuron} steers the T2I model to generate alternatives by modifying the model's internal representations, whereas content filters~\cite{li2025t2isafety_imageguard,helff2025llavaguard,chi2024llamaguard3vision} block generation when the generated content is unsafe.

Content filters offer an advantage of providing non-invasive protection without altering the output of the original model itself.
As illustrated in Table~\ref{tbl:method_comparison}, content filter can be categorized along two independent axes: input-based vs. output-based, and training-based vs. reference-based.
Input-based content filters~\cite{inan2023llamaguardllmbasedinputoutput,liu2024latentguard} check prompts before generation, while output-based content filters~\cite{helff2025llavaguard,li2025t2isafety_imageguard} look into the output imagery.
While training-based content filters require prior training, reference-based ones identify violations by making use of reference information in a training-free manner.

Output-based content filters serve as critical defense mechanisms, successfully catching violative content regardless of how the prompt is phrased.
Unlike input-based methods that rely on prompt or text embedding analysis, these approaches operate on the generated content~\cite{song2024diffsim,li2025t2isafety_imageguard} or content during denoising process~\cite{yang2025seeingIGD}, making them robust against the prompt styles.

Thanks to reference information, reference-based content filters \cite{L2norm2023Carlini,song2024diffsim,klemen2023ffa} can overcome the knowledge cutoff issues of training-based content filters, which struggle to keep pace with a constant influx of new copyrighted works and individuals.
For example, while Vision-Language Models (VLM) used as evaluators~\cite{wang2025internvl35advancingopensourcemultimodal,bai2025qwen3vltechnicalreport,liu2023llava,openai2024gpt4ocard,liu2024llavanext} demonstrate robust filtering capabilities, their efficacy is inherently constrained by their internal knowledge base.
In contrast, reference-based filters can circumvent these limitations, provided that relevant reference data is available.

However, current content filters face separate practical challenges regarding detection latency for output-based ones and inference-time scalability for reference-based ones.
Traditional output-based content filters typically wait for the end of the image generation process, leading to high latency and wasting computational budget for generating unsafe outputs. 
While some methods attempt to mitigate this by operating during the denoising process~\cite{yang2025seeingIGD,liu2026wukongframeworksafework}, they are currently limited to training-based frameworks and lack applicability outside their trained domain. 
Moreover, existing reference-based content filters~\cite{helff2025llavaguard,song2024diffsim} struggle to scale when protecting against multiple reference images, as they often require independent evaluations for each reference target.

To address these challenges, we propose a scalable reference-based content filter called EDGE-Shield which scales efficiently with the number of reference images while preserving high filtering accuracy.
Our method outputs the violation score of the image in generation against reference images based on the similarity between embeddings of reference and generated images, which is used for binary classification with the threshold whether to accept or reject the image in generation.
The embeddings of the reference images can be pre-computed, ensuring high scalability with minimal impact on inference latency, even as the number of reference images increases.
To enable efficient filtering for prevalent ODE-based image generation models, our proposed method introduces a filtering mechanism directly into the denoising process.
By applying the simple yet effective $x$-pred transformation, the proposed method achieves filtering accuracy comparable to output-based approaches, while maintaining inference speeds on par with input-based methods.

In summary, we propose EDGE-Shield to address the limitations of existing methods summarized in Table~\ref{tbl:method_comparison}.
Our main contributions are as follows: 
(\textit{\textbf{i}}) we leverage violation scoring within the denoising process through pre-computed embeddings and the $x$-pred transformation, ensuring scalability relative to the number of reference images and facilitating accurate filtering at an early stage of the denoising process; 
and (\textit{\textbf{ii}}) we demonstrate that EDGE-Shield achieves an ROC-AUC of approximately $0.85$ for both Z-Image-Turbo and Qwen-Image, which is comparable to existing baselines while reducing processing latency by $79\%$ and $50\%$, respectively.

\section{Related Work}

\subsection{Concept Removal}

Concept removal~\cite{lu2024MACE,Gandikota_2024_WACV_UCE,Schramowski_2023_SLD,yoon2024safree,biswas2025cure,gao2024eraseanything} steers the T2I model to generate alternatives to specified concepts by modifying the model's internal representations.
Some approaches neutralize specified concepts at the prompt level, modifying text embeddings or redirecting tokens to bypass the activation of sensitive semantic clusters~\cite{yoon2024safree,cywinski2025saeuron}.
Other approaches intervene directly during the iterative reverse diffusion process, employing guidance or latent steering to suppress the emergence of specific visual features in real-time~\cite{kim2025conceptsteerersleveragingksparse,cywinski2025saeuron,tatiana2026casteer,kim2025trainingfree}.

\subsection{Content Filter}

Whereas various content filters are actively studied, output-based approaches are advantageous due to their prompt-agnostic nature, allowing them to maintain high performance without being biased by the nuances of the input text.
While input-based content filters~\cite{yang2024guardti,liu2024latentguard,inan2023llamaguardllmbasedinputoutput,OpenAILLMInputfilterMarkov_Zhang_Agarwal_EloundouNekoul_Lee_Adler_Jiang_Weng_2023} block or rewrite problematic prompts before processing, output-based content filters~\cite{helff2025llavaguard,li2025t2isafety_imageguard}, recently leveraging VLMs~\cite{liu2023llava,bai2025qwen3vltechnicalreport,wang2025internvl35advancingopensourcemultimodal,li2025t2isafety_imageguard,rando2022redteaming,liu2025copyjudge,li2025t2isafety_imageguard}, use the final generated imagery to prevent the emergence of prohibited concepts.
Recently, some of output-based content filters for NSFW monitor the intermediate state of the output in the denoising process to obtain the low latency~\cite{yang2025seeingIGD}.

Reference-based content filters using reference images~\cite{L2norm2023Carlini,pizzi2022self,song2024diffsim,NEURIPS2024_ICDwang,Shi2025RLCP} offer a promising alternative to training-based methods~\cite{liu2024latentguard} by bypassing inherent knowledge cutoffs to enable protection for newly emerging copyrighted content and individuals not yet represented in the model’s training data.

Waiting for the entire generation process to complete before evaluating and low scalability for multi-reference settings hurts latency and scalability against multiple references of output-based and reference-based content filters shown in Table~\ref{tbl:method_comparison}, respectively.
Our work bridges this gap by proposing reference-based content filter during the denoising process, maintaining low latency and scalability against the number of references.

\section{Preliminaries}
The purpose of this section is to explain the property of the intermediate latent in ODE-based generative models that clean representations can be derived from the noisy intermediate states~\cite{li2026basicsletdenoisinggenerative}.
We first establish the concept of interpolation, which defines the noisy intermediate states connecting pure noise and clean data. 
Next, we formulate the vector field called flow velocity and the relationship with the interpolation, the training target in recent flow-based models. 
Finally, we demonstrate how these formulations can be combined to predict a pseudo-clean data sample directly from any given time step.

\noindent\textbf{Interpolation.}
We first formalize the noisy intermediate states used in ODE-based generative modeling.
Let $\boldsymbol{x} \sim p_{\text{data}}(\boldsymbol{x})$ be a sample from the data distribution and $\boldsymbol{\epsilon} \sim p_{\text{noise}}(\boldsymbol{\epsilon})$ (\textit{e.g.}, $\mathcal{N}(0, \mathbf{I})$) be a sample from a tractable noise distribution.
In these models, a noisy intermediate sample $\boldsymbol{z}_t$ is constructed via an interpolation between the data and noise:
\begin{equation}\label{eq:interpolation}
    \boldsymbol{z}_t = \alpha_t \boldsymbol{x} + \sigma_t \boldsymbol{\epsilon},
\end{equation}
where $\alpha_t$ and $\sigma_t$ are time-dependent coefficients determining the noise schedule. In this work, we define $t \in [0, 1]$ as the time variable representing a path from pure noise at $t=0$ to a clean data sample at $t=1$.

\noindent\textbf{Flow-based Models.}
Specifically, we examine flow-matching models, a dominant paradigm in recent T2I architectures.
These models typically utilize a linear noise schedule where $\alpha_t = t$ and $\sigma_t = 1-t$, resulting in the intermediate state $\boldsymbol{z}_t = t\boldsymbol{x} + (1-t)\boldsymbol{\epsilon}$. 
The model $\boldsymbol{v}_\theta(\boldsymbol{z}_t, t)$ parameterized by $\theta$ is trained to predict the flow velocity $\boldsymbol{v}_t$, defined as the time derivative of $\boldsymbol{z}_t$:
\begin{equation}\label{eq:flow_velocity}
    \boldsymbol{v}_t = \frac{{\rm d}\boldsymbol{z}_t}{{\rm d}t} = \boldsymbol{x} - \boldsymbol{\epsilon}.
\end{equation}
Flow-based models optimize the flow-matching loss $\mathcal{L} = \mathbb{E}_{t, \boldsymbol{x}, \boldsymbol{\epsilon}} \| \boldsymbol{v}_\theta(\boldsymbol{z}_t, t) - \boldsymbol{v}_t \|^2$. 

\noindent\textbf{Inference.}
During inference, a data sample is generated by numerically solving an ordinary differential equation (ODE) $\frac{{\rm d}\boldsymbol{z}_t}{{\rm d}t} = \boldsymbol{v}_\theta(\boldsymbol{z}_t, t)$ starting from $\boldsymbol{z}_0 = \boldsymbol{\epsilon}$. This process is inherently iterative, requiring multiple sequential forward passes through the model $\boldsymbol{v}_\theta$ to compute the trajectory from $t=0$ to $t=1$. This iterative nature is the primary source of inference latency in flow-based models.

\noindent\textbf{Pseudo-Clean Sample Estimation.}
Despite the iterative requirement for generation, we can estimate a pseudo-clean data sample directly from any intermediate state $\boldsymbol{z}_t$. 
By combining Eq.~\ref{eq:interpolation} and Eq.~\ref{eq:flow_velocity} under linear scheduling, the estimate $\boldsymbol{x}_\theta(\boldsymbol{z}_t, t)$ can be derived using the predicted velocity:
\begin{equation}\label{eq:pseudox}
    \boldsymbol{x}_\theta(\boldsymbol{z}_t, t) = \boldsymbol{z}_t + (1 - t) \boldsymbol{v}_\theta(\boldsymbol{z}_t, t).
\end{equation}

As $t \to 1$, this approximation converges to the actual generated sample. While we focus on the pseudo-clean sample estimation in recently used flow-based models~\cite{wu2025qwenimagetechnicalreport,imageteam2025zimageefficientimagegeneration,labs2025flux1kontextflowmatching}, we can apply the estimation to generative models using noise-based models such as SD1.4~\cite{Rombach_2022_CVPR_stablediffusion} as well, detailed in the Appendix.

\section{Methodology}

We propose \textbf{EDGE-Shield}, a content filter during the denoising process designed to overcome the scalability and latency. 
To correspond with large reference sets and obtain low latency, EDGE-Shield classify the intermediate state of the generation process with cached embeddings of references.

\subsection{Task Definition}
We formulate the safety assessment as a binary classification problem.
Consider a T2I model $G$, an input text prompt $c$, and a set of reference images $\mathcal{R} = \{\boldsymbol{r}_1, \boldsymbol{r}_2, \ldots, \boldsymbol{r}_n\}$, where each $\boldsymbol{r}_i \in \mathcal{I}$ exemplifies a specific, mutually exclusive category of violative content with $\mathcal{I}$ denoting the overall image space.
Our objective is to determine a binary label $y \in \{0, 1\}$ for the triplet $(G, c, \mathcal{R})$, where $y=1$ signifies that the generated image $G(c)$ constitutes a violation relative to $\mathcal{R}$, and $y=0$ indicates compliance.

\subsection{Method Overview}

\begin{figure}[t]
    \centering
    \includegraphics[width=1\linewidth]{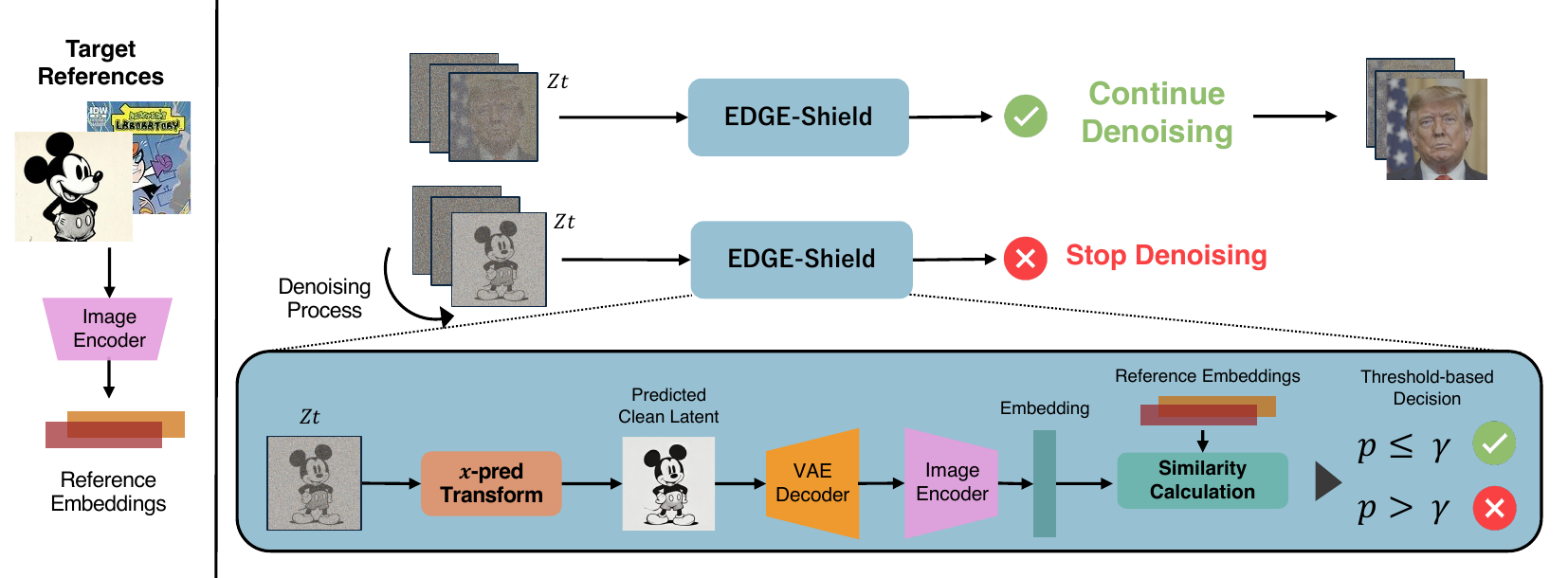}
    \caption{Overview of embedding caching and the detail component of EDGE-Shield. Left: We compute the embeddings of target references of violative content. 
    Right: we transform the intermediate latent into the clean latent by using our $x$-pred transformation. 
    EDGE-Shield calculates the similarity score among the embedding of the decoded clean latent and cached embeddings.}
    \label{fig:method_detail}
\end{figure}

Figure~\ref{fig:method_detail} illustrates the detail workflow of EDGE-Shield, which consists of three steps to achieve the scalable and efficient classification of violative content.
First, we pre-compute embeddings for the set of reference images and store them in a cache, as detailed in Sec.~\ref{subsec:queryembed}, which offers efficient filtering.
Second, we apply $x$-pred transformation to estimate the latent at the final step, which offers accurate filtering in an early denoising step.
The detailed procedure is described in Sec.~\ref{subsec:x-pred_module}.
Finally, we compute the similarity score between the embedding of the decoded $x$-pred transformed latent and the cached reference embeddings to determine compliance of content in generation in Sec.~\ref{subsec:simulatiry_calculation}.

\subsection{Reference Embedding}\label{subsec:queryembed}

To ensure that EDGE-Shield scales effectively with the number of reference images, we pre-compute and store their embeddings as a cached matrix. Since re-encoding a large reference set for every generation step would be computationally prohibitive, this pre-computation is performed once and reused across multiple runs.
An image encoder $E: \mathcal{I} \longmapsto  \mathbb{R}^d$ transforms the set of reference images $\mathcal{R}$ into their corresponding embeddings $\{E(\boldsymbol{r}_i)\}_{i=1}^n$, where each $E(\boldsymbol{r}_i) \in \mathbb{R}^d$ is a $d$-dimensional feature vector. 
For efficient computation, these embeddings are $\ell_2$-normalized and concatenated into a reference embedding matrix:
\begin{equation}
    \mathbf{R} = \left[ \frac{E(\boldsymbol{r}_1)}{\|E(\boldsymbol{r}_1)\|}, \frac{E(\boldsymbol{r}_2)}{\|E(\boldsymbol{r}_2)\|}, \dots, \frac{E(\boldsymbol{r}_n)}{\|E(\boldsymbol{r}_n)\|} \right]^\top \in \mathbb{R}^{n \times d}.
\end{equation}
This matrix $\mathbf{R}$ is stored in memory and subsequently used for similarity scoring in Sec.~\ref{subsec:simulatiry_calculation} to determine whether the generated content violates any references during the denoising process.

\subsection{$x$-pred Transformation}\label{subsec:x-pred_module}

To facilitate accurate classification of violative content at an early stage of the denoising process, we introduce an $x$-pred transformation that refines intermediate latent representations.
Recent flow-based generative models are typically designed with flow velocity prediction ($v$-pred) as the primary objective, which enables the estimation of the final latent from the predicted velocity and the intermediate latent.
Specifically, we compute the estimated final state of latent $\boldsymbol{x}_\theta(\boldsymbol{z}_t, t)$ via Eq.~\ref{eq:pseudox}. 

The transformation to $x$-pred is a well-established technique~\cite{li2026basicsletdenoisinggenerative}; although it is a simple approach, it proves to be effective. As illustrated in Fig~\ref{fig:ablation_booktabs} and Fig~\ref{fig:qwen_image_ablation}, clean images can be reconstructed even at an early stage of the process. %

\subsection{Similarity-based Scoring}\label{subsec:simulatiry_calculation}

The goal of similarity-based scoring is to detect violative content by comparing the embeddings of the decoded intermediate representation with the pre-computed reference matrix $\mathbf{R} \in \mathbb{R}^{n \times d}$. 
After obtaining the pseudo-clean latent $\boldsymbol{x}_\theta(\boldsymbol{z}_t, t)$ in Sec.~\ref{subsec:x-pred_module}, we evaluate its safety by mapping it into the shared embedding space.

We first project the latent $\boldsymbol{x}_\theta(\boldsymbol{z}_t, t)$ into the pixel space using the VAE decoder $D$ of the T2I model $G$. 
By passing the reconstructed image through the image encoder $E$, we obtain the query embedding $\boldsymbol{e}_t = E(D(\boldsymbol{x}_\theta(\boldsymbol{z}_t, t))) \in \mathbb{R}^d$. 
Then, the cosine similarity scores for all reference images can be efficiently computed via a single matrix-vector multiplication with $\mathbf{R}$:
\begin{equation}
    \boldsymbol{s} = \mathbf{R} \frac{\boldsymbol{e}_t}{\|\boldsymbol{e}_t\|},
\end{equation}
where $\boldsymbol{s} = [s_1, s_2, \dots, s_n]^\top \in \mathbb{R}^n$ represents the similarity scores for each reference image. 
Finally, the maximum similarity score $p = \max_i~s_i$ is used as the representative metric for safety classification.
A higher score indicates that the pseudo-clean latent aligns closely with reference images, triggering the safety filter.
The final classification decision is made by comparing the score to a threshold $\gamma$, where the generation process is rejected if $ p > \gamma$ and accepted otherwise.

\section{Experiment}

After introducing experimental setup (Sec.~\ref{subsec:setup}), we evaluate the effectiveness of EDGE-Shield compared to existing content filters (Sec.~\ref{Subsec:results}) with two key aspects: 
(\textit{\textbf{i}}) \textbf{Scalability}, demonstrating low latency and maintained classification ability as the number of target references increase; and 
(\textit{\textbf{ii}}) \textbf{Effectiveness}, showing faster classification while keeping classification ability comparable to existing baselines. 
Finally, we analyze the key properties and design choices of our method to justify its effectiveness (Sec.~\ref{subsec:anaysis}).
The code used for experiments will be published on acceptance.

\subsection{Setup}~\label{subsec:setup}

\noindent\textbf{T2I Model.}
We employ two state-of-the-art T2I models: Z-Image-Turbo~\cite{imageteam2025zimageefficientimagegeneration} and Qwen-Image~\cite{wu2025qwenimagetechnicalreport}. 
The inference steps of Z-Image-Turbo and Qwen-Image are set to the default values of 9 and 50 steps, respectively. 
To ensure deterministic reproducibility and a fair comparison, all experiments use fixed random seeds, ensuring that identical prompts generate identical output images across runs.

\noindent\textbf{Datasets.}
We evaluate our method on two datasets: HUB dataset~\cite{Moon_2025_ICCV_HUB} and CPDM dataset~\cite{ma2024datasetbenchmarkcopyrightinfringementCPDM}. 
The CPDM dataset comprises 200 individual faces and 81 intellectual properties (IPs), each with corresponding prompts and reference images. 
The HUB dataset contains 10 individual faces, 10 IPs, and 10 artistic styles, also with associated prompts and reference images. 
Importantly, in the original dataset papers, each prompt has been verified to enable the T2I model to successfully generate the corresponding the same category as the target reference, ensuring the validity of our evaluation.

\noindent\textbf{Evaluation Protocol.}
We evaluate EDGE-Shield from two perspective. 
(\textit{\textbf{i}}) Scalability: We assess that the proposed method maintain high accuracy and low latency even as the number of references increases. we use the CPDM dataset and incrementally increase the reference set size from 10 to 140 categories in steps of 10.\footnote{It is the available maximum number of categories in the dataset.} 
For a set of $N$ reference images, the evaluation includes $N$ matching prompts and $N$ unrelated prompts.
(\textit{\textbf{ii}}) Effectiveness: To evaluate overall classification accuracy and latency of blocking for generation by content filters, we perform experiments using a single-category reference set. 
This process is repeated across all single categories in the CPDM and HUB datasets, with one matching and one unrelated prompt per category.

To evaluate filtering performance at each timestep, we measure the similarity scores of EDGE-Shield at nine discrete timesteps for Z-Image-Turbo and at all 50 timesteps for Qwen-Image, respectively.
Furthermore, we investigated the impact of an image encoder $E$ on filtering performance by evaluating Qwen3-VL-embedding-2B~\cite{li2026qwen3vlembeddingqwen3vlrerankerunifiedframework}, CLIP~\cite{Radford2021clip}, SigLIP~\cite{Zhai_2023_ICCV_siglip}, and SigLIP2~\cite{tschannen2025siglip2multilingualvisionlanguage}.

\noindent\textbf{Baselines.}
We use reference-based content filters including perceptual-based methods and VLM-based methods as a baseline following the previous research~\cite{li2025t2isafety_imageguard,liu2025copyjudge}. Implementation details are provided in Appendix.

\begin{enumerate}
    \item \textbf{Perceptual-based methods}: Following the evaluation protocols established in previous research baselines, we directly compare the generated images with reference images. 
    Specifically, we employ Normalized L2~\cite{L2norm2023Carlini} and LPIPS~\cite{zhang2018perceptualLPIPS} to quantify image-to-image similarity.
    As a state-of-the-art among perceptual-based methods, we employ DiffSim~\cite{song2024diffsim}.
    \item \textbf{VLM-based methods}: VLM-based method includes the use of LLaVaGuard~\cite{helff2025llavaguard} as a specialized model for safety assessment. Furthermore, we evaluate performance using several open-source VLMs, including InternVL3.5-8B~\cite{wang2025internvl35advancingopensourcemultimodal}, Qwen3-VL-8B~\cite{bai2025qwen3vltechnicalreport}, and LLaVA-NEXT-7B~\cite{liu2023llava} with VLLM~\cite{kwon2023vllm} for fast inference. 
    Following the prompt of LLaVAGuard, we use the prompt, e.g., \textit{"Compare Image 2 (generated) to Image 1 (reference) for (O1) IP violation, (O2) right of publicity, and (O3) style mimicry, then output 'True' if any violation is found or 'False' otherwise."} for VLM-based methods.
    We use $P("True")$ as an output score to get the probability for each inference.
    Additionally, we assess the performance of gpt-4o-mini~\cite{openai2024gpt4ocard} whose implementation detailed in Appendix.
\end{enumerate}

\noindent\textbf{Evaluation Metrics.}
To evaluate both classification ability and efficiency, we measure ROC-AUC, PR-AUC, and latency. 
ROC-AUC and PR-AUC measure the classification ability without being affected by the threshold, respectively. 
For latency, we measure the time from the start of the generation process until the content filter outputs the similarity score. 
For our proposed method, the latency refers the time from when the generation starts to when EDGE-Shield outputs the similarity score. 
For baselines, it is measured as the total time from the start of generation to its completion, followed by the classification process on the resulting image.

\subsection{Comparison with Baselines}\label{Subsec:results}
We here show the performance of EDGE-Shield compared to baseline methods. 
First, we examine how time to classification scales with an increasing number of references. 
Subsequently, we analyze the time to classification and AUC to highlight the efficiency of our approach.

\begin{figure}[t]
  \centering
  
  \begin{subfigure}[b]{0.45\textwidth}
    \centering
    \includegraphics[width=\textwidth]{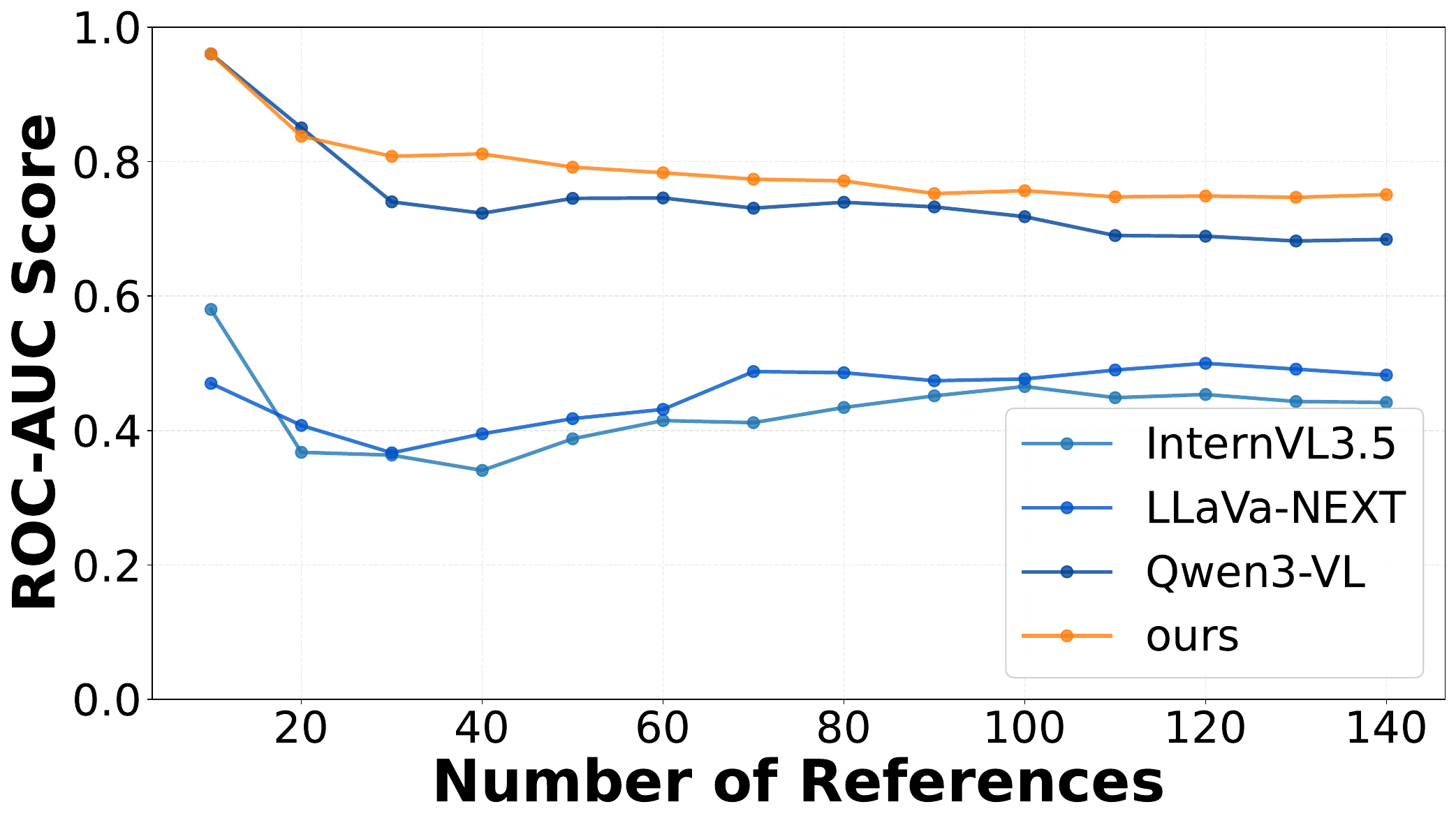}
    \caption{}
    \label{fig:two_sub1}
  \end{subfigure}
  \hfill %
  \begin{subfigure}[b]{0.45\textwidth}
    \centering
    \includegraphics[width=\textwidth]{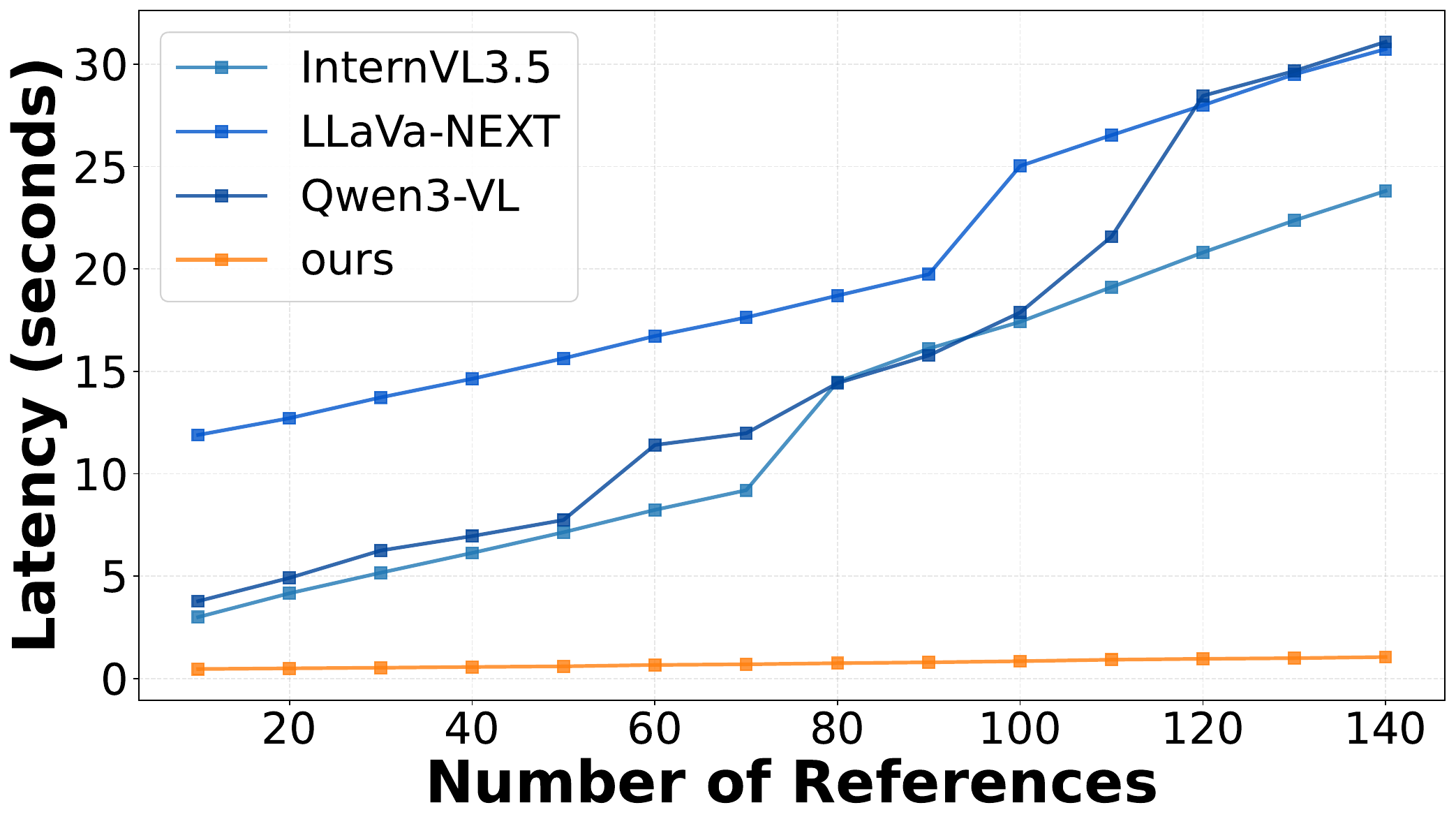}
    \caption{}
    \label{fig:two_sub2}
  \end{subfigure}
  \caption{Evaluation results of ROC-AUC, Latency, and the number of references on Z-Image-Turbo.
  (a) Relationship between the number of references and the ROC-AUC score. (b) Relationship between the number of references and the latency time.}
  \label{fig:scalability}
\end{figure}

\noindent\textbf{Scalability.}
Fig. \ref{fig:scalability} demonstrates the robustness and efficiency of the proposed method across varying reference set sizes.
As shown in the left plot, our method exhibits a positive correlation between the number of reference images and classification performance, consistently achieving superior ROC-AUC scores.
Simultaneously, the right plot evaluates the corresponding computational cost.
While conventional filters suffer from prohibitive computational overhead as the reference set expands, our method maintains a near-constant inference latency by leveraging reference caching, thereby ensuring high scalability

\noindent\textbf{Effectiveness.}
Table~\ref{tbl:efficiency} summarizes the filtering performance of EDGE-Shield in comparison to existing baselines.
Our method achieves the highest scores in both ROC-AUC and PR-AUC across the experimental setup.
Notably, EDGE-Shield significantly outperforms perceptual-based metrics—such as Normalized $L_2$ and LPIPS—as well as established VLM-based filters.
Furthermore, it surpasses the performance of the highly capable Qwen3-VL, demonstrating its superior ability to accurately identify violative content.

\begin{table}[t]
\setlength{\tabcolsep}{2pt}
\caption{ROC-AUC and PR-AUC scores of violation binary classification for CPDM and HUB, and the time by the average score of single generation. 
The score of EDGE-Shield shown in this table is calculated at step 1 for Z-Image, at step 25 for Qwen-Image, respectively.
}\label{tbl:efficiency}
\begin{tabular}{@{}lcccccc@{}}
\toprule
\multirow{2}{*}{T2I Model} & \multicolumn{3}{c}{Z-Image} & \multicolumn{3}{c}{Qwen-Image} \\ \cmidrule(lr){2-4} \cmidrule(l){5-7} 
                           & ROC-AUC & PR-AUC & Time (s) & ROC-AUC & PR-AUC     & Time (s) \\ \midrule
Normalized   $L_2$~\cite{L2norm2023Carlini} & 0.417 & 0.449 & 2.137 & 0.431 & 0.454 & 22.003 \\ 
LPIPS~\cite{zhang2018perceptualLPIPS}              & 0.407 & 0.441 & 2.505 & 0.428 & 0.451 & 22.016 \\
DjffSim~\cite{song2024diffsim}            & 0.706 & 0.758 & 2.168 & 0.719 & 0.773 & 22.233 \\ \midrule
LLaVaGuard~\cite{helff2025llavaguard}         & 0.493 & 0.497 & 2.425 & 0.474 & 0.483 & 22.498 \\
LLaVa-NEXT~\cite{liu2024llavanext}         & 0.493 & 0.492 & 2.257 & 0.494 & 0.494 & 22.868 \\
InternVL3.5~\cite{wang2025internvl35advancingopensourcemultimodal}        & 0.541 & 0.523 & 2.251 & 0.522 & 0.524 & 23.820 \\
Qwen3-VL~\cite{bai2025qwen3vltechnicalreport}           & 0.841 & 0.888 & 2.413 & 0.831 & 0.875 & 23.081 \\ \midrule
gpt-4o-mini~\cite{openai2024gpt4ocard} & 0.696   & 0.705  & 25.656   & 0.691   & 0.714  & 32.423   \\ \midrule
Ours w/ CLIP               & 0.827 & 0.846 & 0.404 & 0.830 & 0.844 & 11.621 \\
Ours w/ SigLIP             & 0.846 & 0.875 & \textbf{0.402} & 0.843 & 0.862 & 11.619 \\
Ours w/ SigLIP2            & 0.835 & 0.869 & 0.407 & 0.833 & 0.854 & \textbf{11.617} \\
Ours w/ Q3VLEmbed  & \textbf{0.857} & \textbf{0.898} & 0.454 & \textbf{0.844} & \textbf{0.883} & 12.107 \\ \bottomrule
\end{tabular}
\end{table}

\begin{figure}[t]
  \centering
  \begin{subfigure}[b]{0.48\linewidth}
    \centering
    \includegraphics[width=\linewidth]{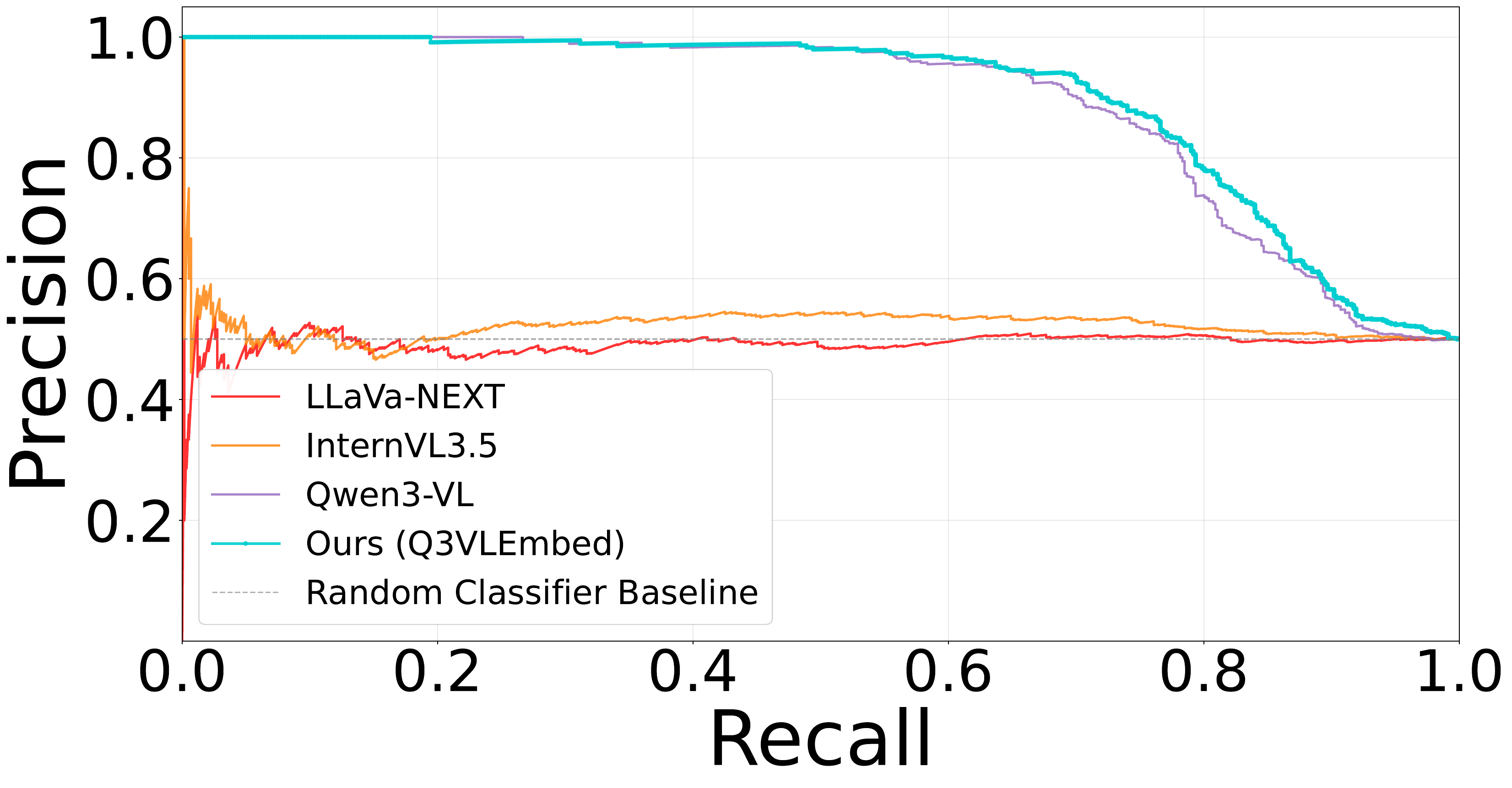}
    \caption{Z-Image-Turbo}
    \label{fig:ts0}
  \end{subfigure}  
  \hfill
  \begin{subfigure}[b]{0.48\linewidth}
    \centering
    \includegraphics[width=\linewidth]{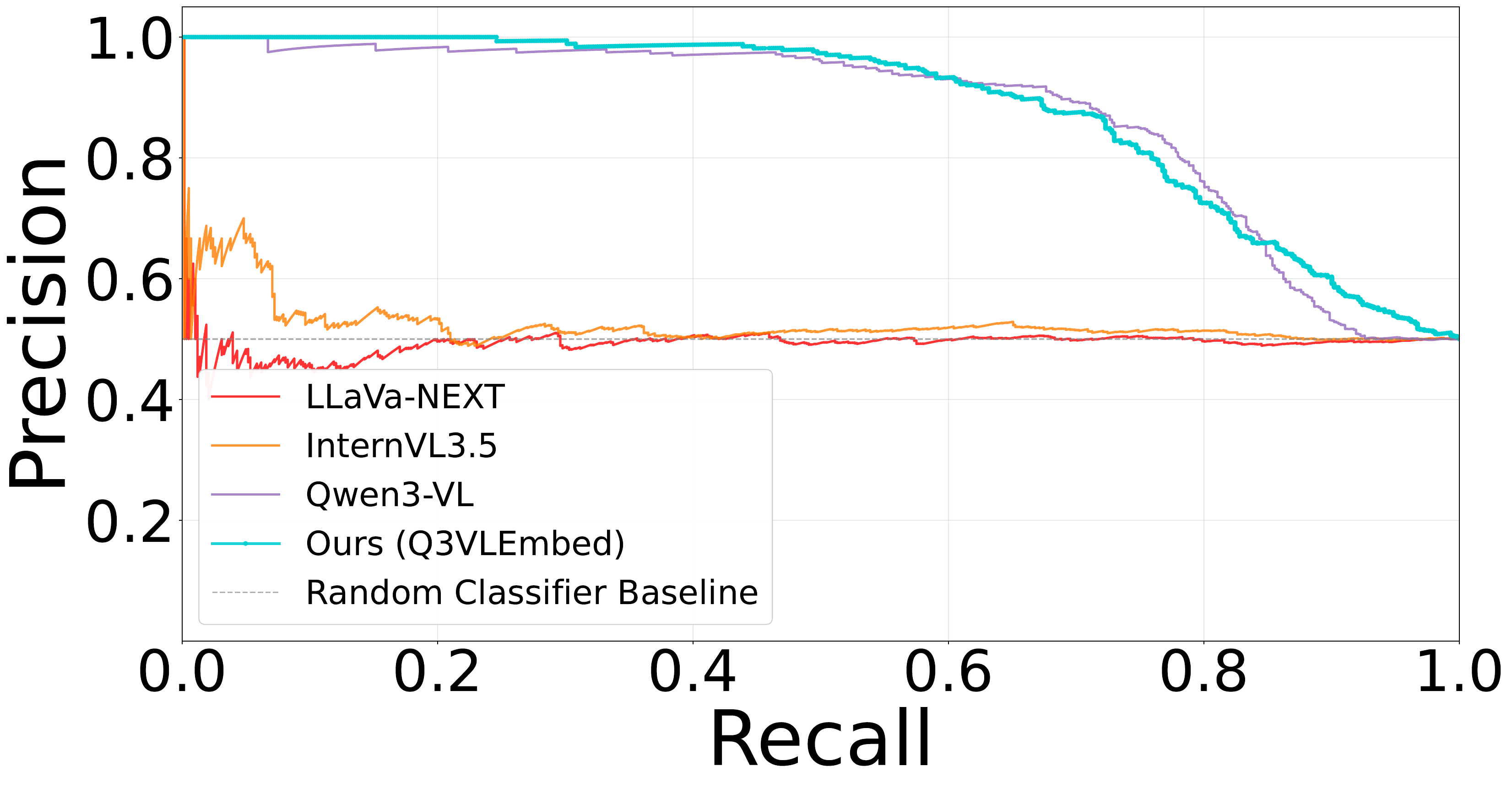}
    \caption{Qwen-Image}
    \label{fig:ts30}
  \end{subfigure}  
  
  \caption{Visualization of Precision-Recall curve.}
  \label{fig:efficiency_prcurve}
\end{figure}

More importantly, Table~\ref{tbl:efficiency} also shows the advantage of EDGE-Shield lies in classification ability in the earlier time. 
While baseline methods require more than 2.1 seconds to evaluate and halt the generation, our method drastically reduces this latency to just 0.454 seconds for Z-Image-Turbo. 

Figure~\ref{fig:efficiency_prcurve} presents the Precision-Recall curves, illustrating that our proposed method achieves a superior trade-off between recall and precision.
EDGE-Shield dominates the other models by occupying the largest area in the upper-right region.
Notably, it maintains near-ideal precision up to a recall level of 0.6 and exhibits a distinct performance gain over the Qwen3-VL baseline in the high-recall regime (0.7–0.9).
These results underscore the effectiveness of our approach in identifying violative content with minimal false positives, even under stringent recall requirements.

\subsection{Analysis}\label{subsec:anaysis}

Here, we analyze the effectiveness of each EDGE-Shield component and the filtering ability across different categories.
Specifically, we conduct an ablation study to verify that the $x$-pred transformation effectively enhances classification performance during the denoising process. 
We also examine how the choice of image encoder $E$ influences filtering capability. 
Furthermore, we assess threshold-wise analysis of the similarity score and the consistency of filtering performance across diverse categories, including intellectual property (IP), individual faces, and artistic styles.

\begin{figure}[t]
  \centering
  \begin{subfigure}{0.48\linewidth}
    \centering
    \includegraphics[width=\linewidth]{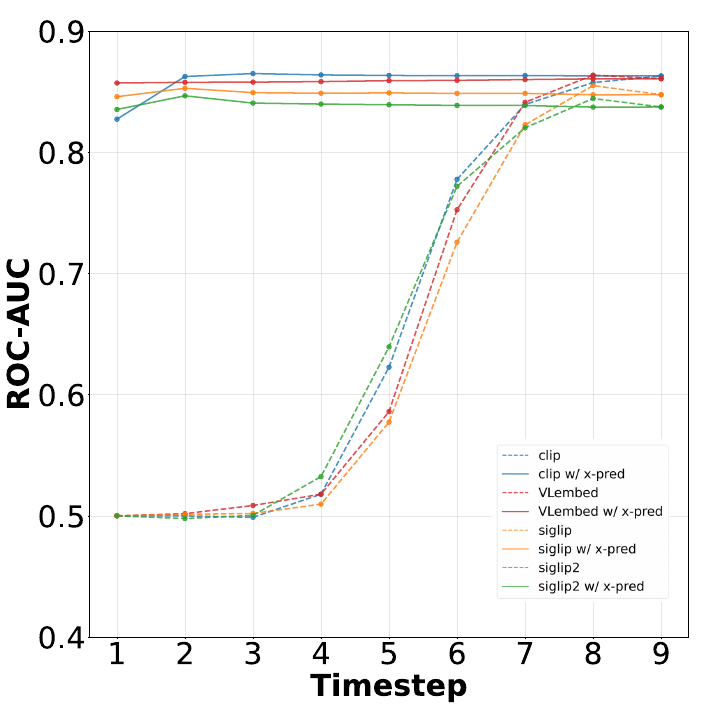}
    \caption{Z-Image-Turbo}
    \label{fig:ablation-qwen}
  \end{subfigure}
  \begin{subfigure}{0.48\linewidth}
    \centering
    \includegraphics[width=\linewidth]{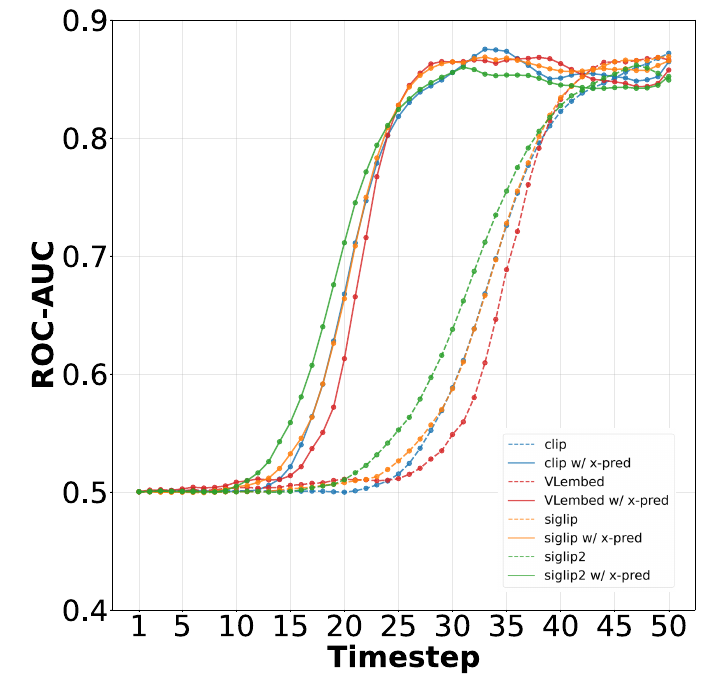}
    \caption{Qwen-Image} %
    \label{fig:ablation-third}
  \end{subfigure}
  \caption{ROC-AUC visualization of $x$-pred transformation ablation. This figure shows the ROC-AUC scores across each timesteps for three different models.}
  \label{fig:x-pred-ablation}
\end{figure}

\noindent\textbf{\textit{x}-pred Transformation Ablation.}
Fig.~\ref{fig:x-pred-ablation} (a) demonstrates that by introducing $x$-pred transformation, violative content can be determined at an early timestep on Z-Image-Turbo.
As shown, EDGE-Shield without $x$-pred transformation struggle to identify violative content during the initial generation stages, yielding scores of approximately 0.500 from $T=1$ to $T=3$. 
They only begin to show reliable performance at later timesteps ($T=7$ to $T=9$). 
In contrast, integrating the $x$-pred transformation triggers an immediate and dramatic performance boost right from $T=1$.  
Fig.~\ref{fig:x-pred-ablation} (b) shows that while introducing our $x$-pred transformation to Qwen-Image also enables the classification of violative content at earlier timesteps, the improvement in early-stage classification capability is not as significant as that observed in Z-Image-Turbo.
Specifically, a noticeable performance improvement is observed between steps $T=20$ and $T=30$.

\begin{figure}[t]
  \centering
  \small %
  \resizebox{0.9\textwidth}{!}{\begin{tabular}{c@{\hspace{0.4mm}}ccccccccc}
    \toprule
    Timestep ($T$) & $1$ & $2$ & $3$ & $4$ & $5$ & $6$ & $7$ & $8$ & $9$ \\
    \midrule
    
    \begin{tabular}[c]{@{}c@{}}Vanilla \\ $D(\boldsymbol{z}_t)$\end{tabular} & 
    \includegraphics[width=.09\textwidth, valign=c]{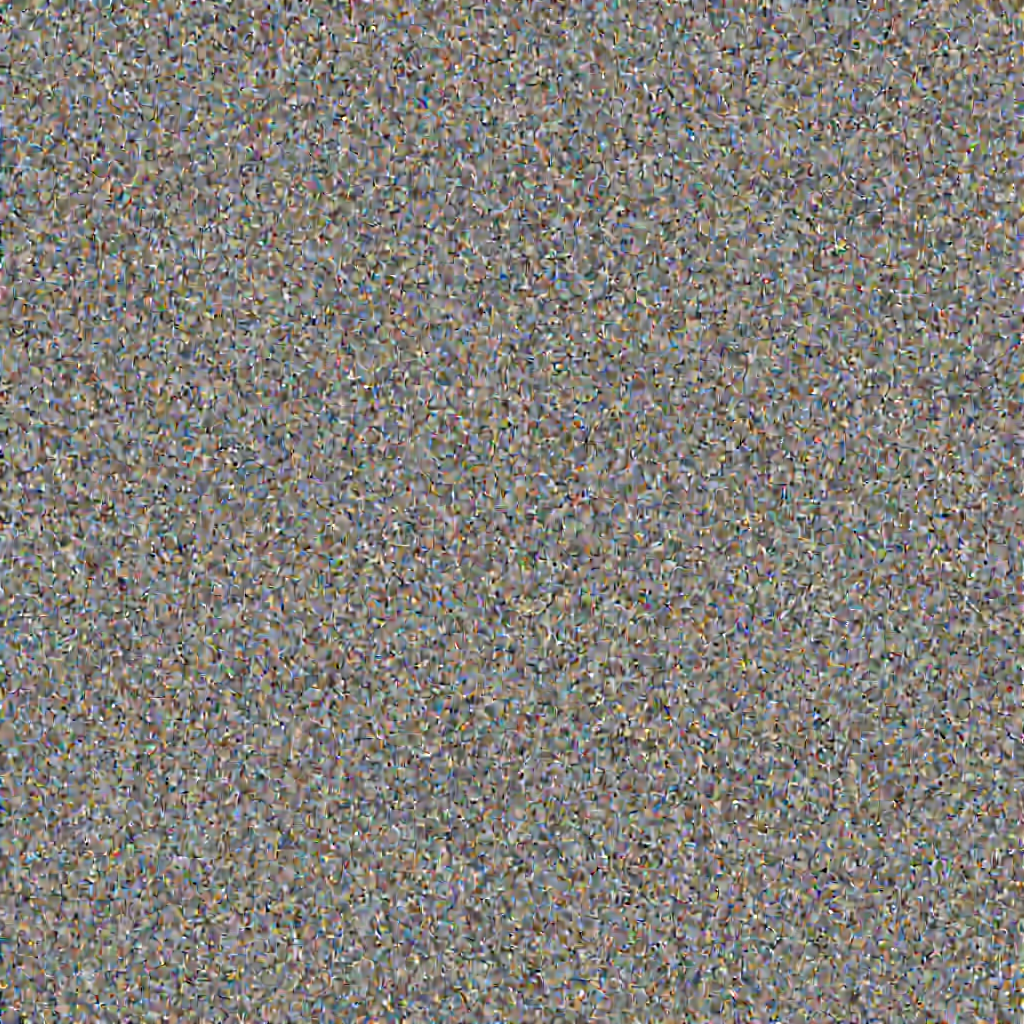} &
    \includegraphics[width=.09\textwidth, valign=c]{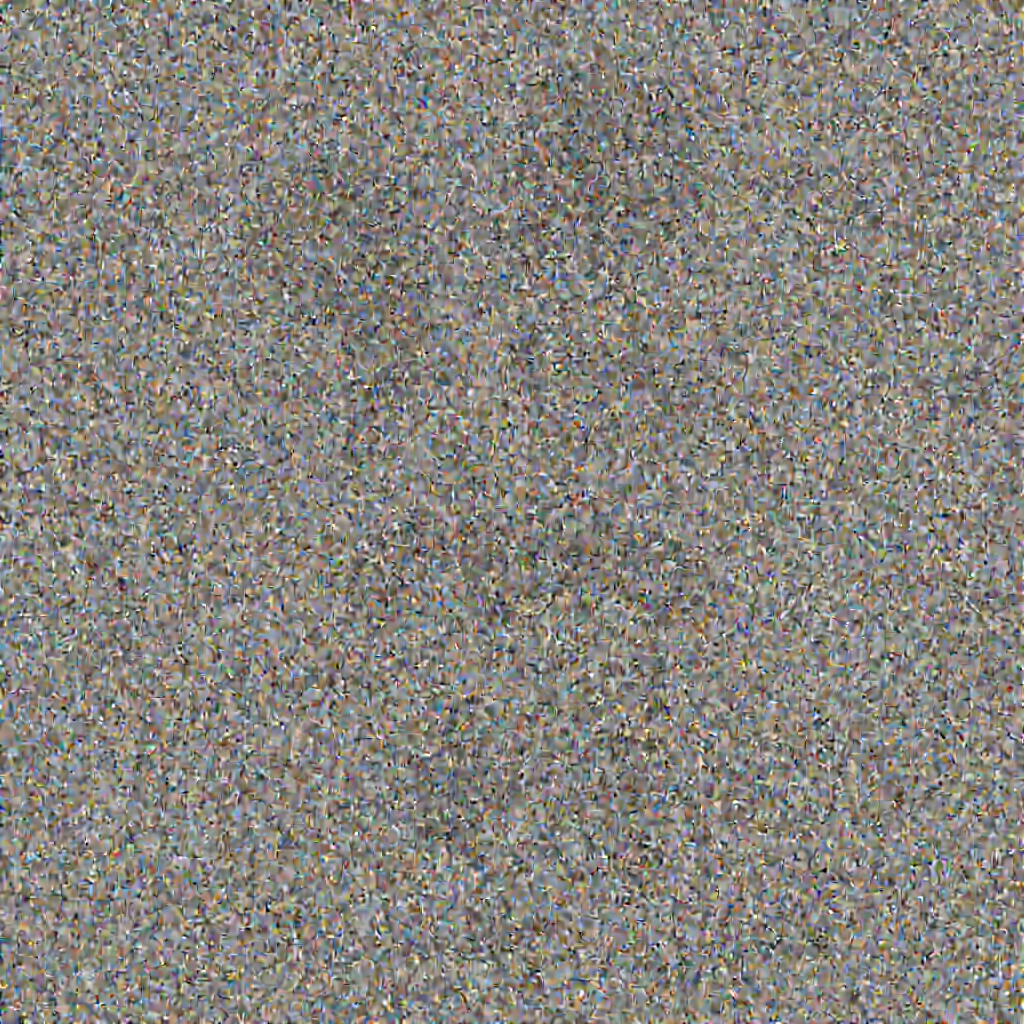} &
    \includegraphics[width=.09\textwidth, valign=c]{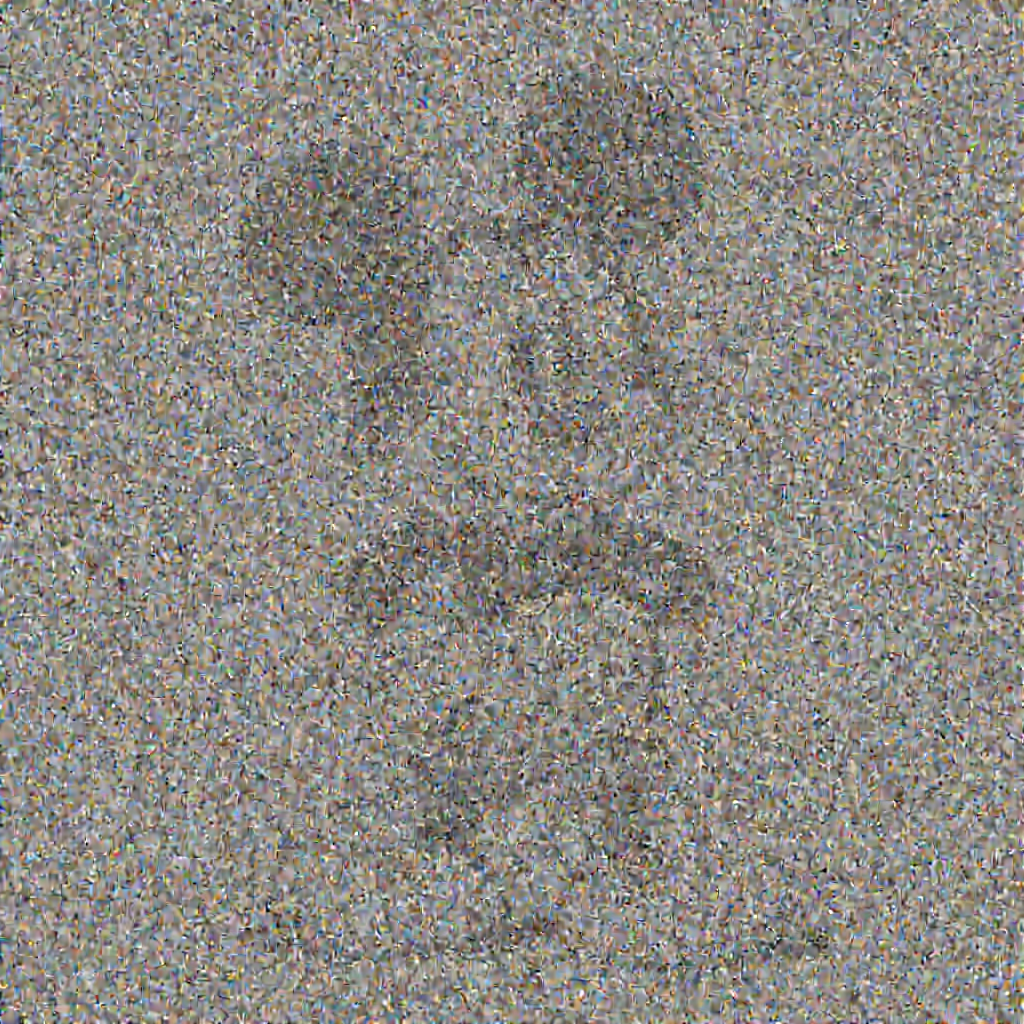} &
    \includegraphics[width=.09\textwidth, valign=c]{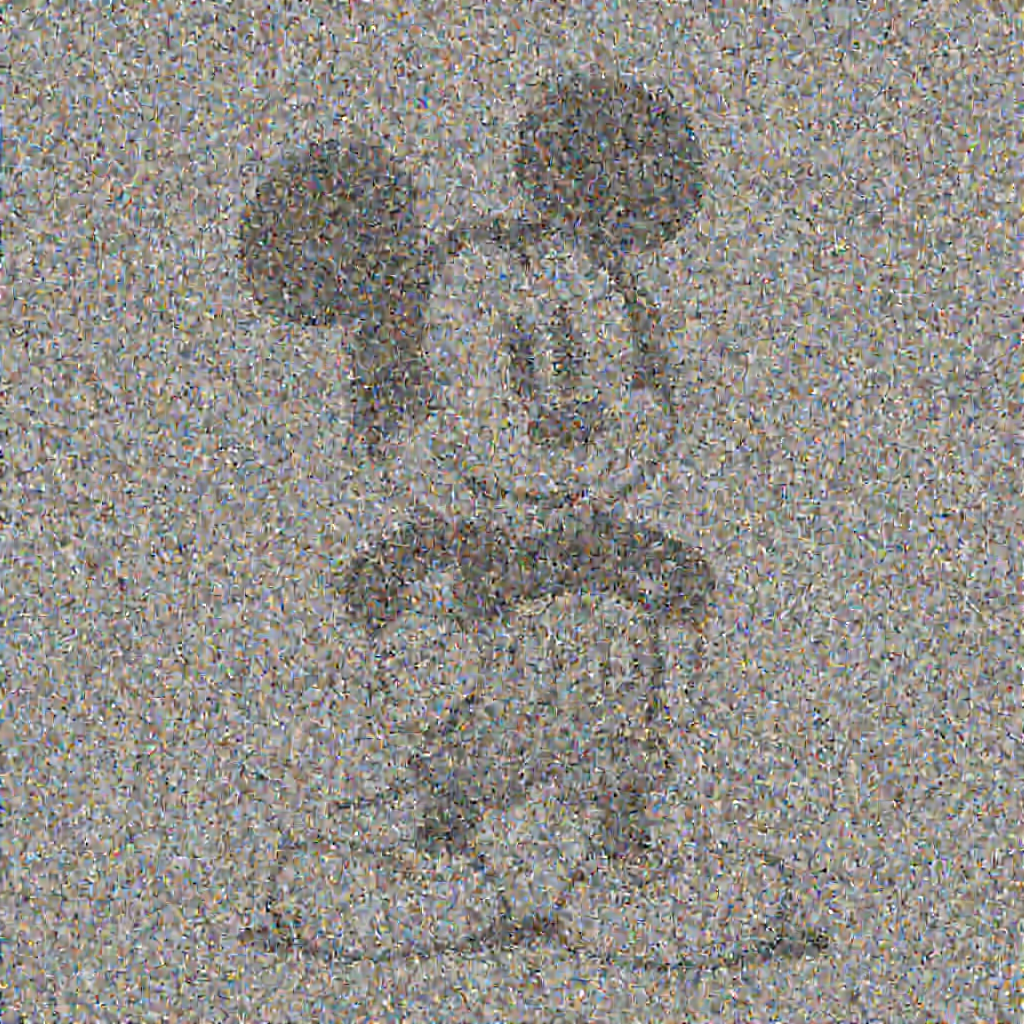} &
    \includegraphics[width=.09\textwidth, valign=c]{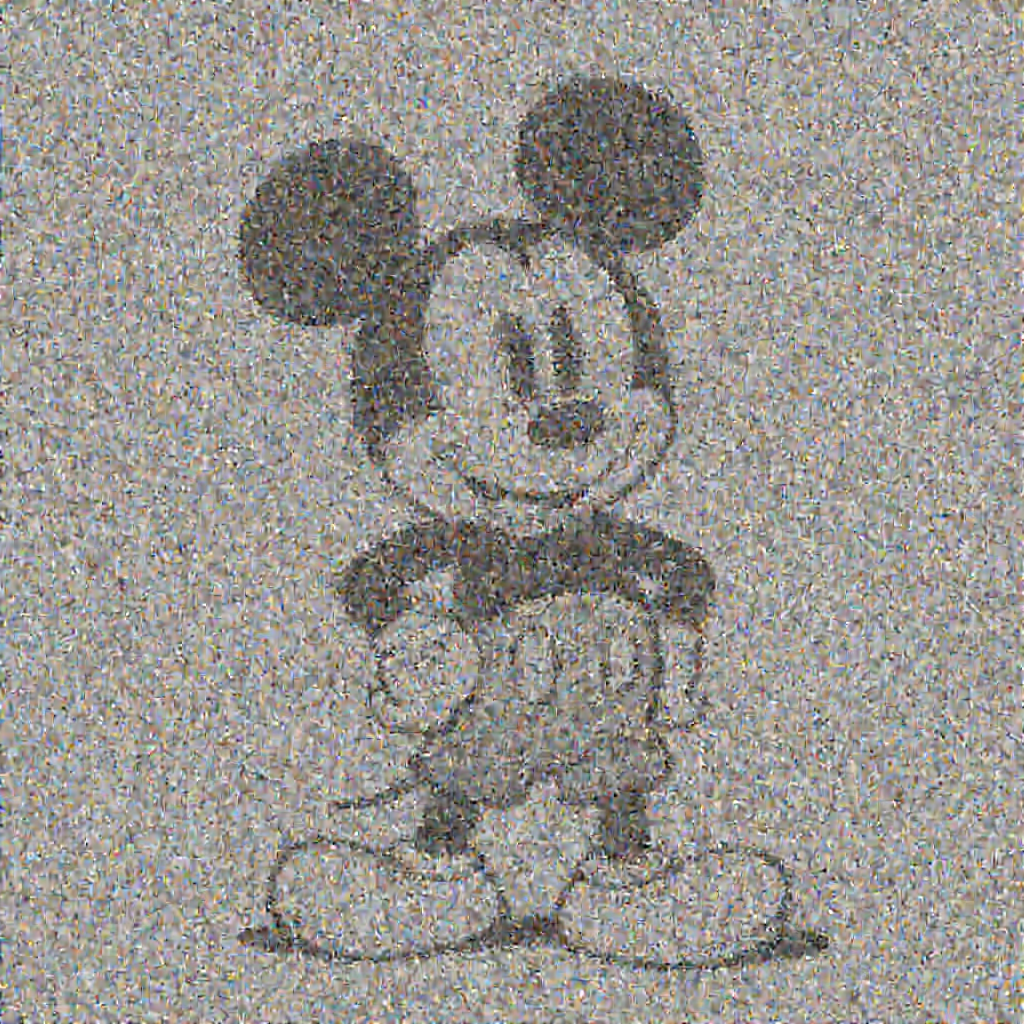} &
    \includegraphics[width=.09\textwidth, valign=c]{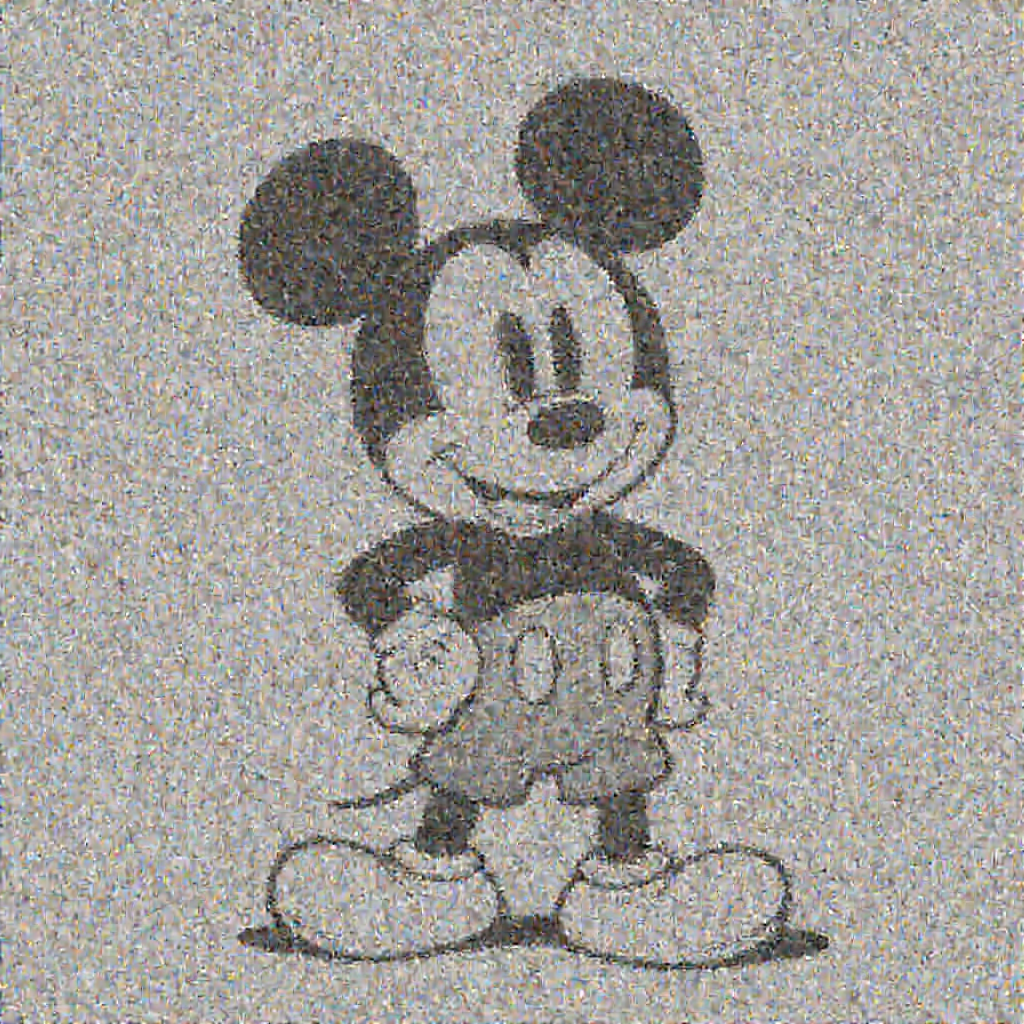} &
    \includegraphics[width=.09\textwidth, valign=c]{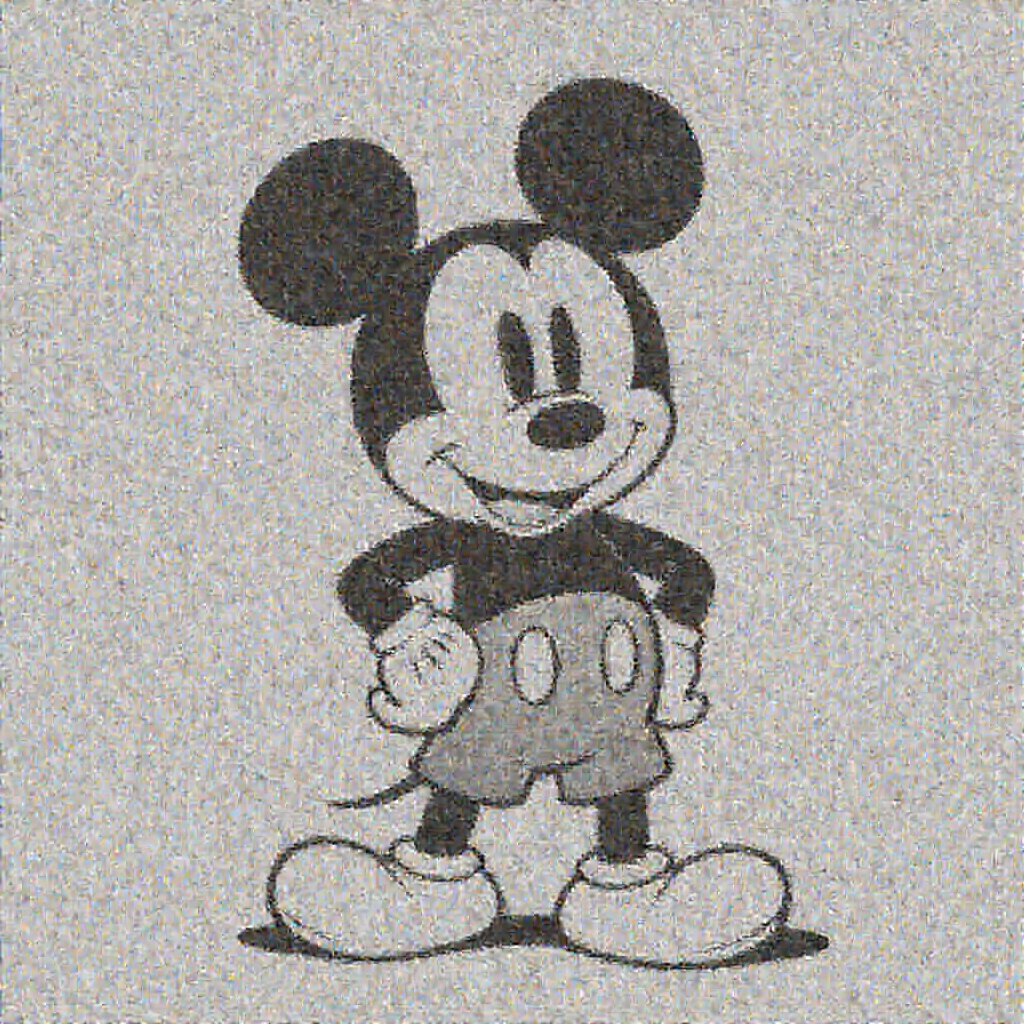} &
    \includegraphics[width=.09\textwidth, valign=c]{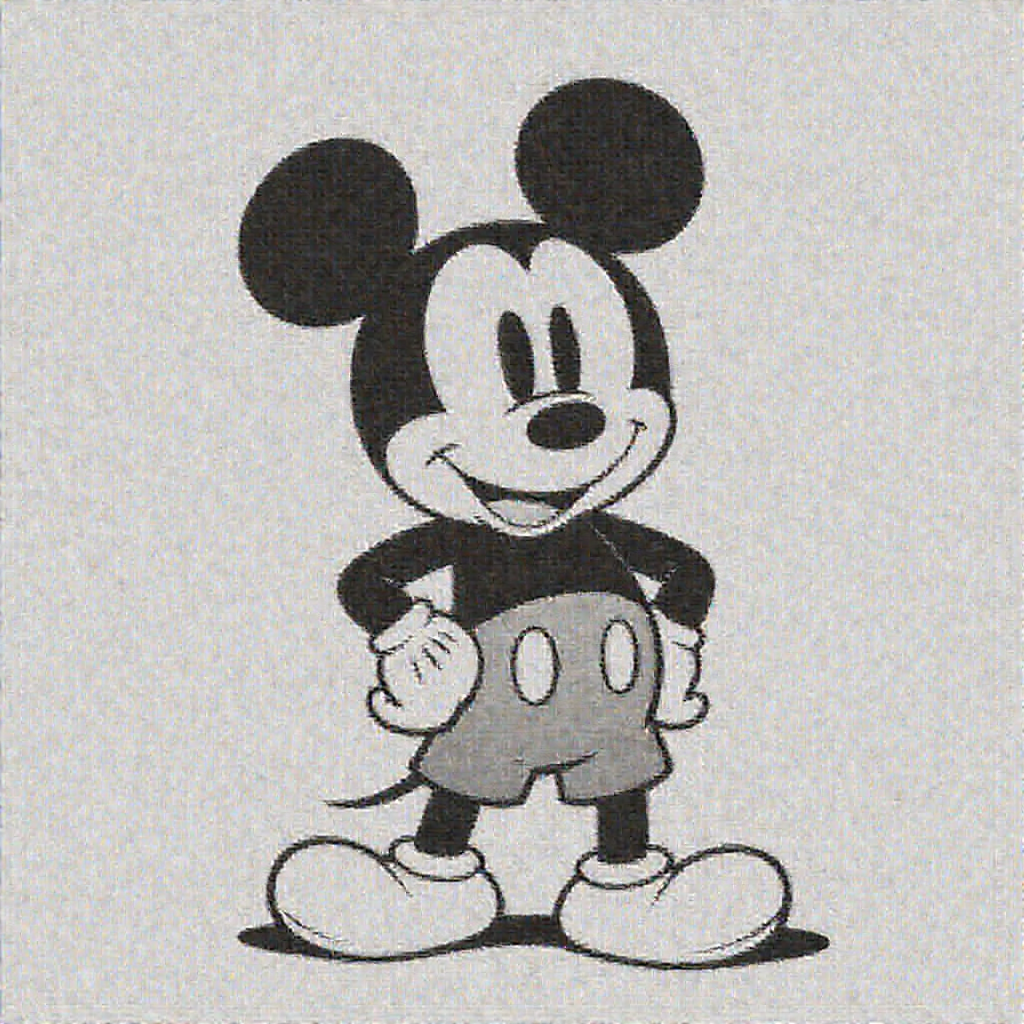} &
    \includegraphics[width=.09\textwidth, valign=c]{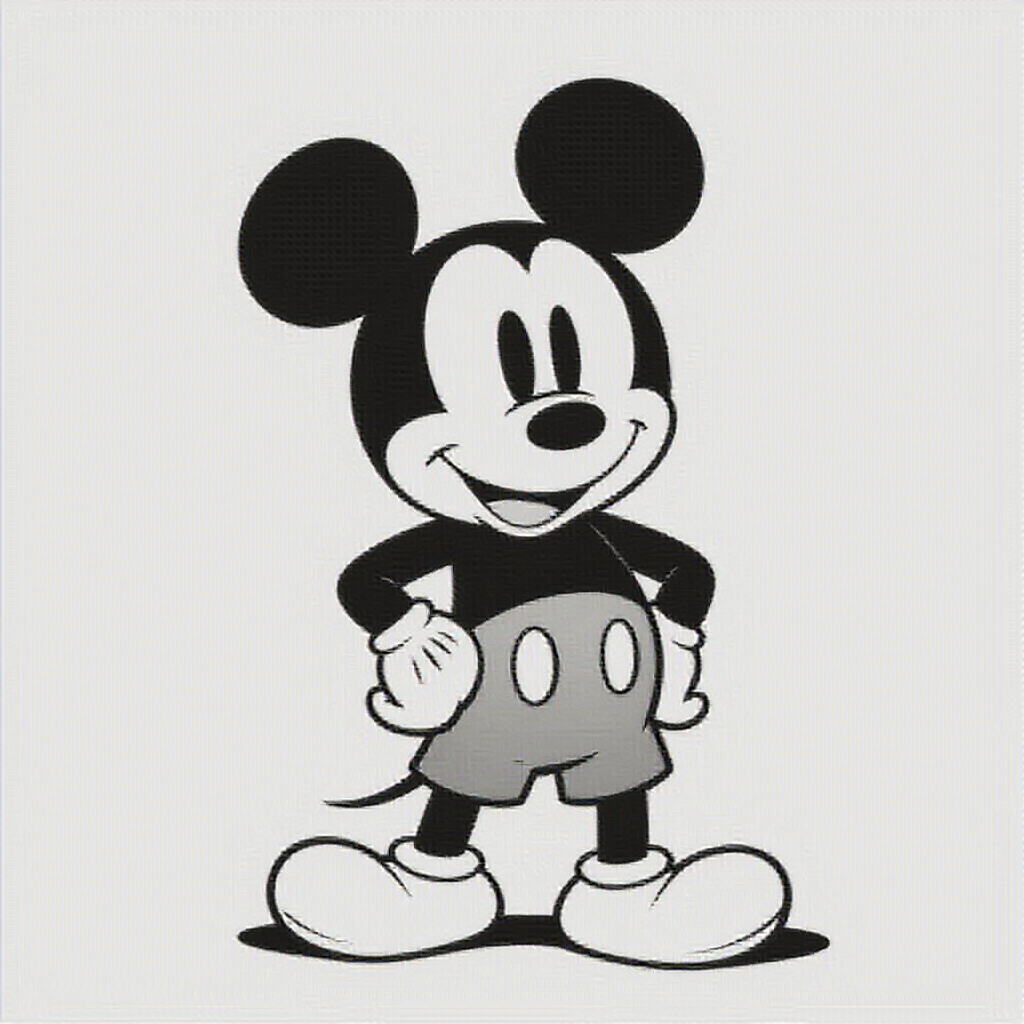}  \\
    \addlinespace[2mm] %
    
    \begin{tabular}[c]{@{}c@{}}$x$-pred \\ $D(\boldsymbol{x}_\theta(\boldsymbol{z}_t, t))$\end{tabular} & 
    \includegraphics[width=.09\textwidth, valign=c]{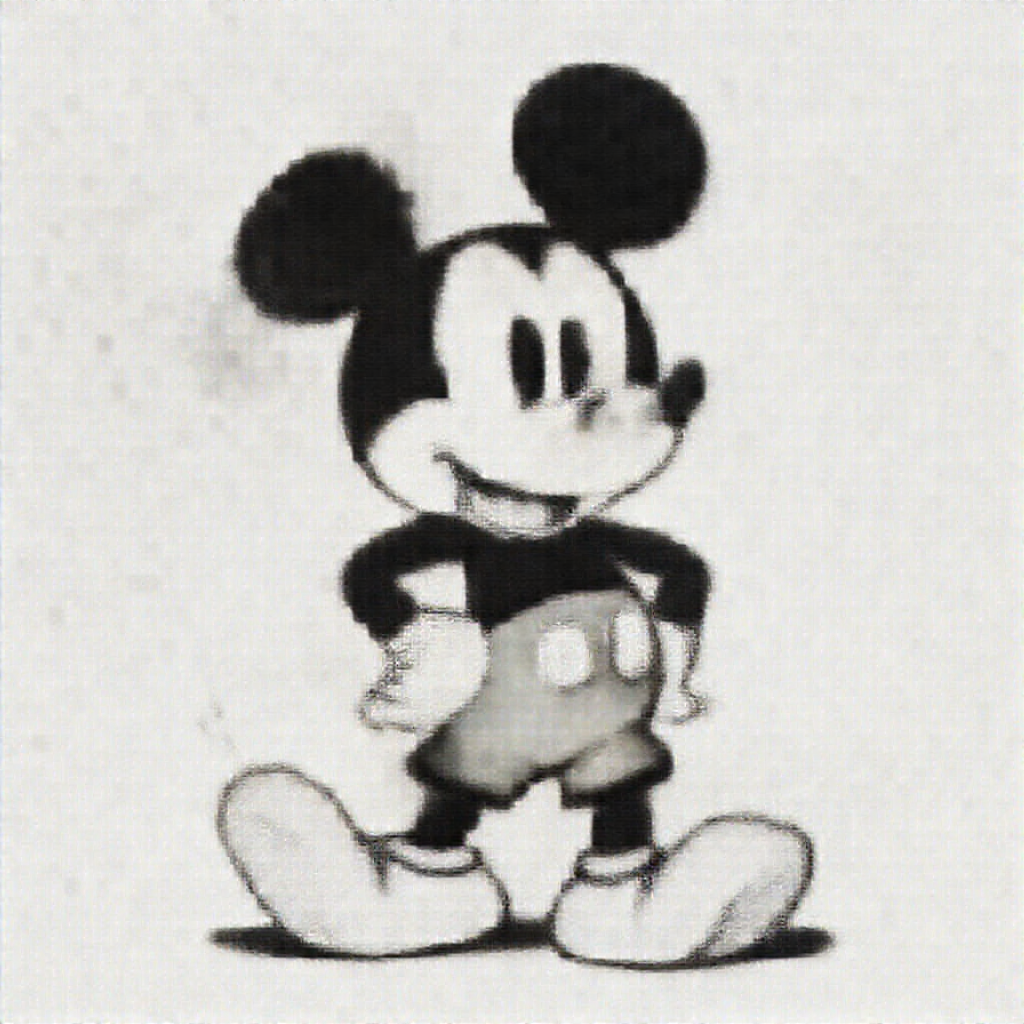} &
    \includegraphics[width=.09\textwidth, valign=c]{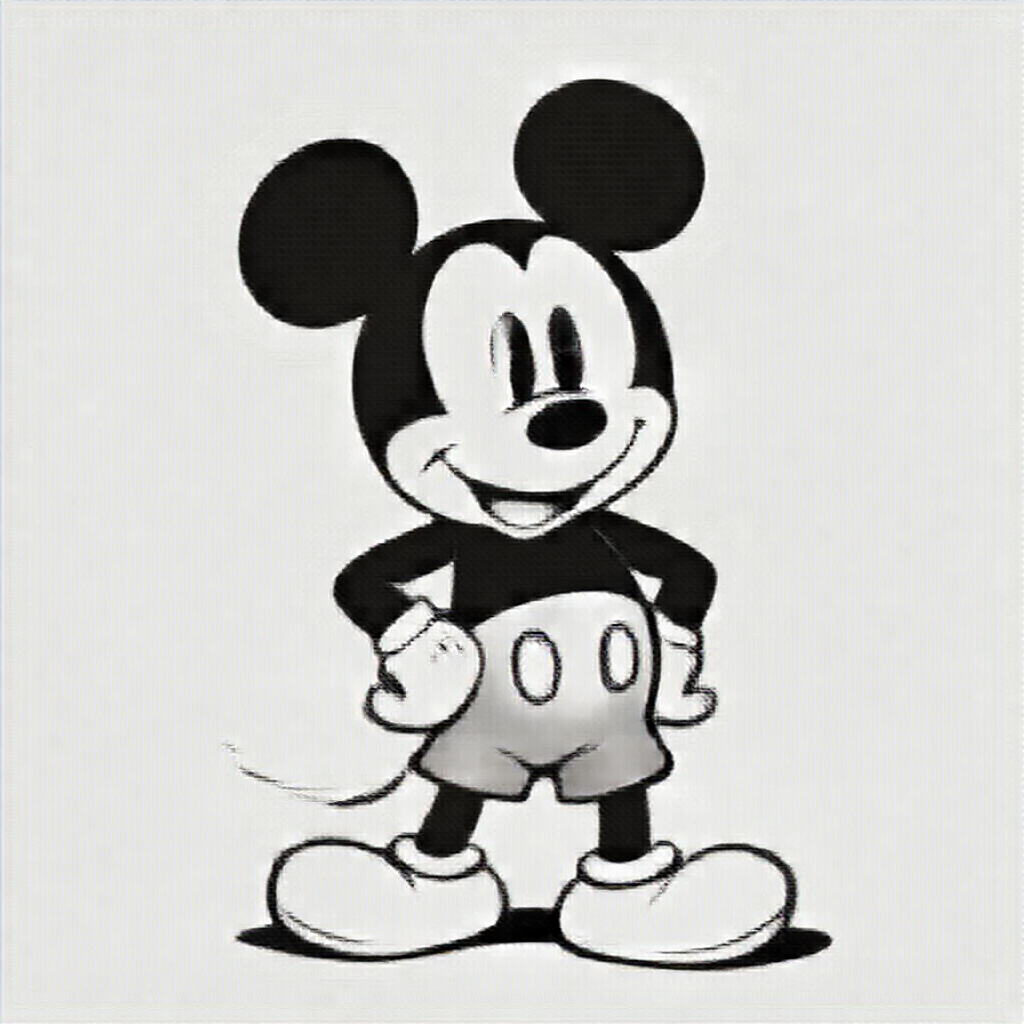} &
    \includegraphics[width=.09\textwidth, valign=c]{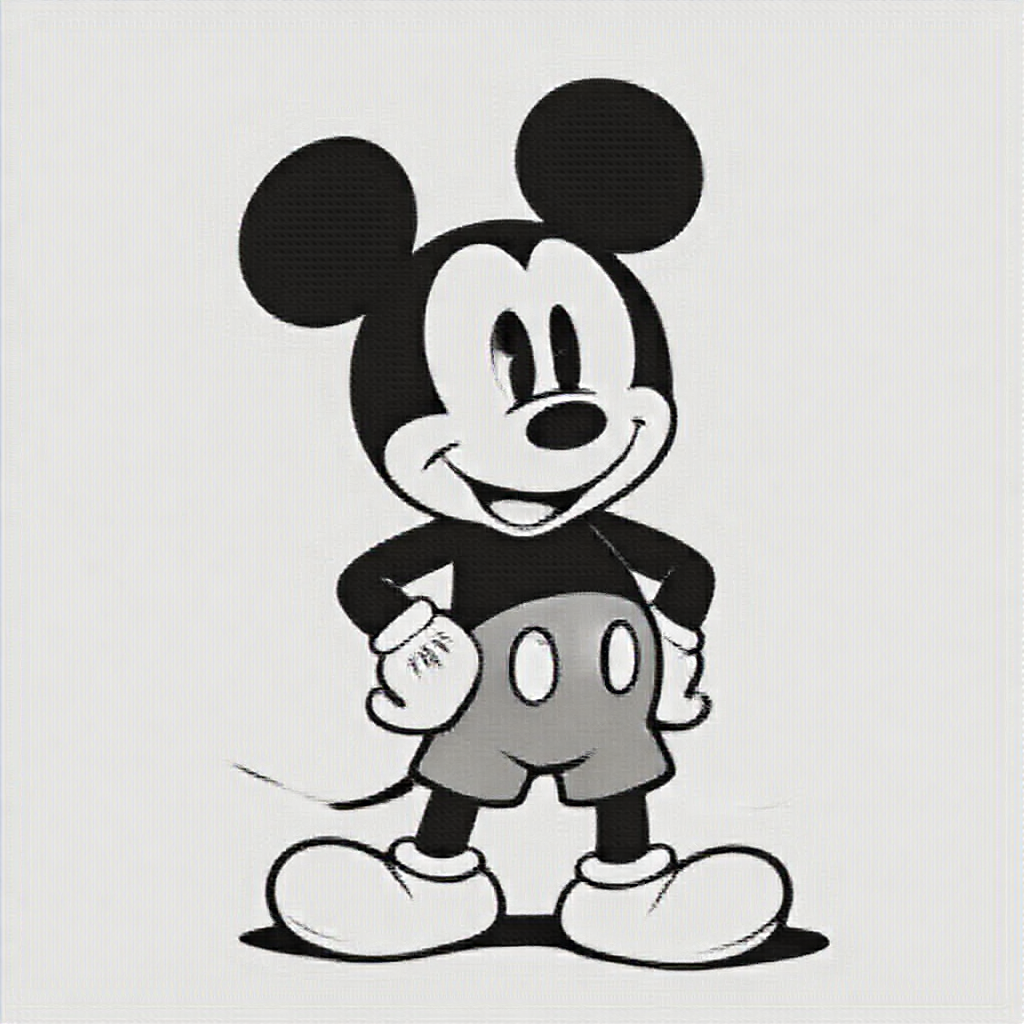} &
    \includegraphics[width=.09\textwidth, valign=c]{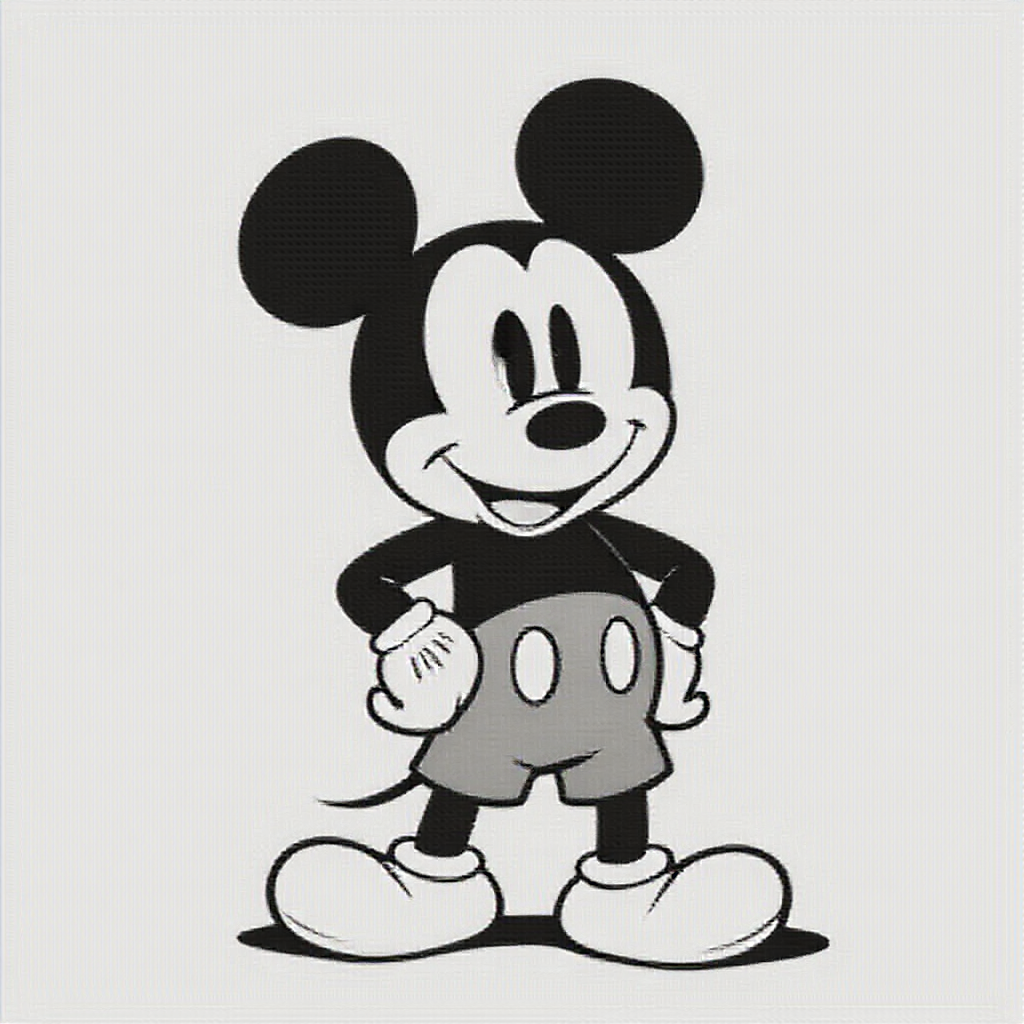} &
    \includegraphics[width=.09\textwidth, valign=c]{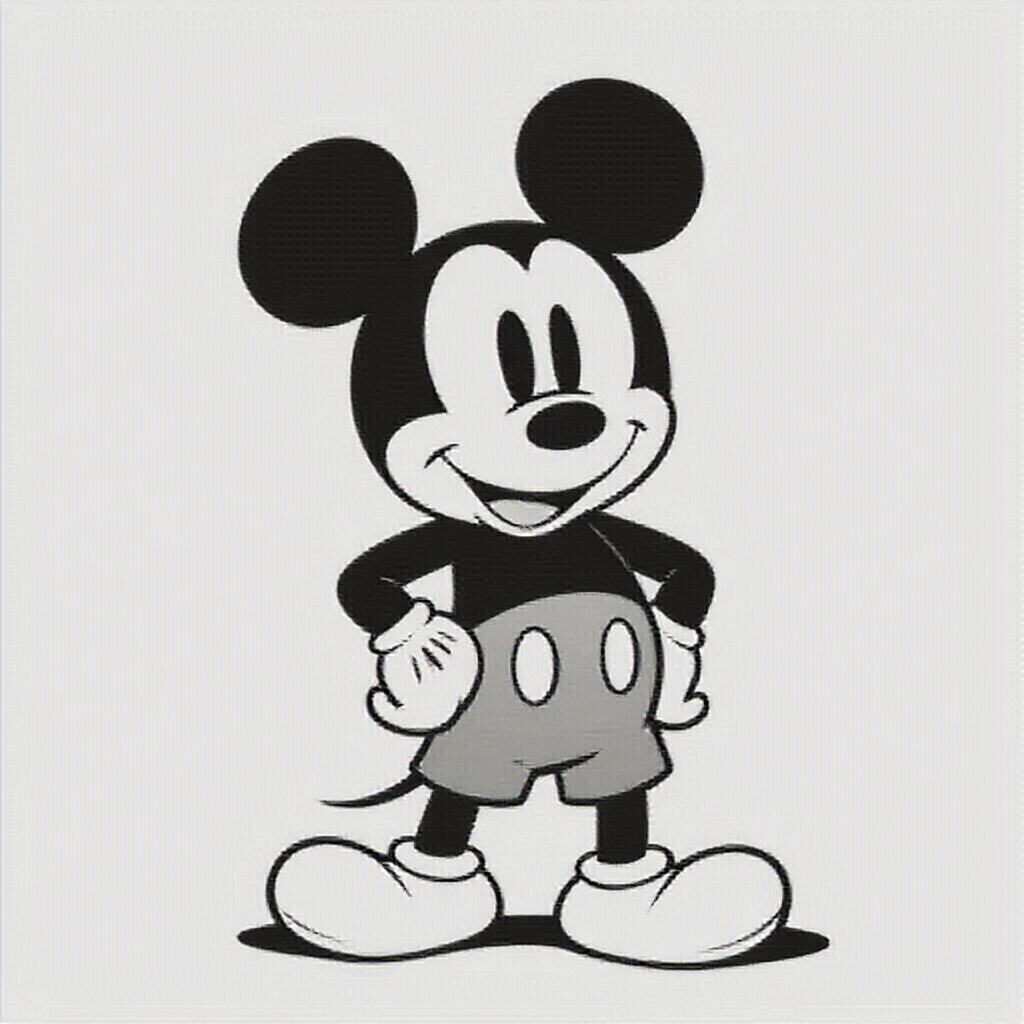} &
    \includegraphics[width=.09\textwidth, valign=c]{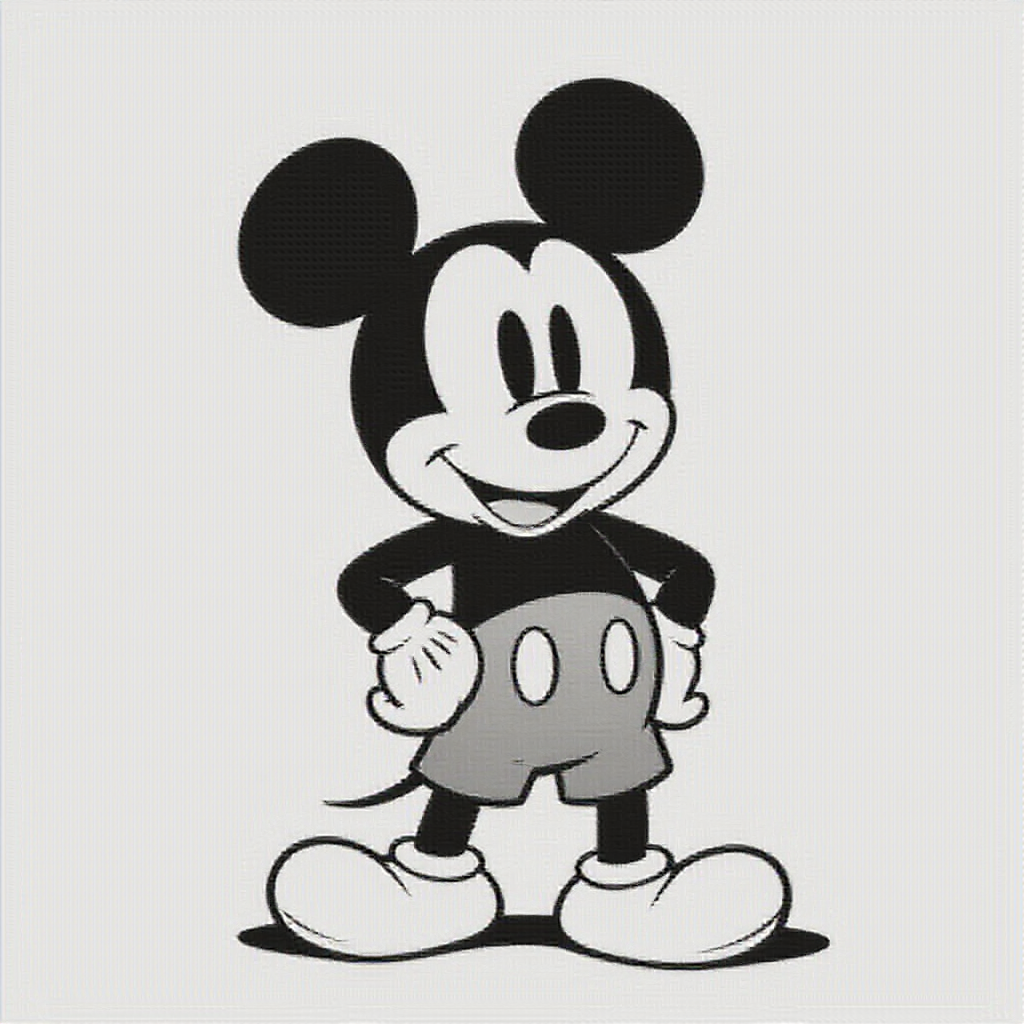} &
    \includegraphics[width=.09\textwidth, valign=c]{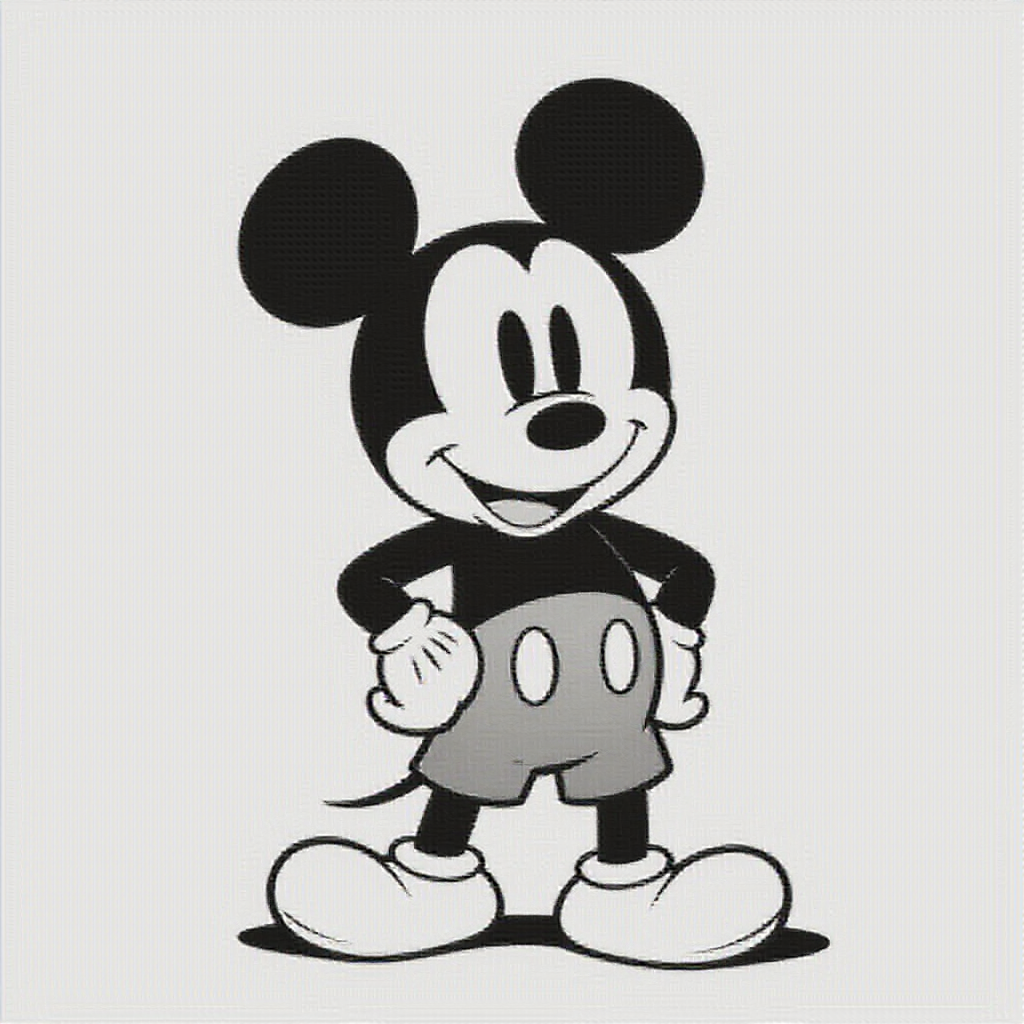} &
    \includegraphics[width=.09\textwidth, valign=c]{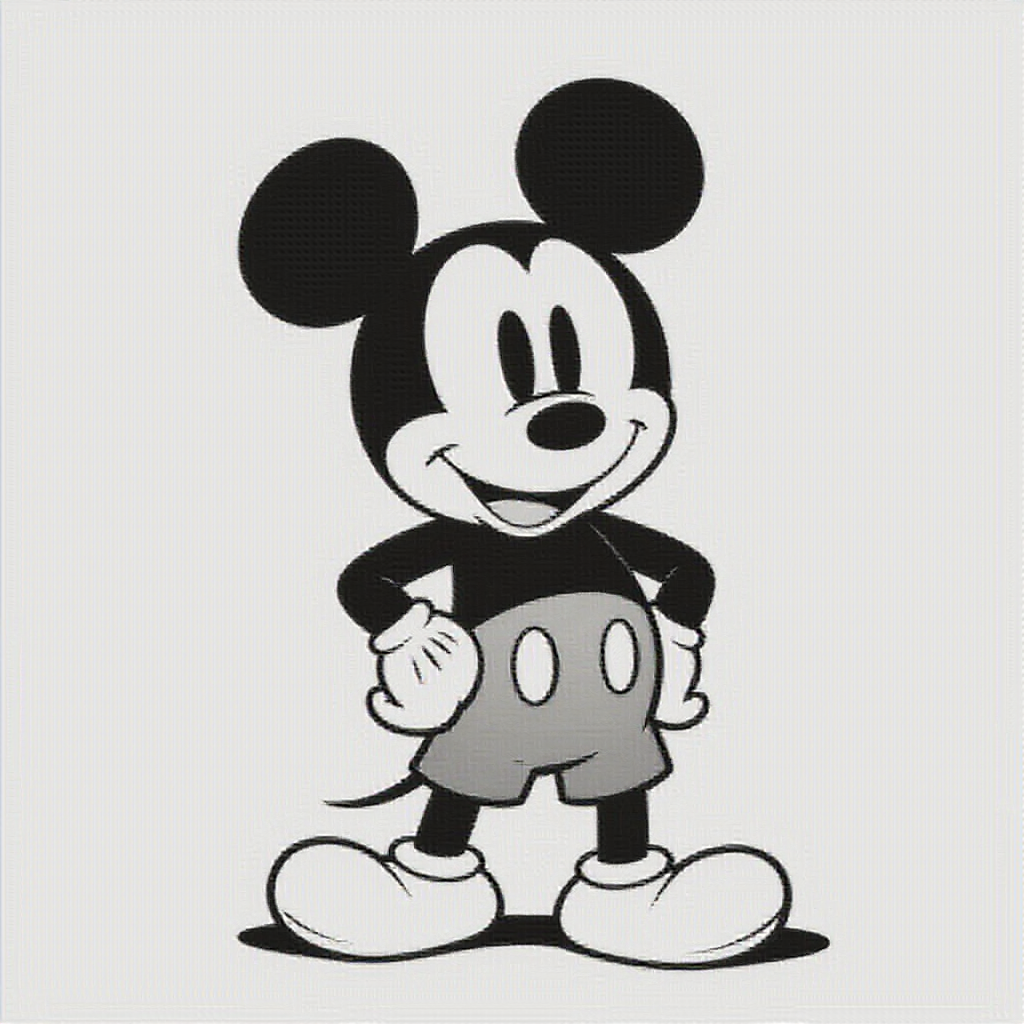} &
    \includegraphics[width=.09\textwidth, valign=c]{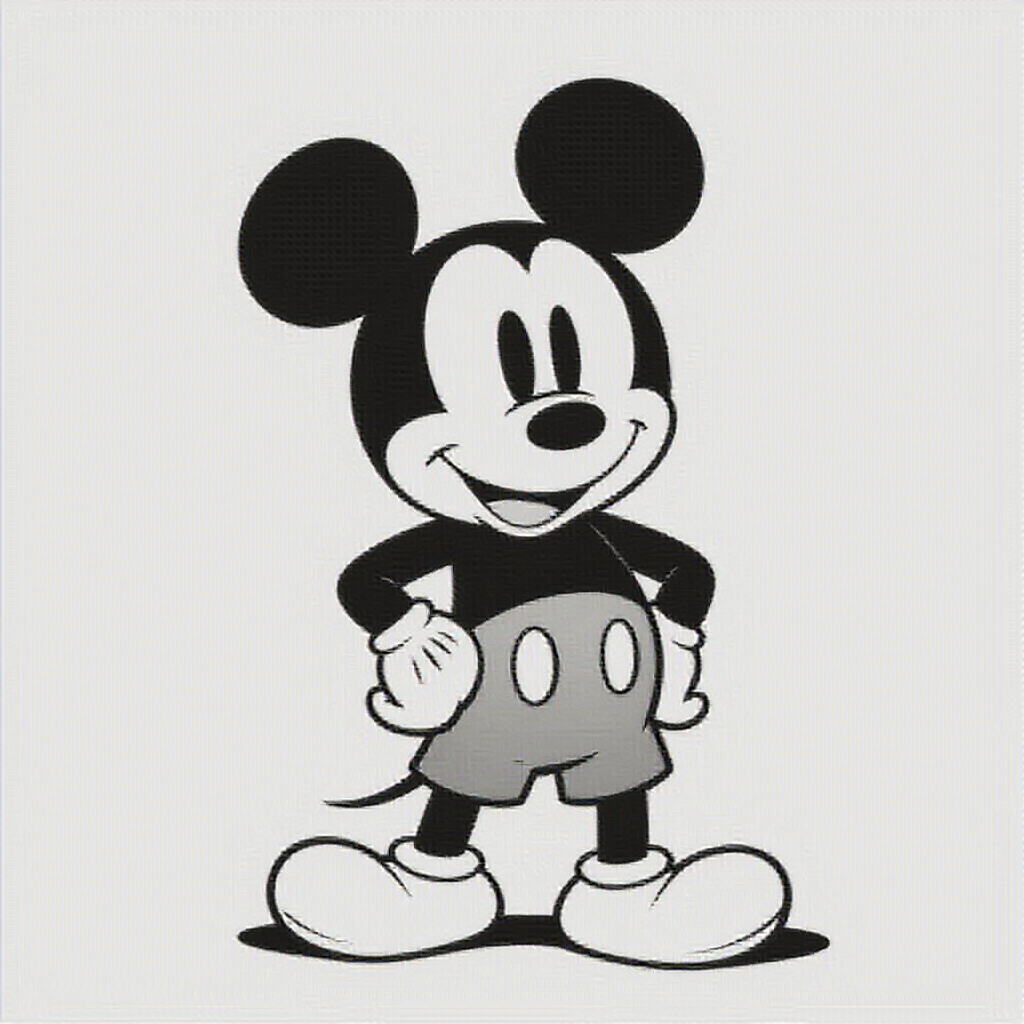} \\
    
    \bottomrule
  \end{tabular}}
  \caption{Visualized intermediate latents of Z-Image-Turbo, comparing states with and without the $x$-pred transformation across 9 generation timesteps.}
  \label{fig:ablation_booktabs}
\end{figure}

\begin{figure}[t]
  \centering
  \small
  \resizebox{0.9\textwidth}{!}{\begin{tabular}{c@{\hspace{0.4mm}}ccccccccccc}
    \toprule
    Timestep ($T$) & $1$ & $5$ & $10$ & $15$ & $20$ & $25$ & $30$ & $35$ & $40$ & $45$ & $50$ \\
    \midrule
    
    \begin{tabular}[c]{@{}c@{}}Vanilla \\ $D(\boldsymbol{z}_t)$\end{tabular} & 
    \includegraphics[width=.076\textwidth, valign=c]{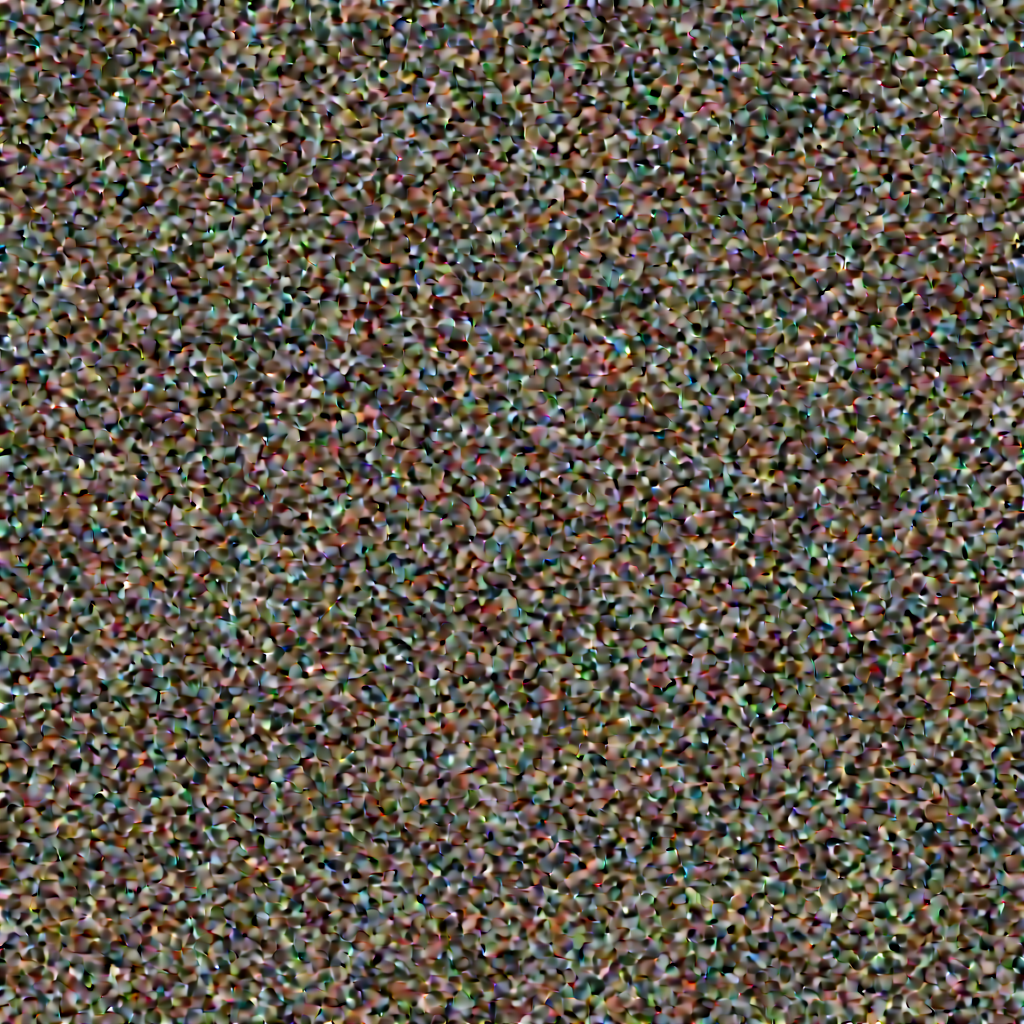} &
    \includegraphics[width=.076\textwidth, valign=c]{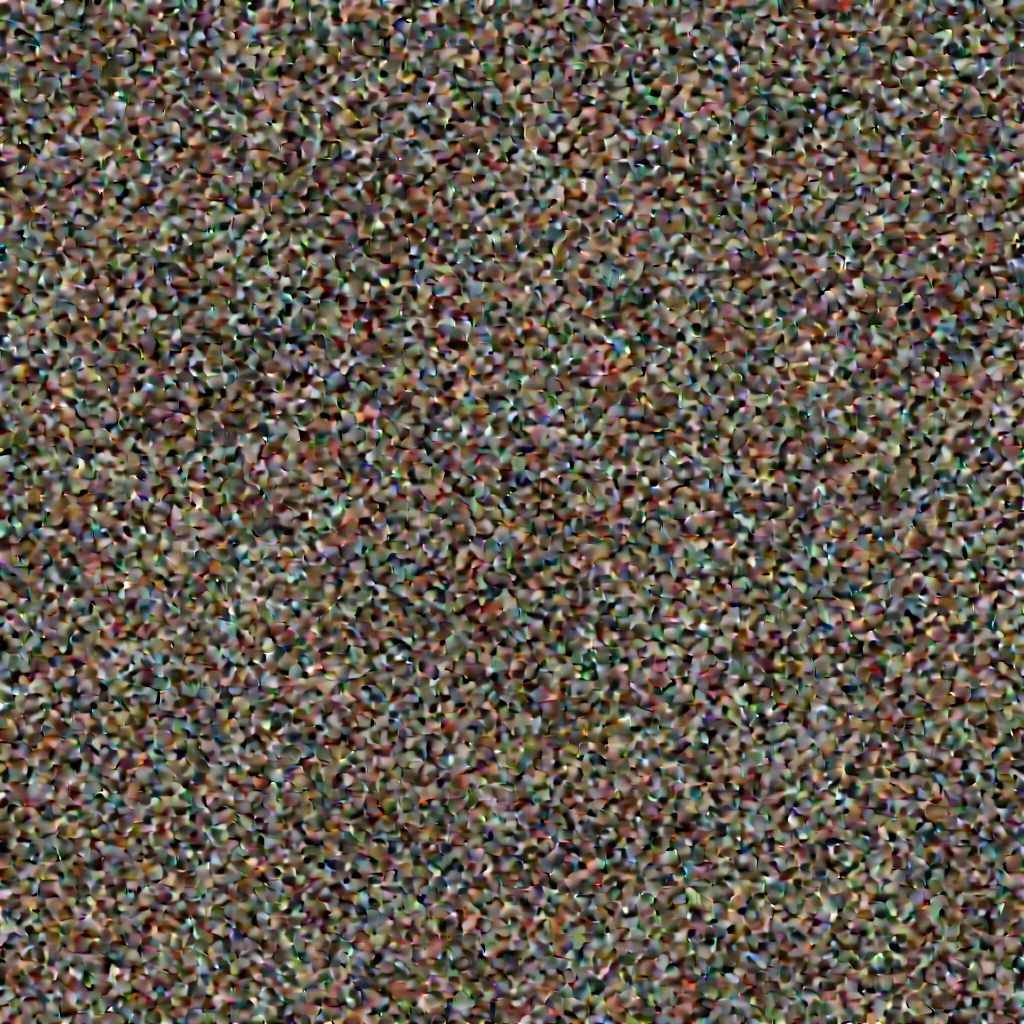} &
    \includegraphics[width=.076\textwidth, valign=c]{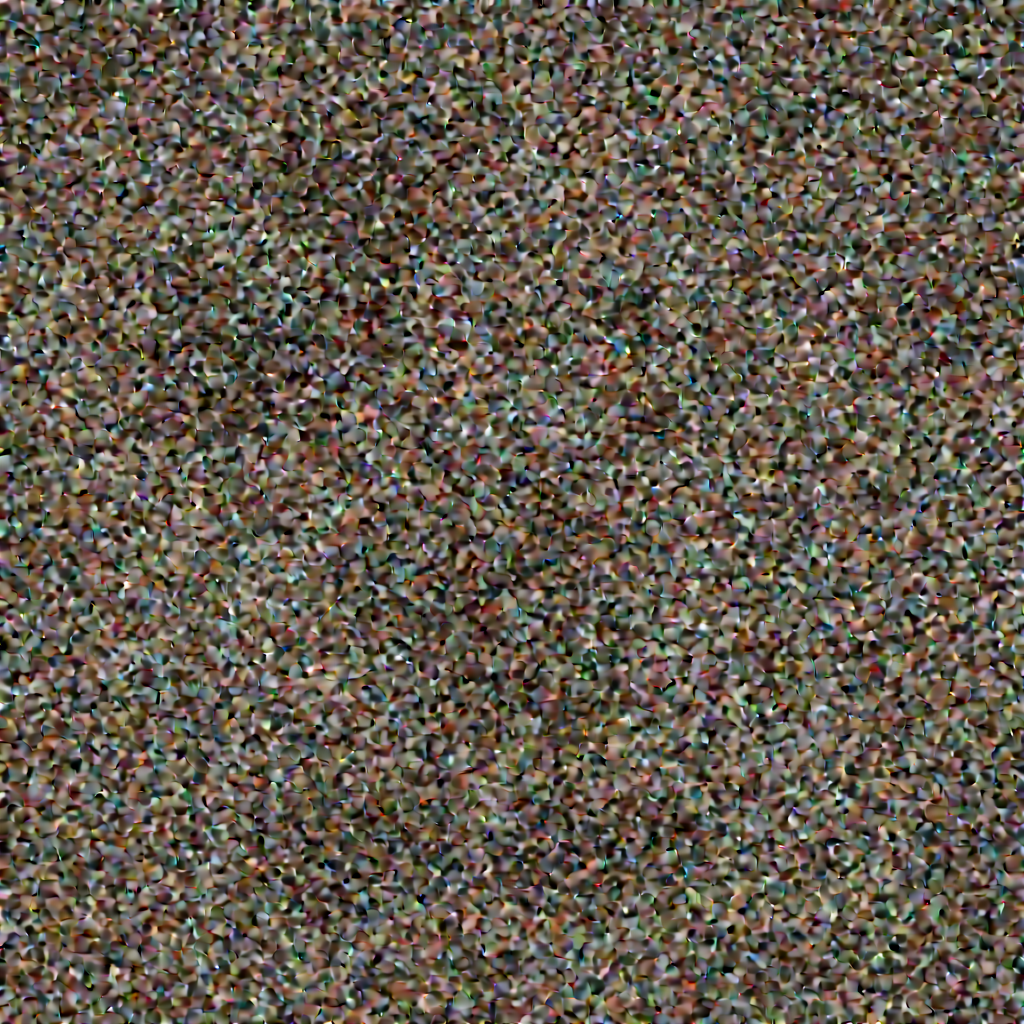} &
    \includegraphics[width=.076\textwidth, valign=c]{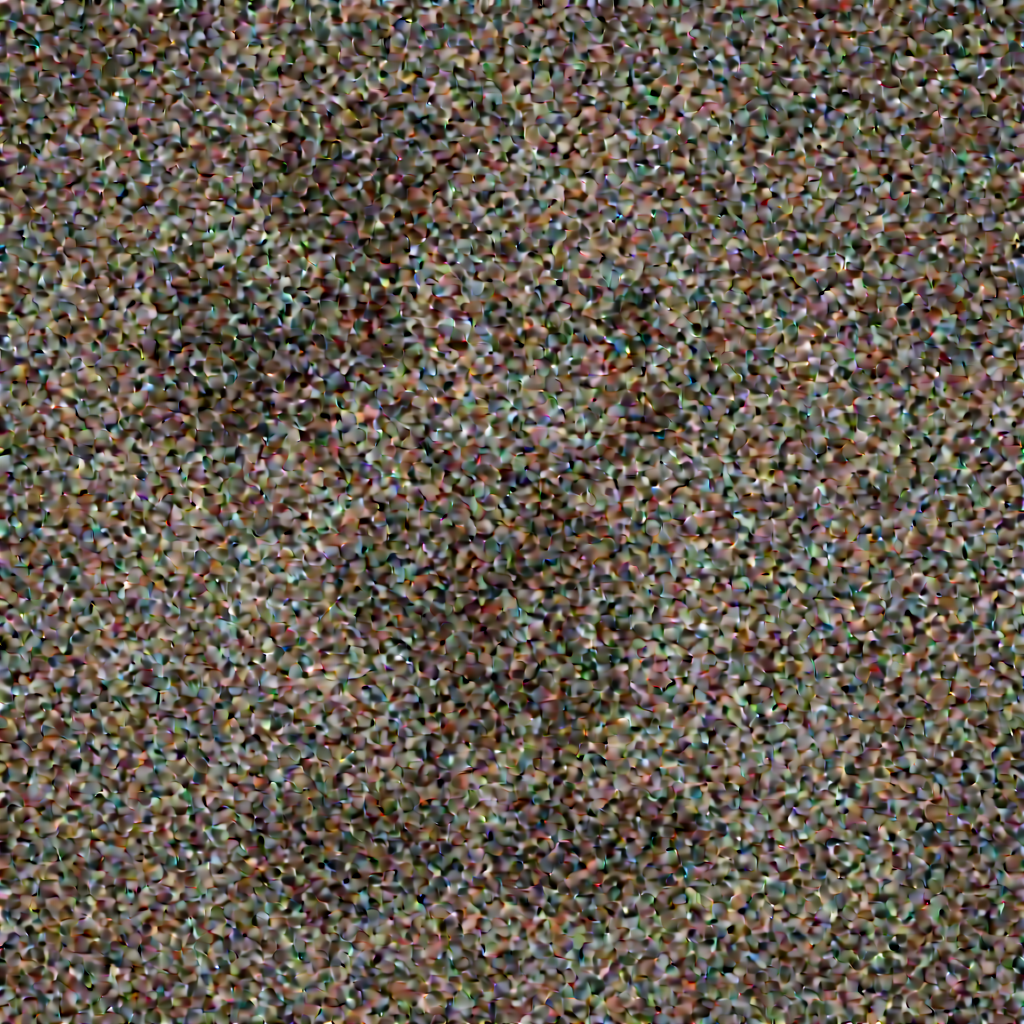} &
    \includegraphics[width=.076\textwidth, valign=c]{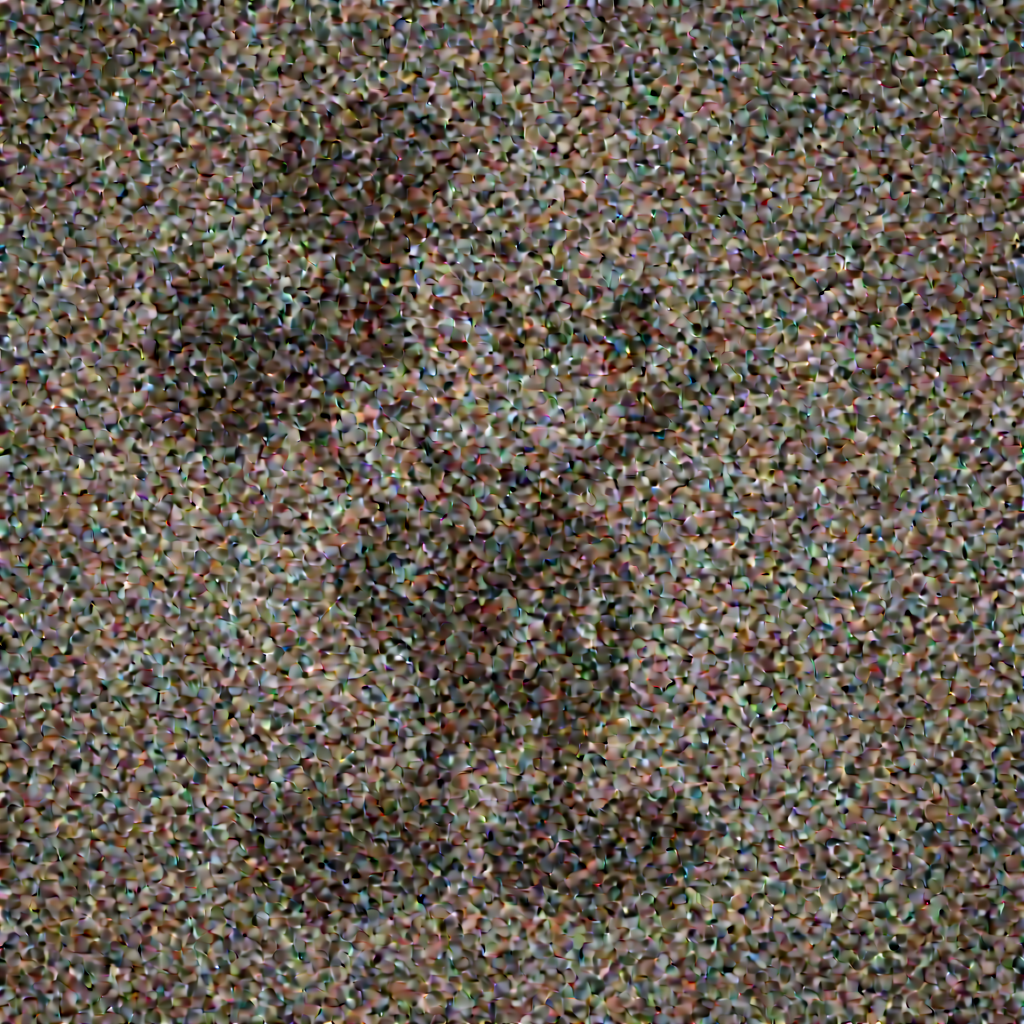} &
    \includegraphics[width=.076\textwidth, valign=c]{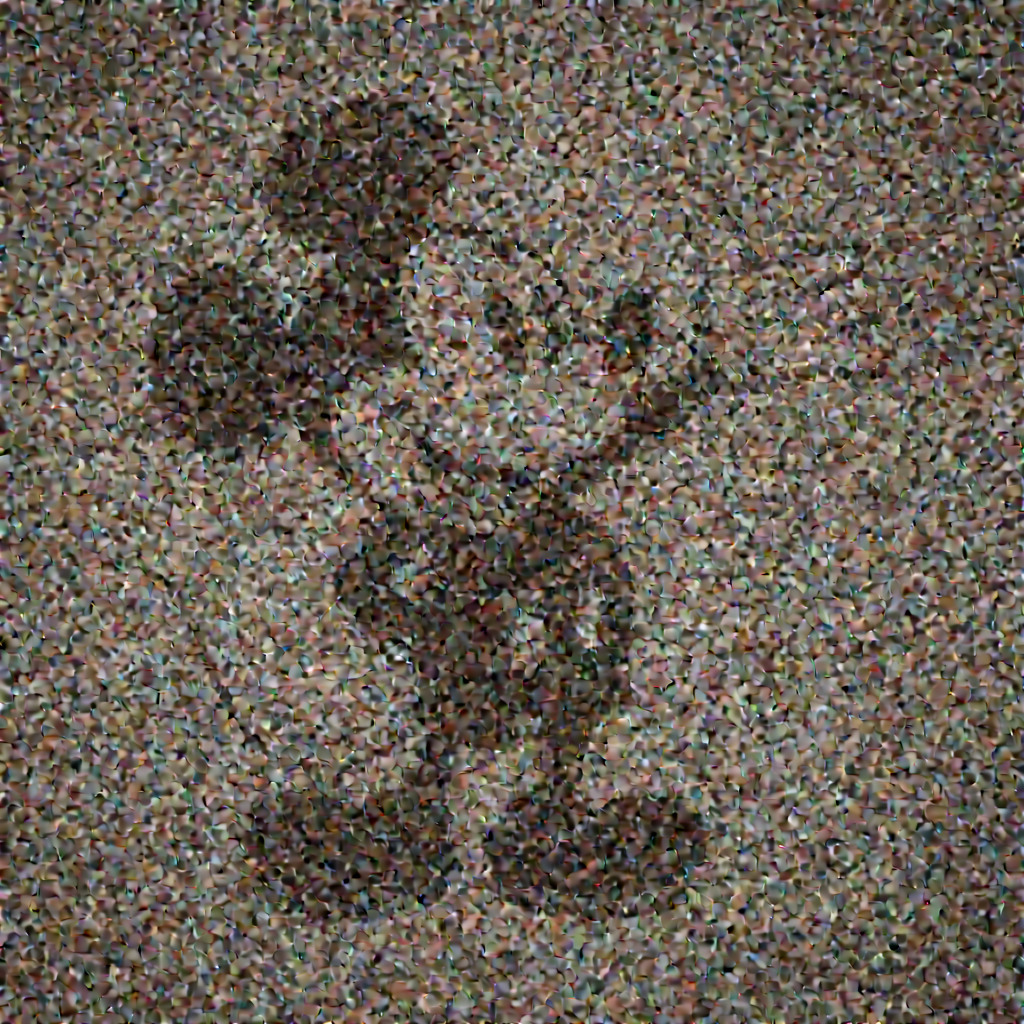} &
    \includegraphics[width=.076\textwidth, valign=c]{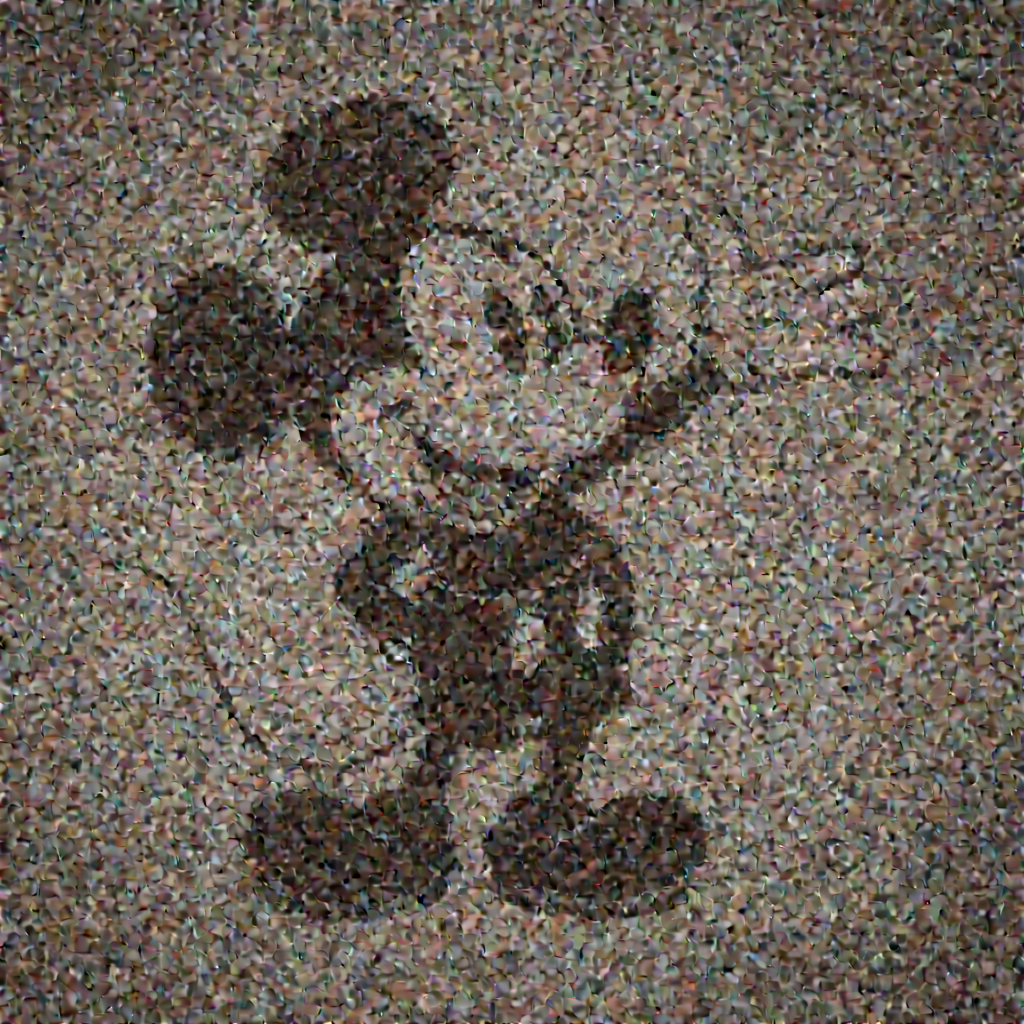} &
    \includegraphics[width=.076\textwidth, valign=c]{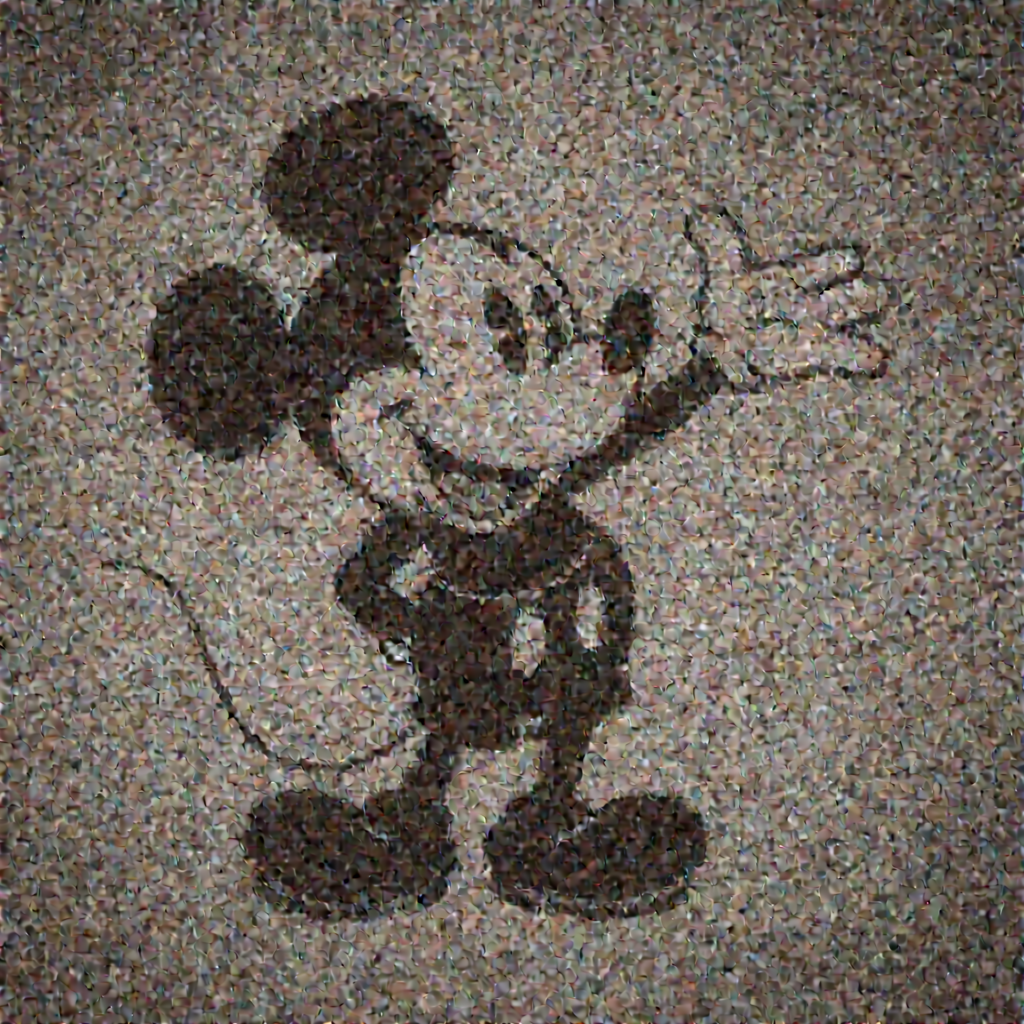} &
    \includegraphics[width=.076\textwidth, valign=c]{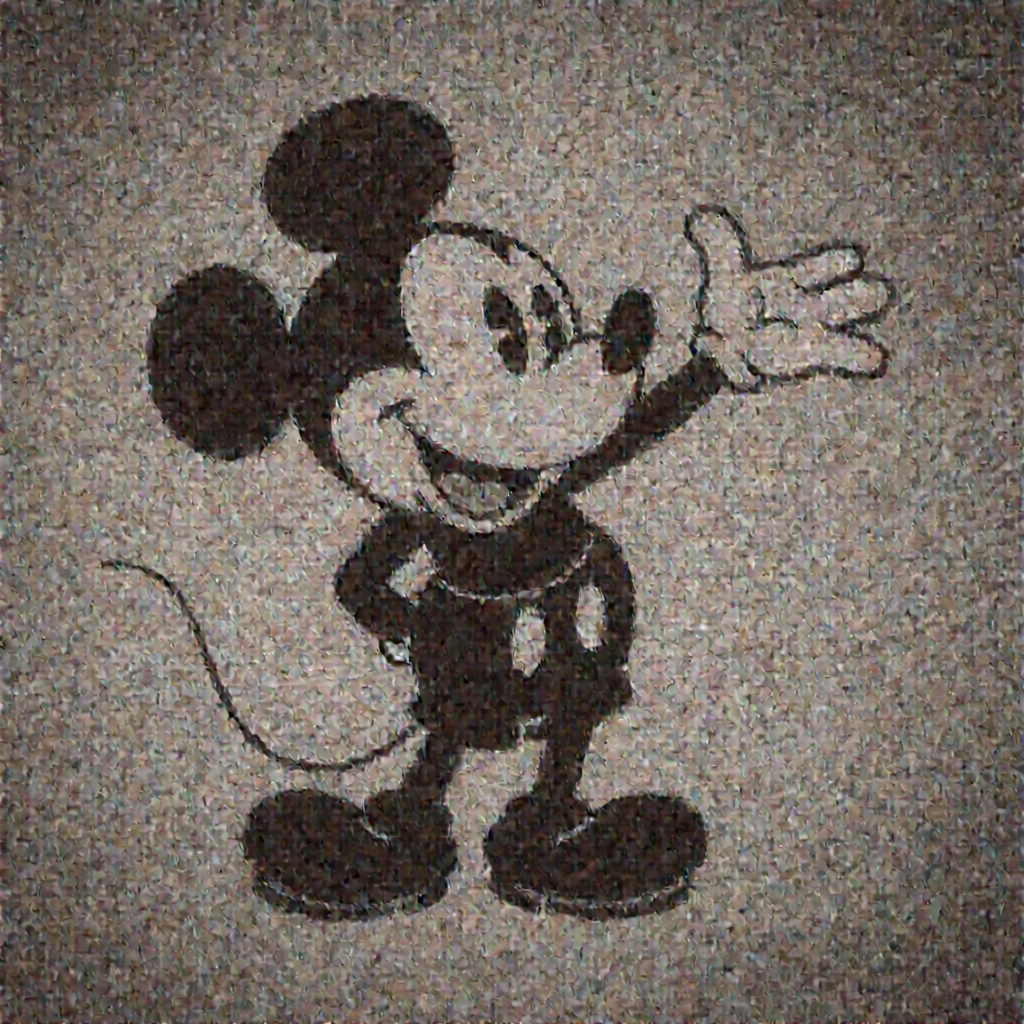} &
    \includegraphics[width=.076\textwidth, valign=c]{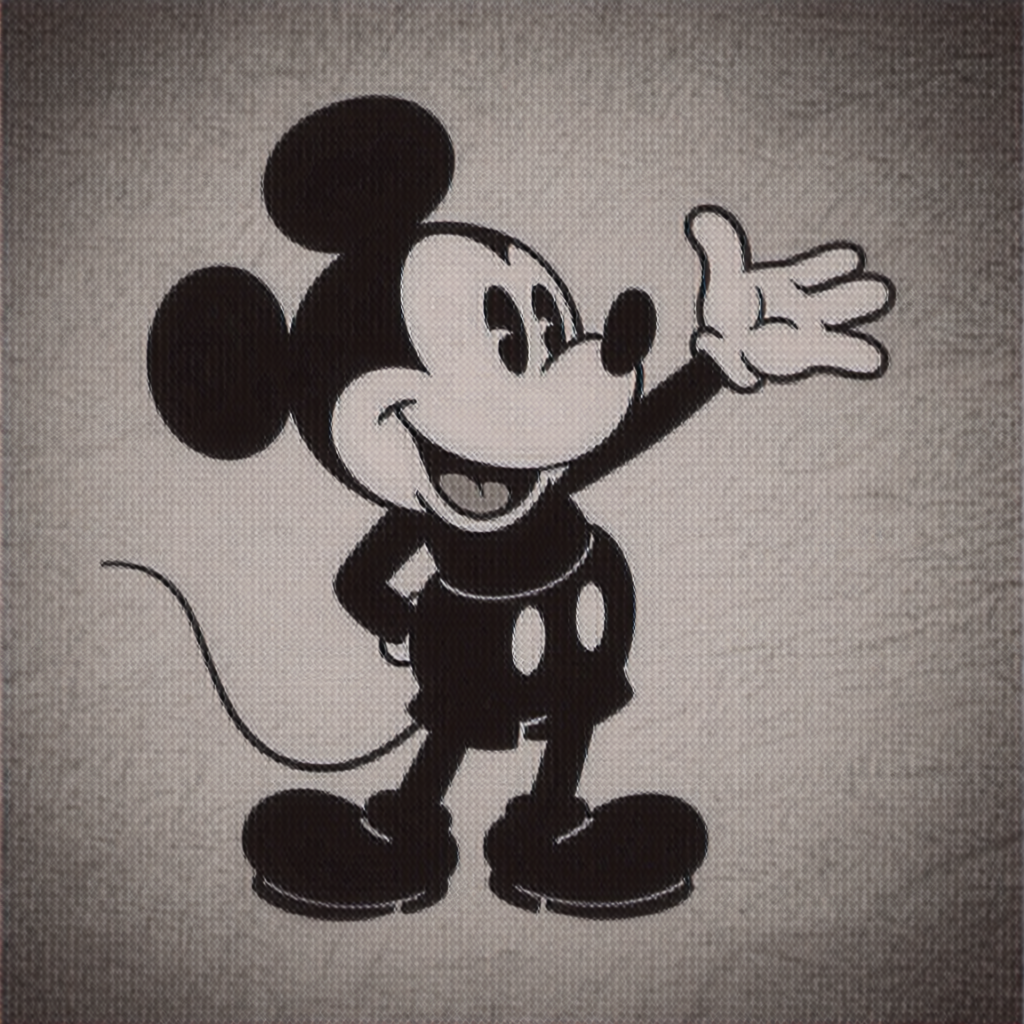} &
    \includegraphics[width=.076\textwidth, valign=c]{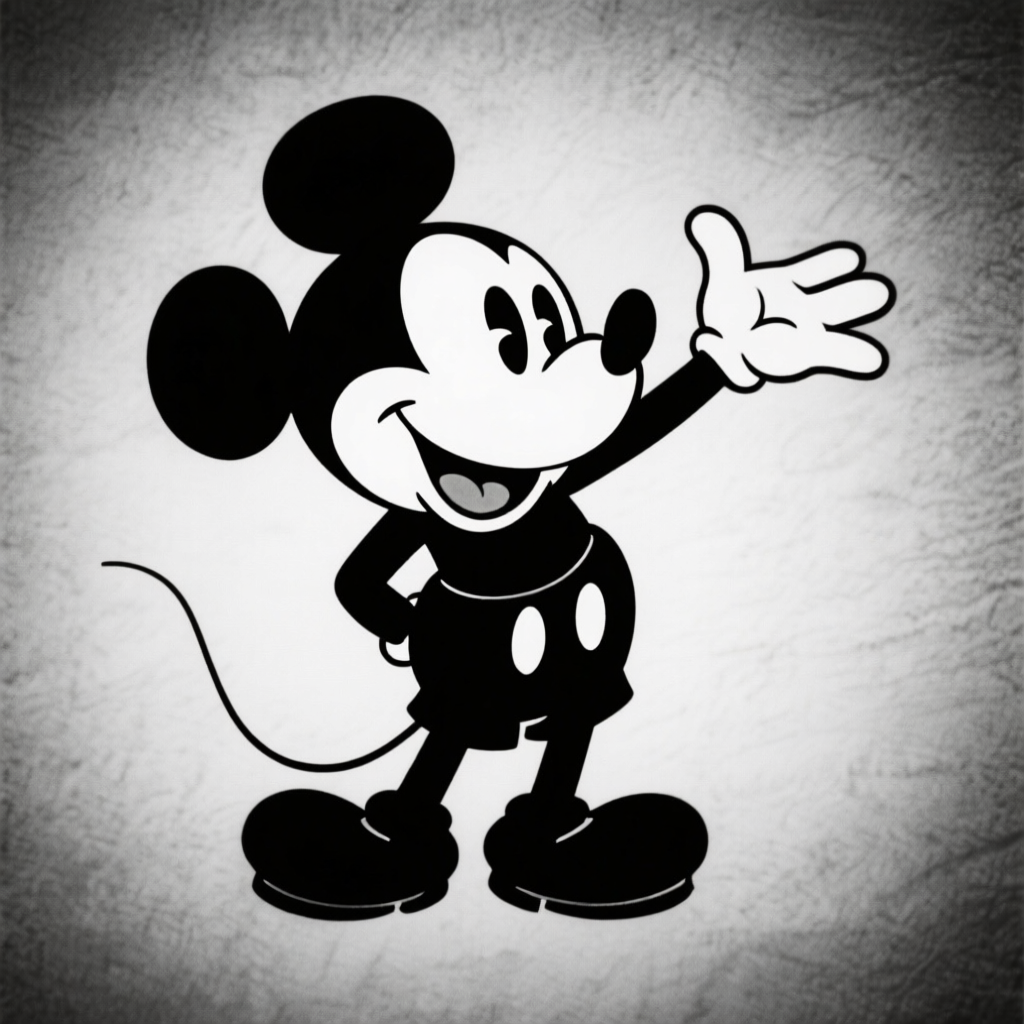} \\
    \addlinespace[1.5mm]
    
    \begin{tabular}[c]{@{}c@{}}$x$-pred \\ $D(\boldsymbol{x}_\theta(\boldsymbol{z}_t, t))$\end{tabular} & 
    \includegraphics[width=.076\textwidth, valign=c]{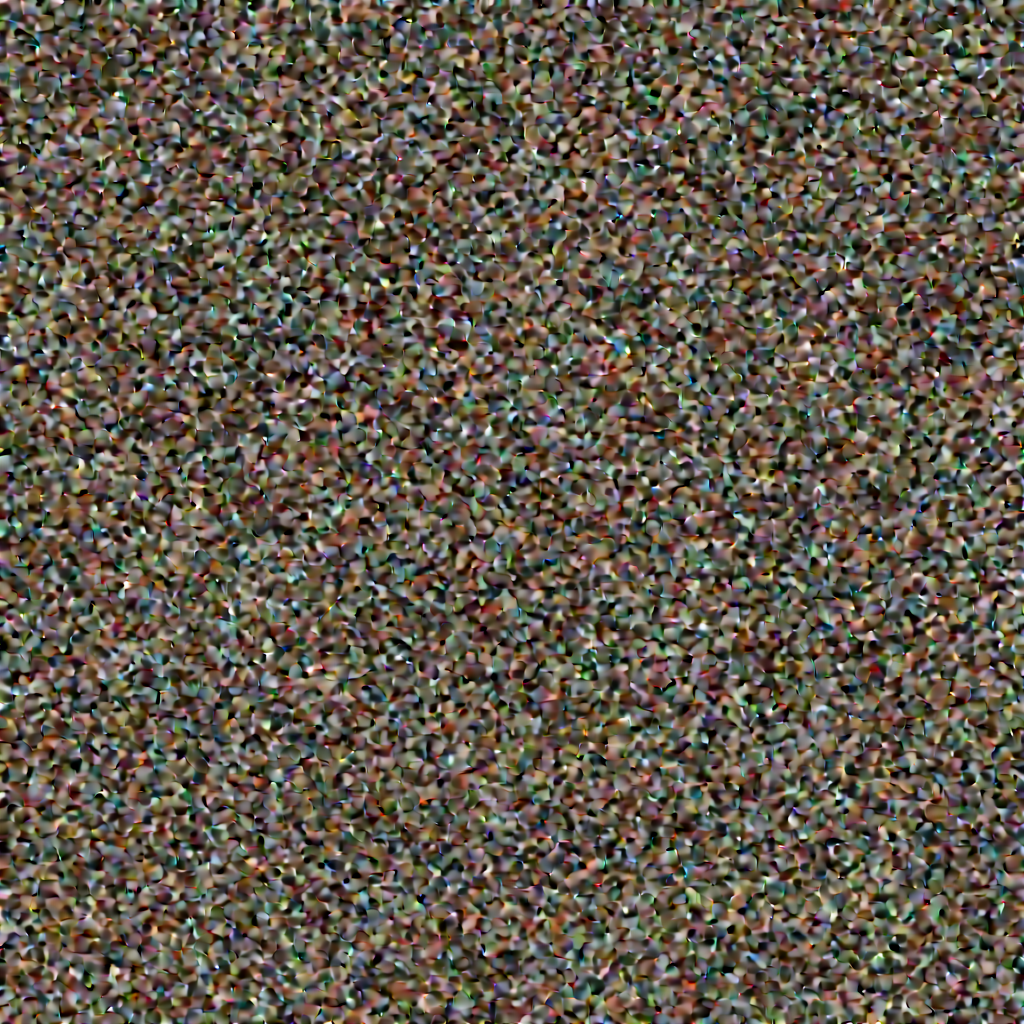} &
    \includegraphics[width=.076\textwidth, valign=c]{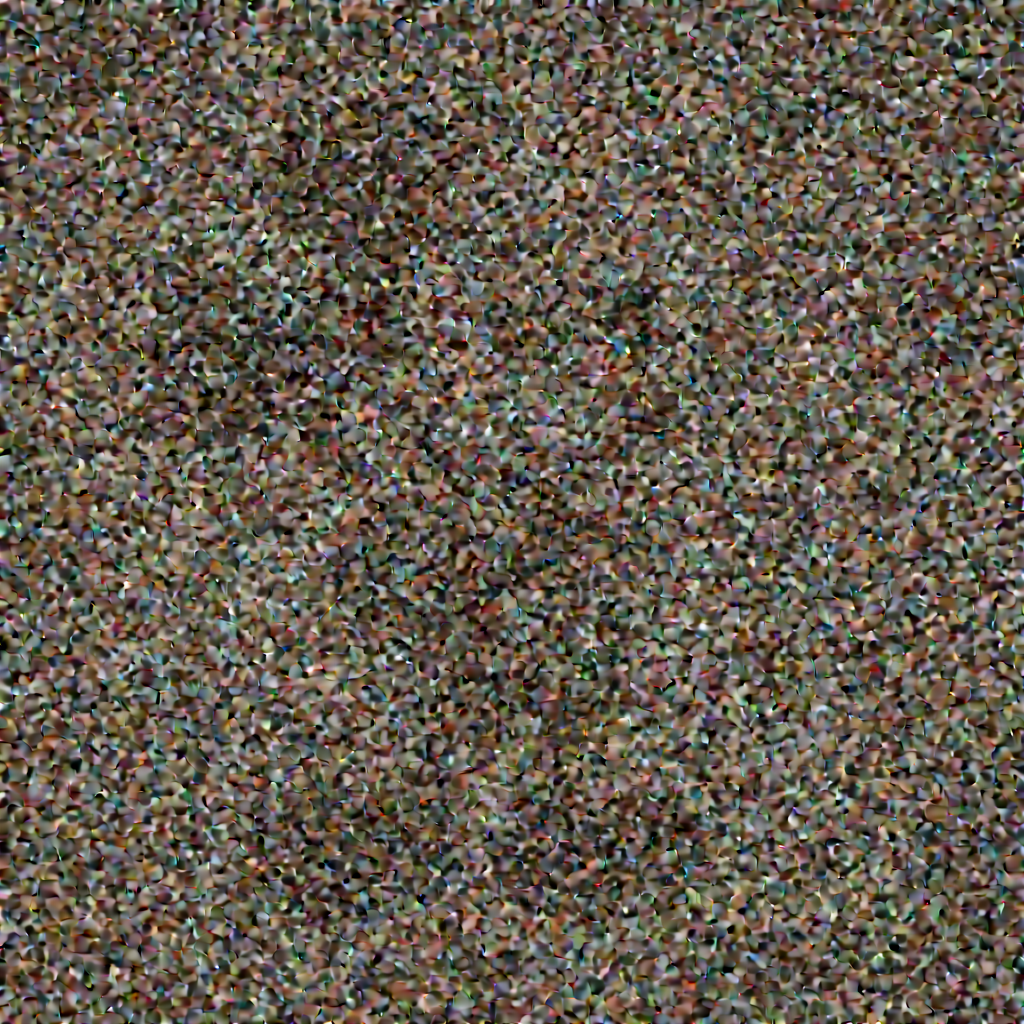} &
    \includegraphics[width=.076\textwidth, valign=c]{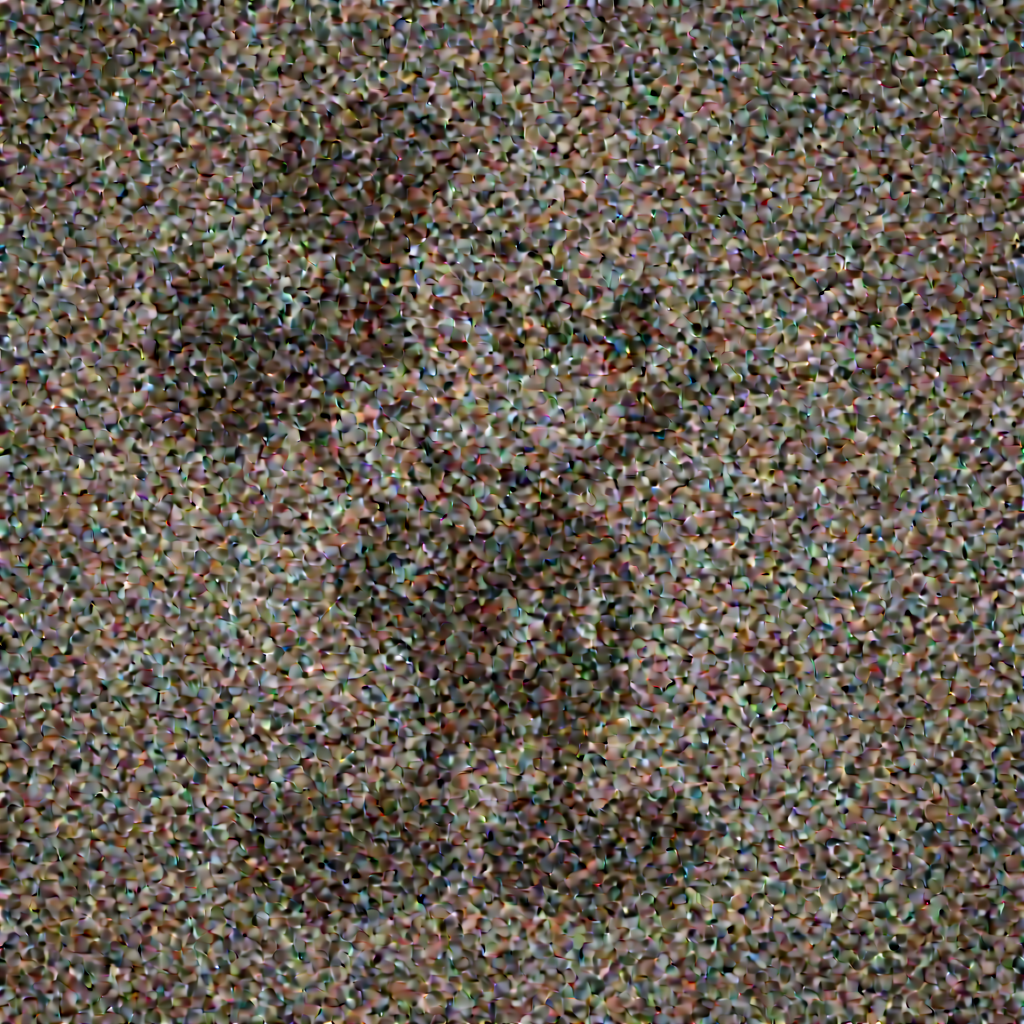} &
    \includegraphics[width=.076\textwidth, valign=c]{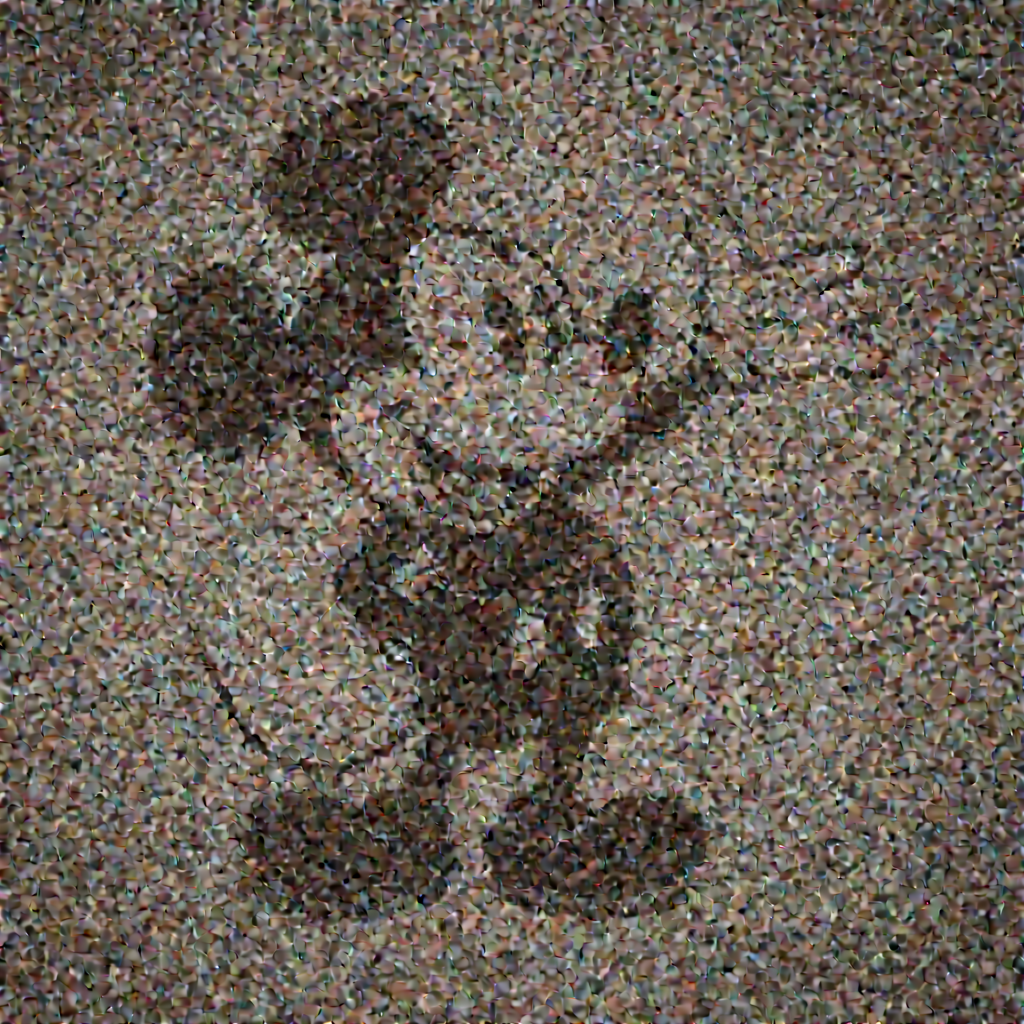} &
    \includegraphics[width=.076\textwidth, valign=c]{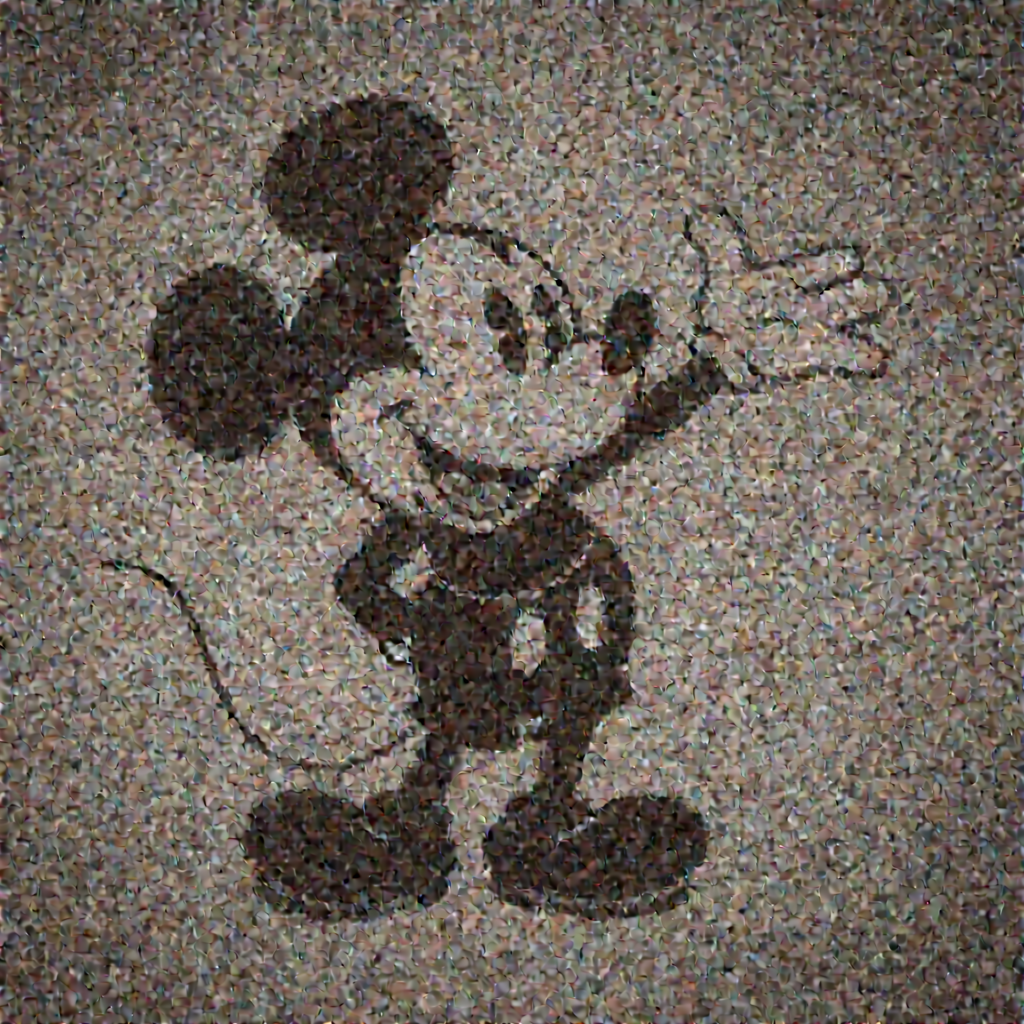} &
    \includegraphics[width=.076\textwidth, valign=c]{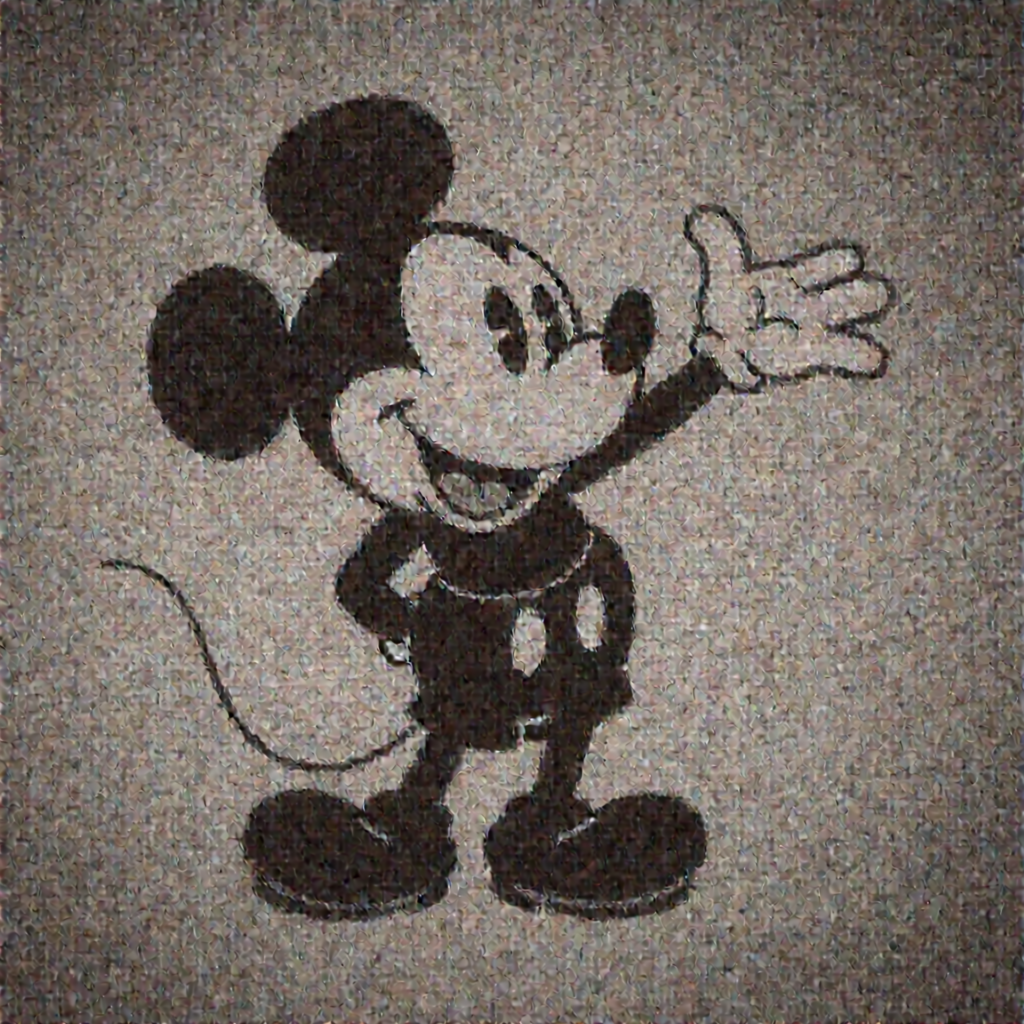} &
    \includegraphics[width=.076\textwidth, valign=c]{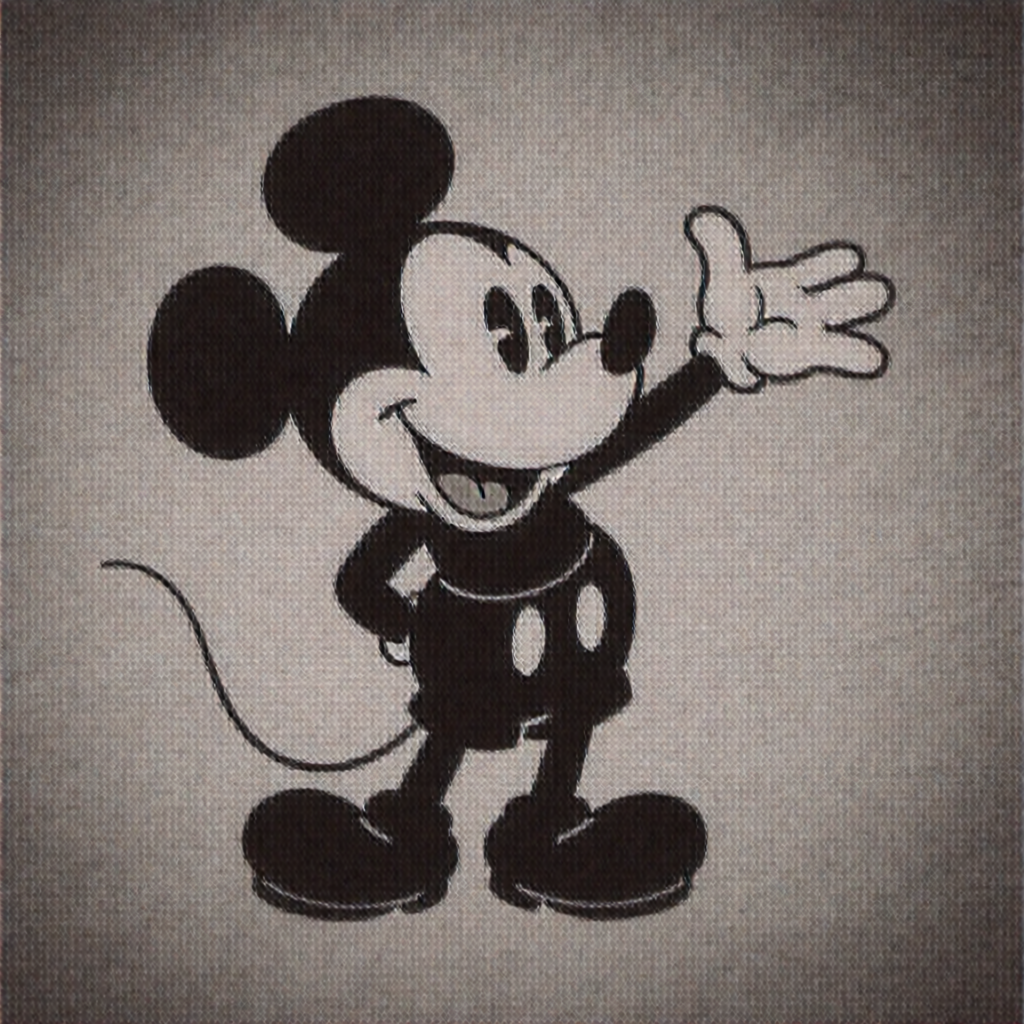} &
    \includegraphics[width=.076\textwidth, valign=c]{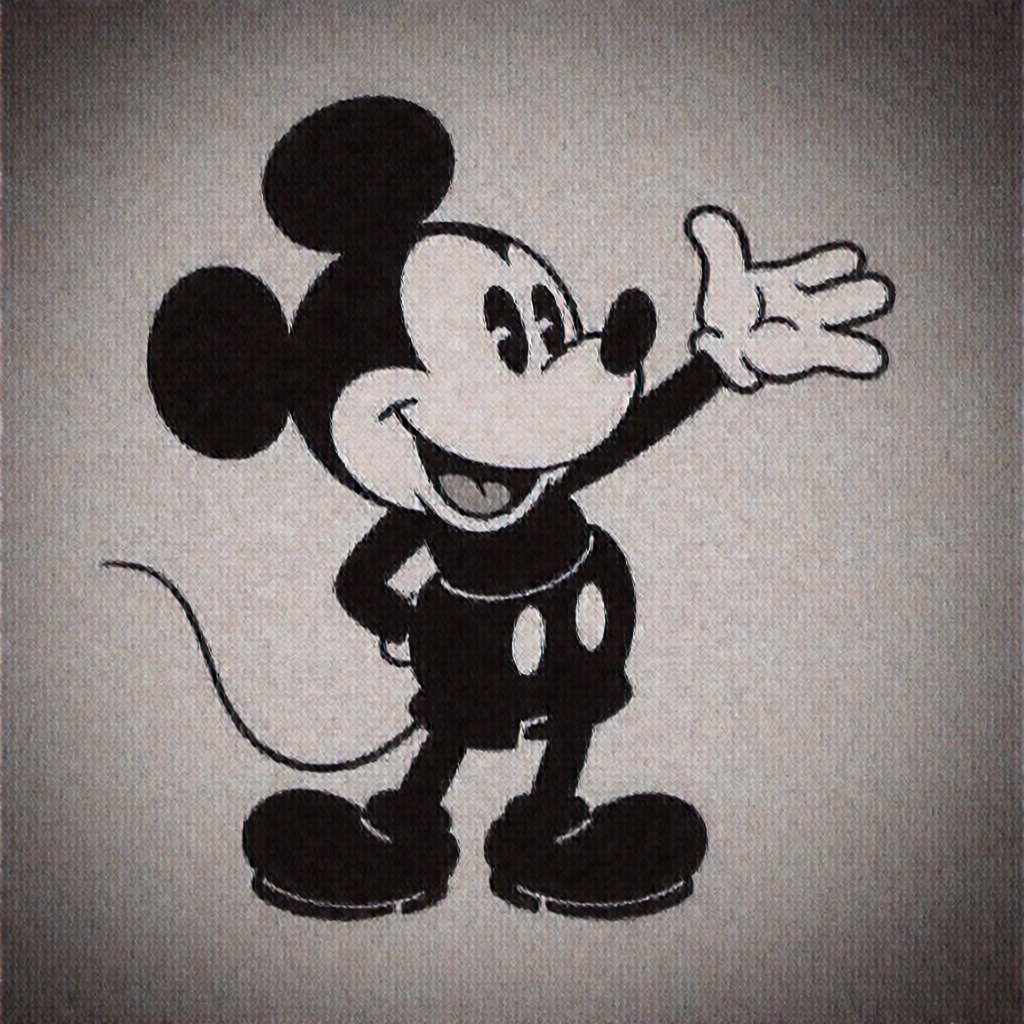} &
    \includegraphics[width=.076\textwidth, valign=c]{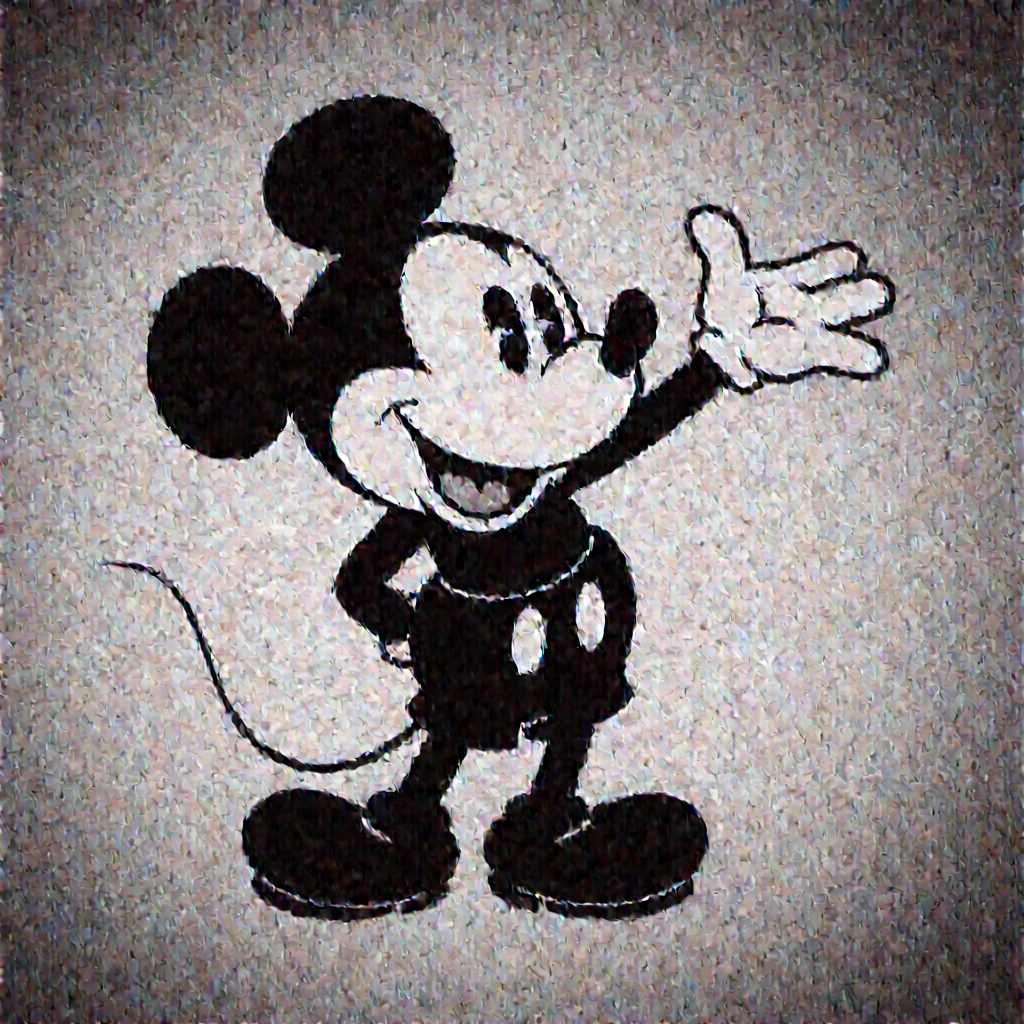} &
    \includegraphics[width=.076\textwidth, valign=c]{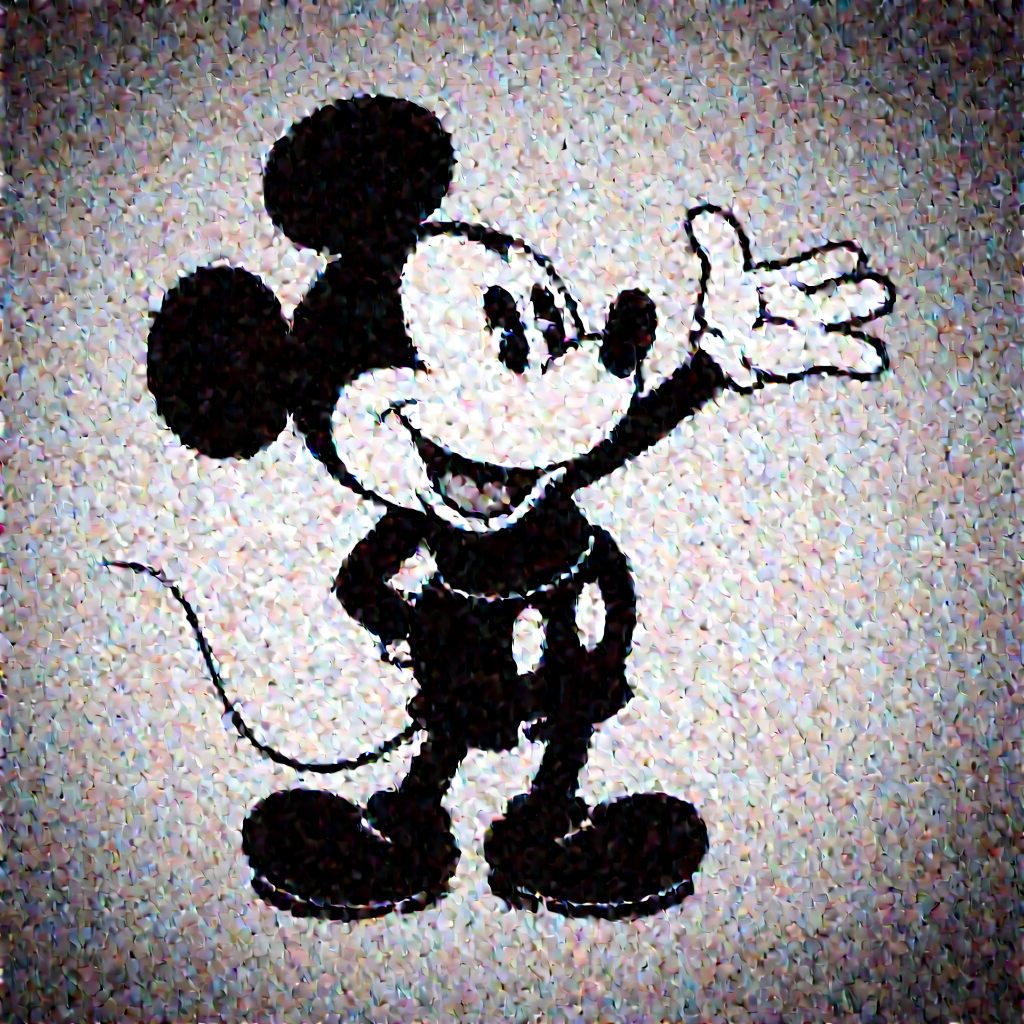} &
    \includegraphics[width=.076\textwidth, valign=c]{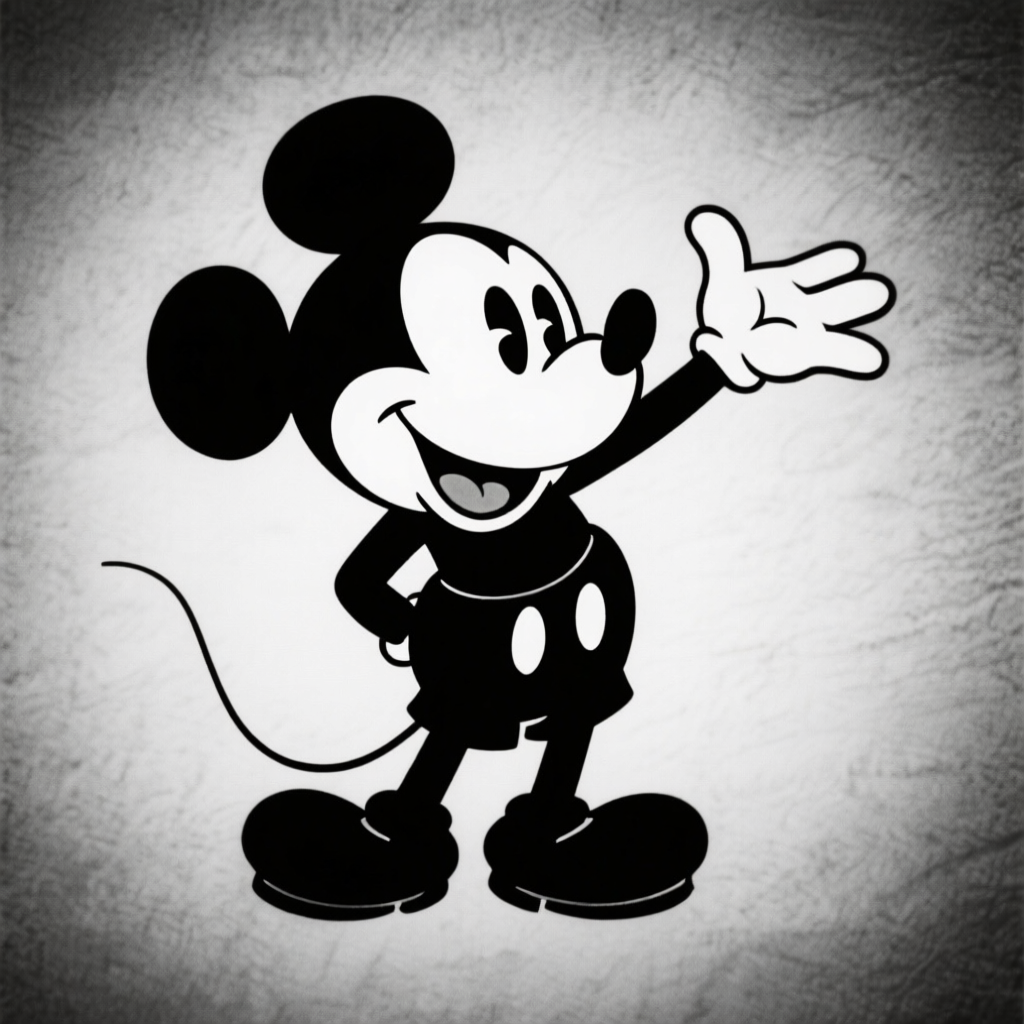} \\
    
    \bottomrule
  \end{tabular}}
  \caption{Visualized intermediate latents of Qwen-Image, comparing states with and without the $x$-pred transformation across 50 generation timesteps.}
  \label{fig:qwen_image_ablation}
\end{figure}

Fig.~\ref{fig:ablation_booktabs} demonstrates that, for Z-Image-Turbo, the $x$-pred transformation enables the retrieval of visual content highly consistent with the final samples even during the early stages of inference.
While direct visualizations of the latent states at each timestep are dominated by noise from $T=1$ to approximately $T=5$, applying the $x$-pred transformation yields results that align with the final $T=8$ state much earlier in the process.
Conversely, as shown in Fig.~\ref{fig:qwen_image_ablation}, this transformation in Qwen-Image is less effective during initial denoising; meaningful enhancement only becomes apparent between steps $T=20$ and $T=30$, rather than within the $T=1$ to $T=10$ range.
Furthermore, a slight degradation in quality is observed in the final steps of the $x$-pred transformation.

\noindent\textbf{Encoder-wise Analysis.}
Table~\ref{tbl:category-wise} indicates that the performance fluctuates depending on the choice of the image encoder $E$, suggesting that our method has the potential for further performance improvements by switching to better encoders.
Specifically, while substituting the standard CLIP encoder with SigLIP or SigLIP2 yields comparable or slightly varied results, employing the more advanced Qwen3VL-embedding produces a performance leap.

\noindent\textbf{Category-wise Analysis.}
Table~\ref{tbl:category-wise} also demonstrates that EDGE-Shield excels at classifying violation content on IPs and individual faces. 
Specifically, the ROC-AUC scores for the IP and Individual categories are consistently high, whereas the classification of artistic styles (Style) remains more challenging for the current framework.
The weakness on the Style category is attributed to the reference style is differ from the generated ones, whose qualitative analysis is detailed in the Appendix.
\begin{table}[t]
\centering
\setlength{\tabcolsep}{7pt}
\caption{ROC-AUC at timestep 0 on Z-Image across image encoders inclding CLIP, SigLIP, SigLIP2, Qwen3-VLM-Embedding.}\label{tbl:category-wise}
\begin{tabular}{@{}lcccccc@{}}
\toprule 
  & \multicolumn{2}{c}{CPDM} & \multicolumn{3}{c}{HUB} &  \\ \cmidrule(lr){2-3} \cmidrule(lr){4-6} 
EDGE-Shield              & IP     & Individual & IP     & Individual & Style   \\ \midrule
\hspace{1em}w/ CLIP              & 0.903 & 0.970 & 0.767 & \textbf{0.854} & 0.616  \\
\hspace{1em}w/ SigLIP            & 0.943 & 0.965 & 0.804 & 0.822 & 0.635  \\
\hspace{1em}w/ SigLIP2           & 0.942 & 0.957 & 0.813 & 0.799 & 0.629  \\
\hspace{1em}w/ Qwen3VL-Embedding & \textbf{0.988} & \textbf{0.988} & \textbf{0.827} & 0.845 & \textbf{0.645}  \\ \bottomrule
\end{tabular}
\end{table}

\begin{figure}[h]
  \centering
  \begin{subfigure}{0.48\linewidth}
    \centering
    \includegraphics[width=\linewidth]{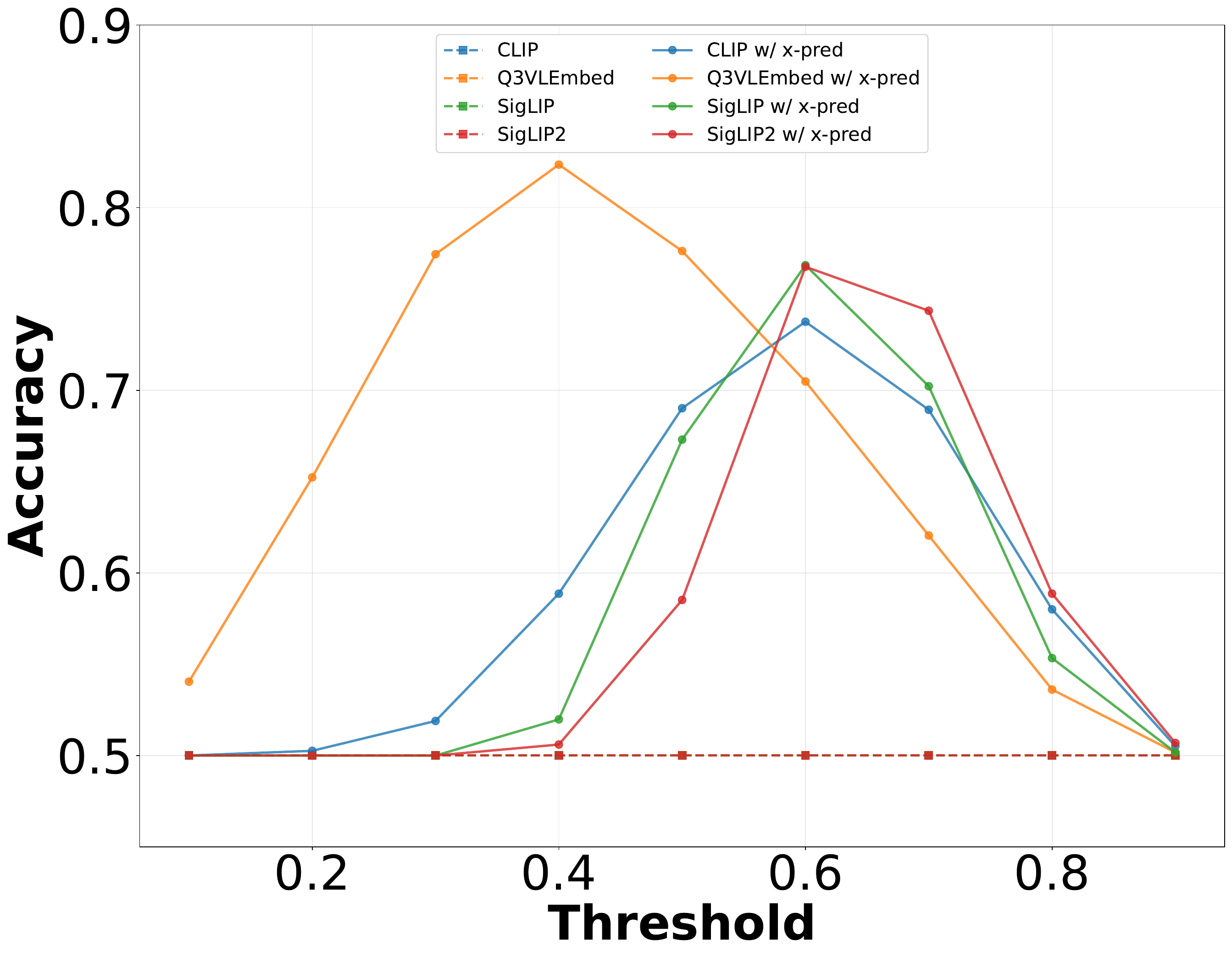}
    \caption{Z-Image-Turbo}
    \label{fig:ablation-qwen}
  \end{subfigure}
  \begin{subfigure}{0.48\linewidth}
    \centering
    \includegraphics[width=\linewidth]{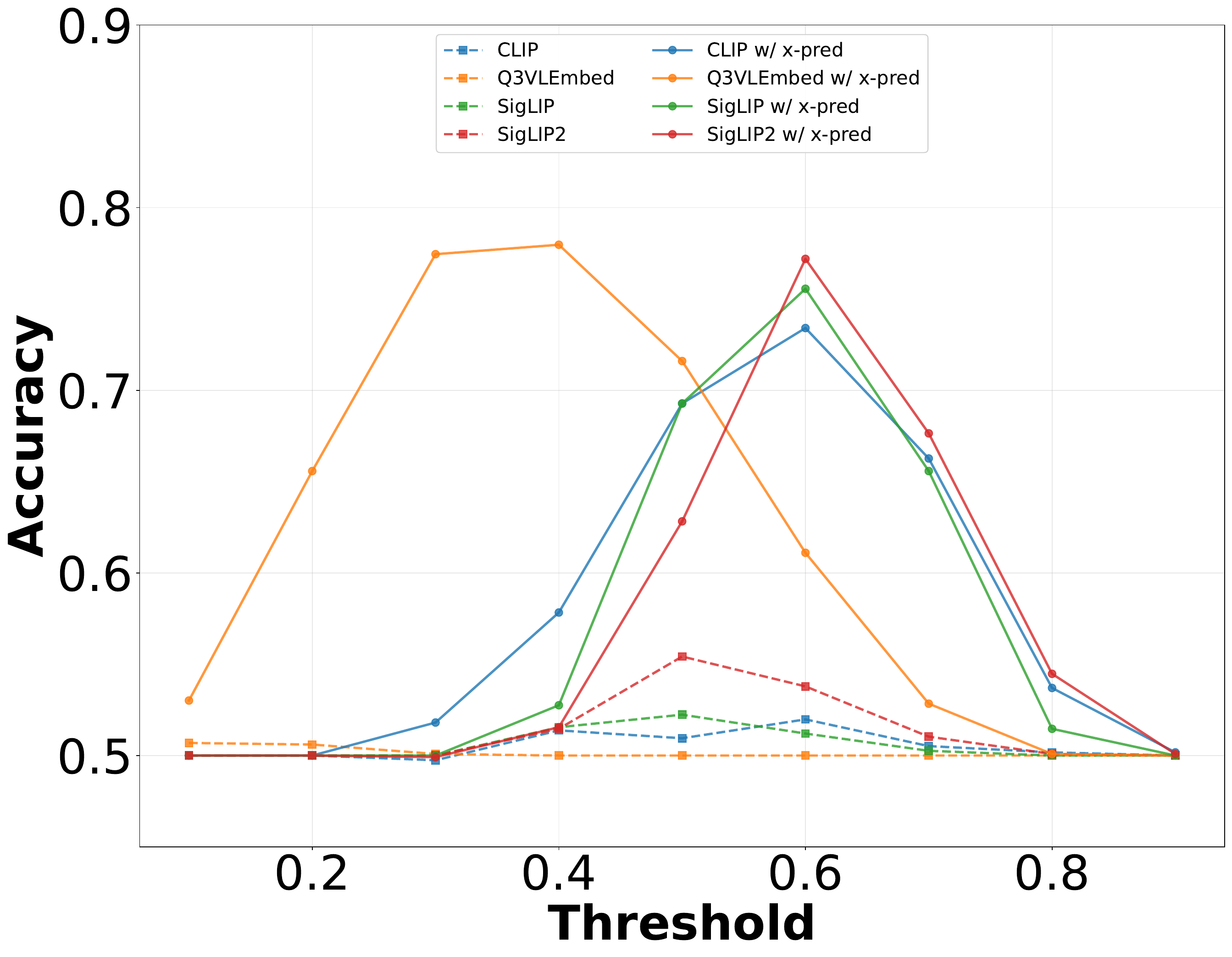}
    \caption{Qwen-Image} %
    \label{fig:ablation-third}
  \end{subfigure}
  \caption{Accuracy across thresholds from 0.1 to 0.9. (a) Results on Z-Image-Turbo at timestep 1 (b) Results on Qwen-Image at timestep 25.}\label{fig:ablation_accuracy}
\end{figure}

\noindent\textbf{Analysis on the optimal threshold.}
Fig.~\ref{fig:ablation_accuracy} shows the variation in accuracy of EDGE-Shield across different thresholds, indicating that an optimal threshold exists for each embedding model.
For CLIP, SigLIP, and SigLIP2, the threshold around 0.7 is the best for accuracy.
Conversely, threshold around 0.4 is the best for Qwen3VL-Embedding.

\noindent\textbf{Robustness against Roundabout Prompts.}
Table~\ref{tbl:input-filter-weekness} demonstrates that violative content can still be elicited through implicit or roundabout prompts that can bypass standard text-based filters.
While input-based guardrails typically rely on explicit keywords or semantic patterns in the prompt, output-based approaches operate on the actual visual synthesis process. This allows for a robust defense by detecting violative characteristics in the generated content, regardless of the prompt's linguistic framing.

\begin{table}[t]
    \centering
    \setlength{\tabcolsep}{8pt}      %
    \caption{Comparison of Straight and Roundabout prompting methods. Both images were generated by Qwen-Image.}\label{tbl:input-filter-weekness}    
    \begin{tabular}{m{3.5cm} c m{3.5cm} c}
        \toprule
        \multicolumn{2}{c}{\textbf{Straight}} & \multicolumn{2}{c}{\textbf{Roundabout}} \\
        \cmidrule(lr){1-2} \cmidrule(lr){3-4} %
        Prompt & Image & Prompt & Image \\
        \midrule
        Mickey mouse, old, black and white. & 
        \includegraphics[width=1.2cm, valign=m]{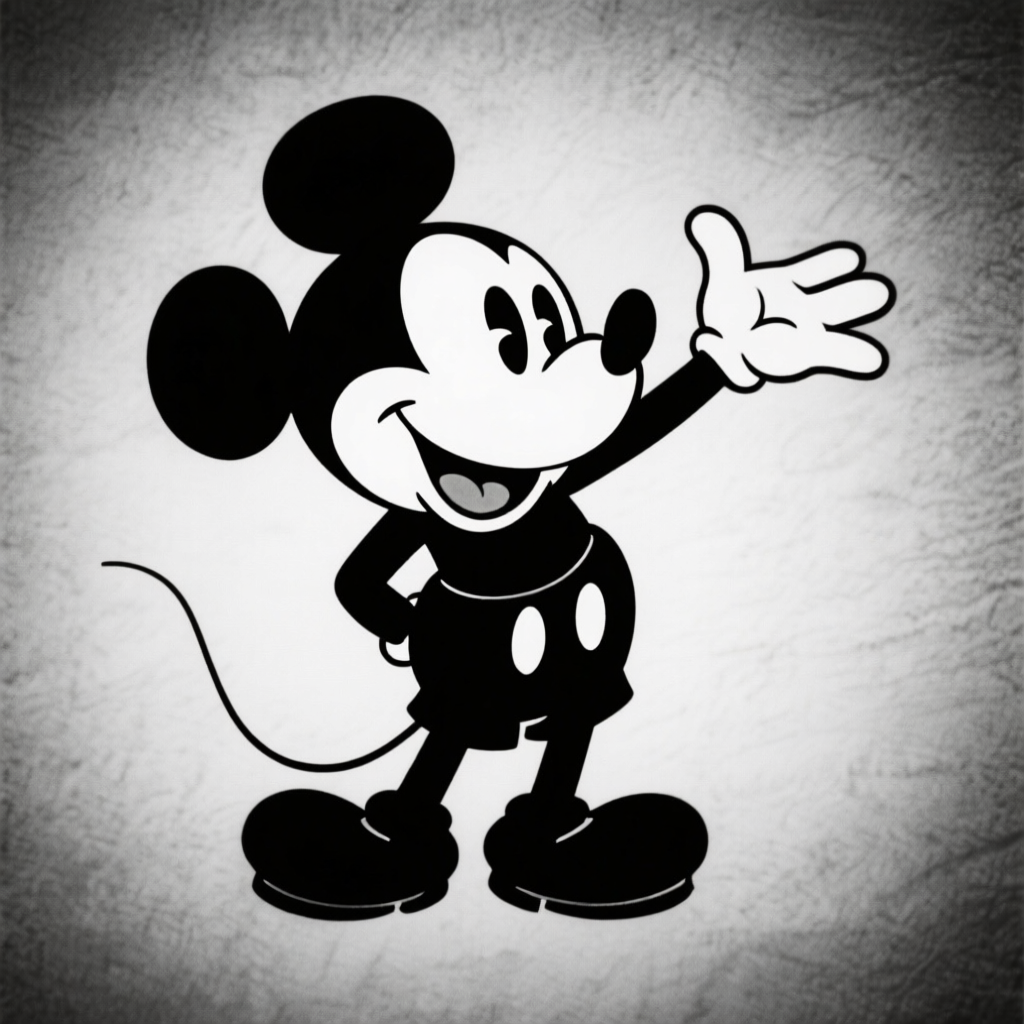} & 
        \color{red}Mouse famous cartoon character\color{black}, old, black and white. & 
        \includegraphics[width=1.2cm, valign=m]{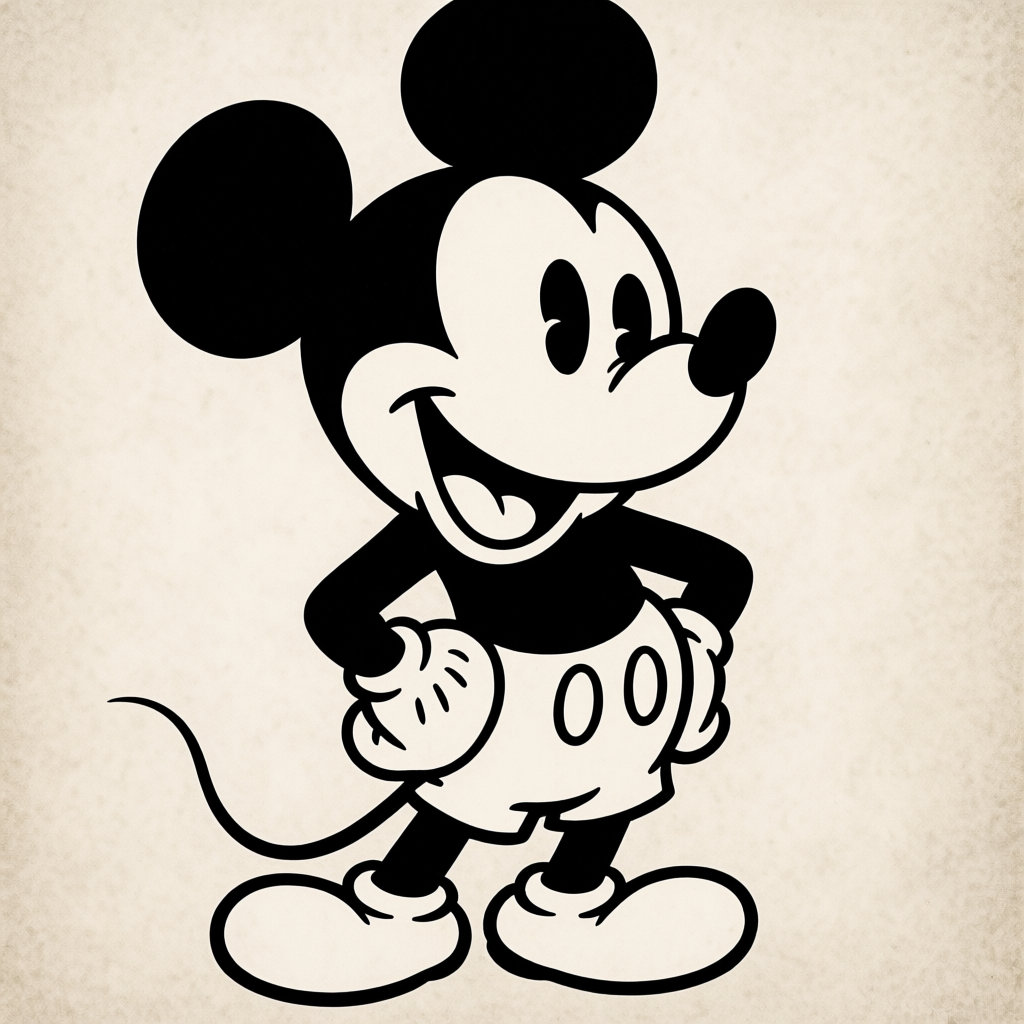} \\ 
        \bottomrule
    \end{tabular}
\end{table}

\noindent\textbf{Supplementary Empirical Evaluations.}
To further validate the extensibility and performance of EDGE-Shield, we provide additional evaluations in the Appendix, including:
(\textit{\textbf{i}}) the superior performance of EDGE-Shield compared to the the proprietary model, 
(\textit{\textbf{ii}}) the effectiveness of our methods to noise-based generative models shown by the qualitative experiments, and 
(\textit{\textbf{iii}}) robustness of our method compared to input-based content filters shown by the empirical results.

\section{Conclusion}
We propose EDGE-Shield, a reference-based content filter that scales efficiently with the number of references and detects violative content at the early stages of the denoising process. 
To achieve this, EDGE-Shield incorporates two improvements: 
(\textit{\textbf{i}}) it gains scalability by using an embedding model that allows pre-computing and caching reference embeddings, enabling efficient batch similarity calculations, and 
(\textit{\textbf{ii}}) it facilitates early detection by transforming intermediate latent into the estimated final clean images. 
The limitation of this study is the marked decrease in classification performance regarding the stylistic mimicry.
The decline in performance is tied to the inherent nature of cosine similarity, which struggles to differentiate styles whose feature vectors are less distinct than those of specific faces or IPs.
For future work, it would be promising to further refine the architecture to maintain performance when the reference set expands to tens of thousands of entries.

\bibliographystyle{splncs04}
\bibliography{main}

\appendix
\section{Additional Results}

\subsection{Application to noise-based generative models}

\noindent\textbf{Noise-based Models.}
In accordance with Eq. 1 in the main draft, noise-based models employ a linear noise schedule defined by $\alpha_t = t$ and $\sigma_t = 1-t$. This formulation yields the intermediate state $\boldsymbol{z}_t = t\boldsymbol{x} + (1-t)\boldsymbol{\epsilon}$. 
The noise-based model $\boldsymbol{\epsilon}_\theta(\boldsymbol{z}_t, t)$, parameterized by $\theta$, is optimized to predict the noise component $\boldsymbol{\epsilon}$ by minimizing the loss $\mathcal{L} = \mathbb{E}_{t, \boldsymbol{x}, \boldsymbol{\epsilon}} \| \boldsymbol{\epsilon}_\theta(\boldsymbol{z}_t, t) - \boldsymbol{\epsilon} \|^2$.

\noindent\textbf{Pseudo-Clean Sample Estimation in noise-based models.}
We can also obtain a pseudo-clean data sample in noise-based models with an interpolation.
Using Eq. 1 under linear scheduling, the estimate $\boldsymbol{x}_\theta(\boldsymbol{z}_t, t)$ can be derived using the predicted noise:
\begin{equation*}
    \boldsymbol{x}_\theta(\boldsymbol{z}_t, t) = \frac{ \boldsymbol{z}_t - (1 - t) \boldsymbol{\epsilon}_\theta(\boldsymbol{z}_t, t)}{t}.
\end{equation*}
Theoretically, the $x$-pred transformation also can be applied to the noise-based model.

\begin{figure}[t]
  \centering
  \small %
  \resizebox{0.9\textwidth}{!}{\begin{tabular}{c@{\hspace{0.4mm}}ccccccccc}
    \toprule
    Timestep ($T$) & $1$ & $5$ & $10$ & $15$ & $20$ & $25$ & $30$ & $35$ & $40$ \\
    \midrule
    
    \begin{tabular}[c]{@{}c@{}}Vanilla \\ $D(\boldsymbol{z}_t)$\end{tabular} & 
    \includegraphics[width=.09\textwidth, valign=c]{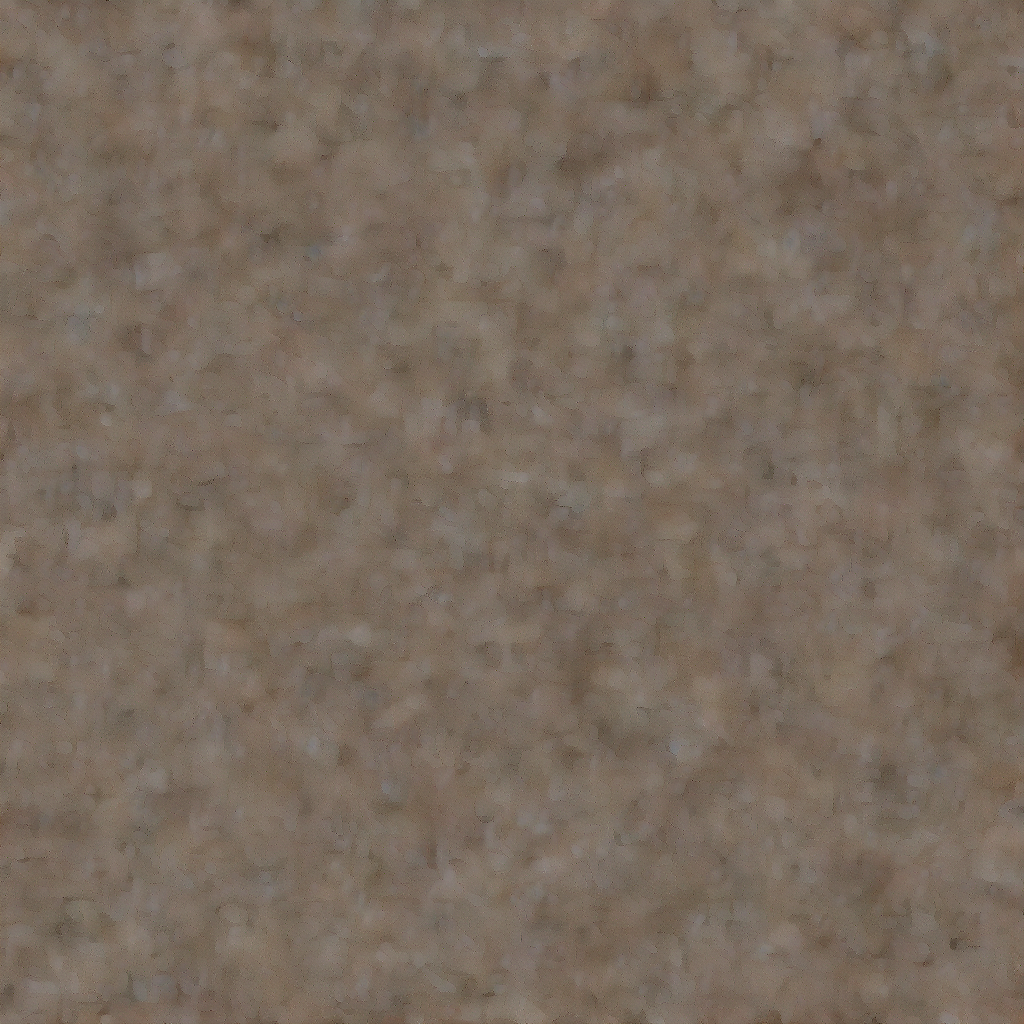} &
    \includegraphics[width=.09\textwidth, valign=c]{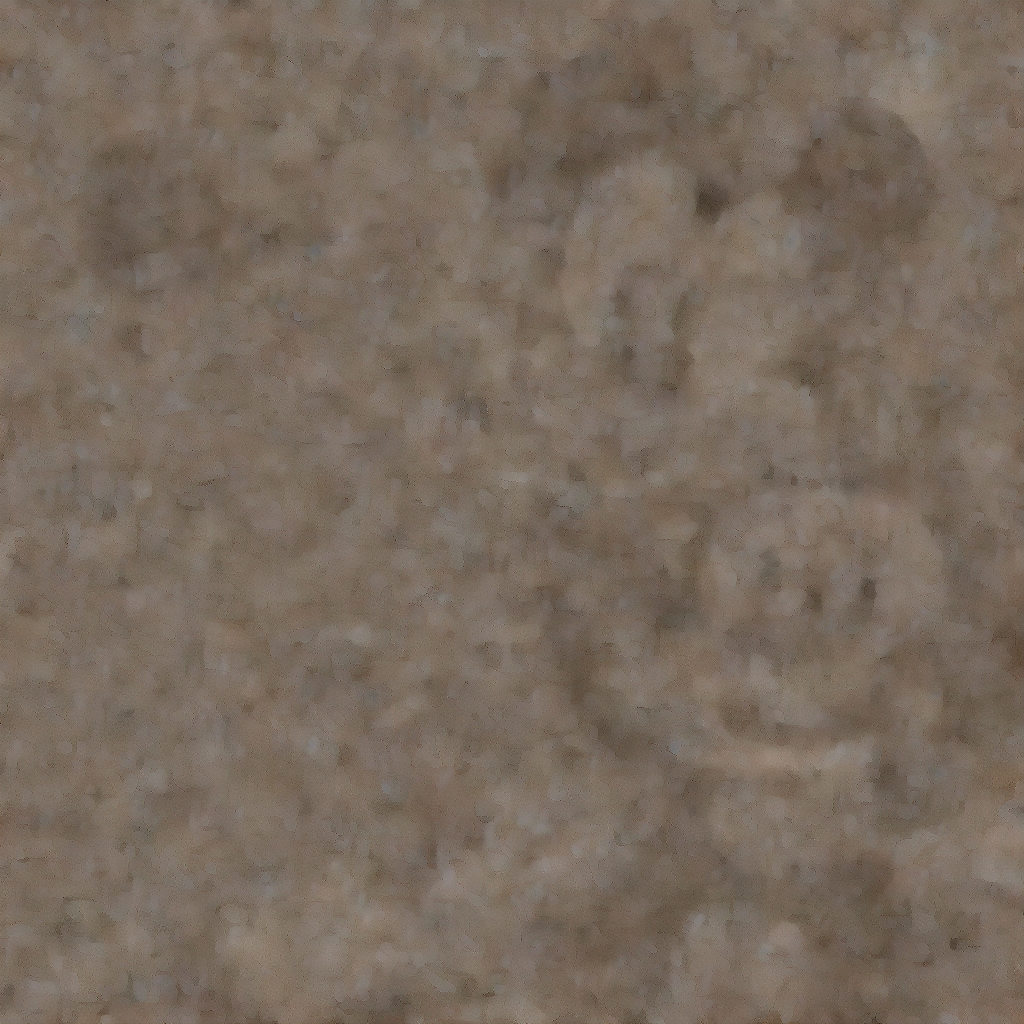} &
    \includegraphics[width=.09\textwidth, valign=c]{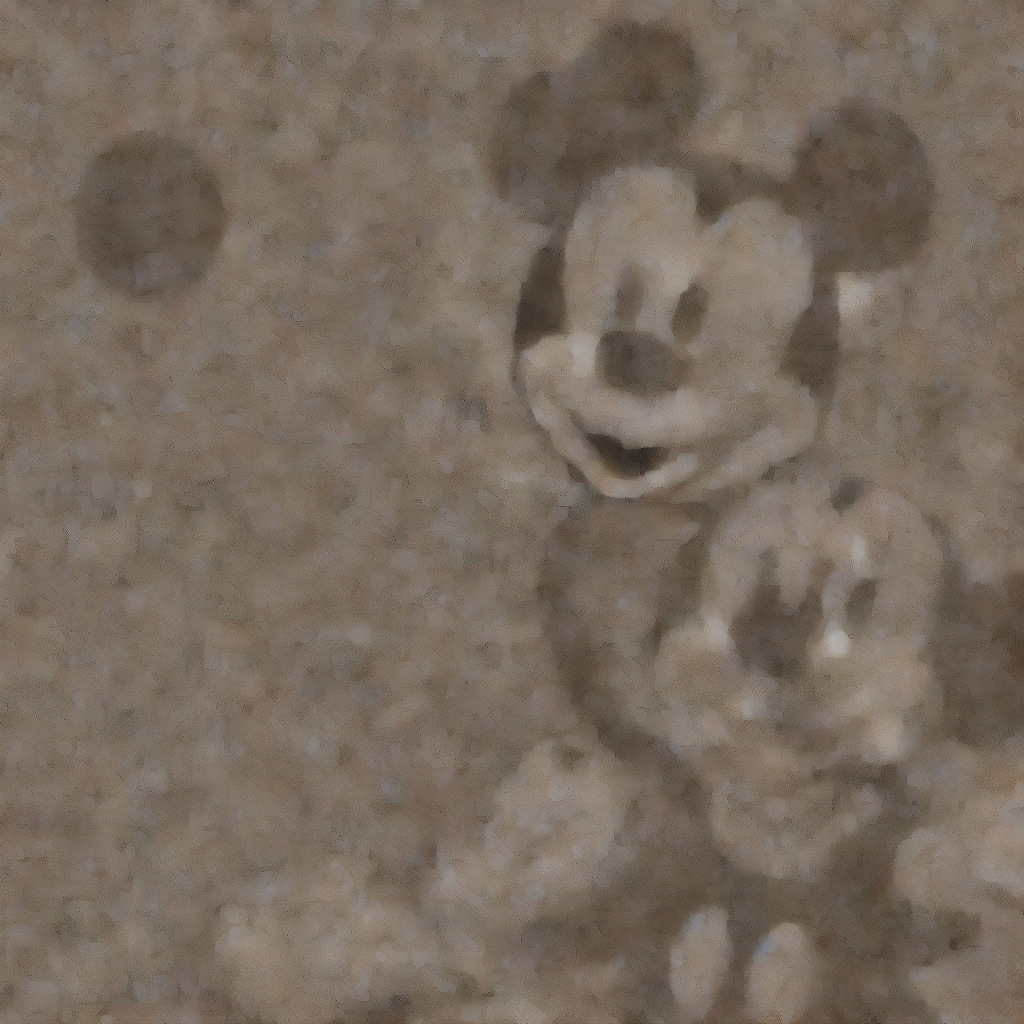} &
    \includegraphics[width=.09\textwidth, valign=c]{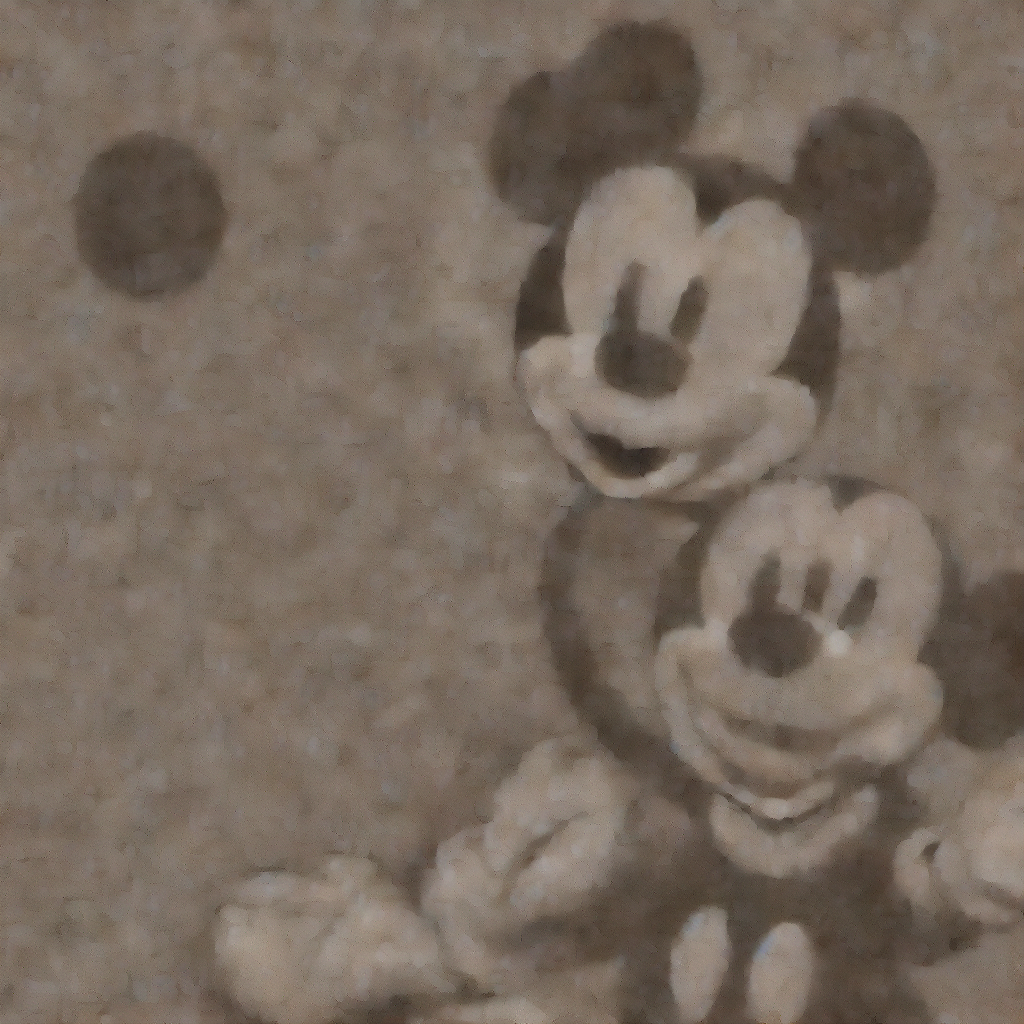} &
    \includegraphics[width=.09\textwidth, valign=c]{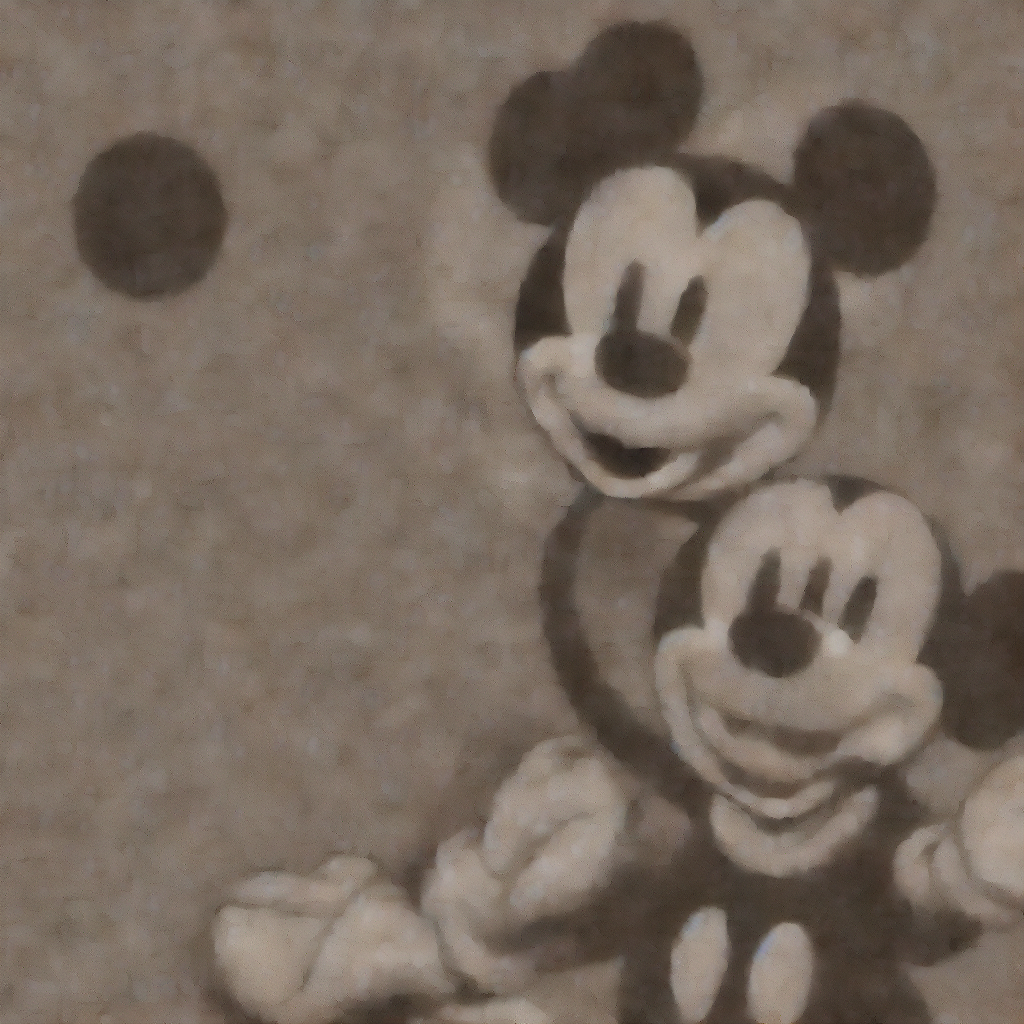} &
    \includegraphics[width=.09\textwidth, valign=c]{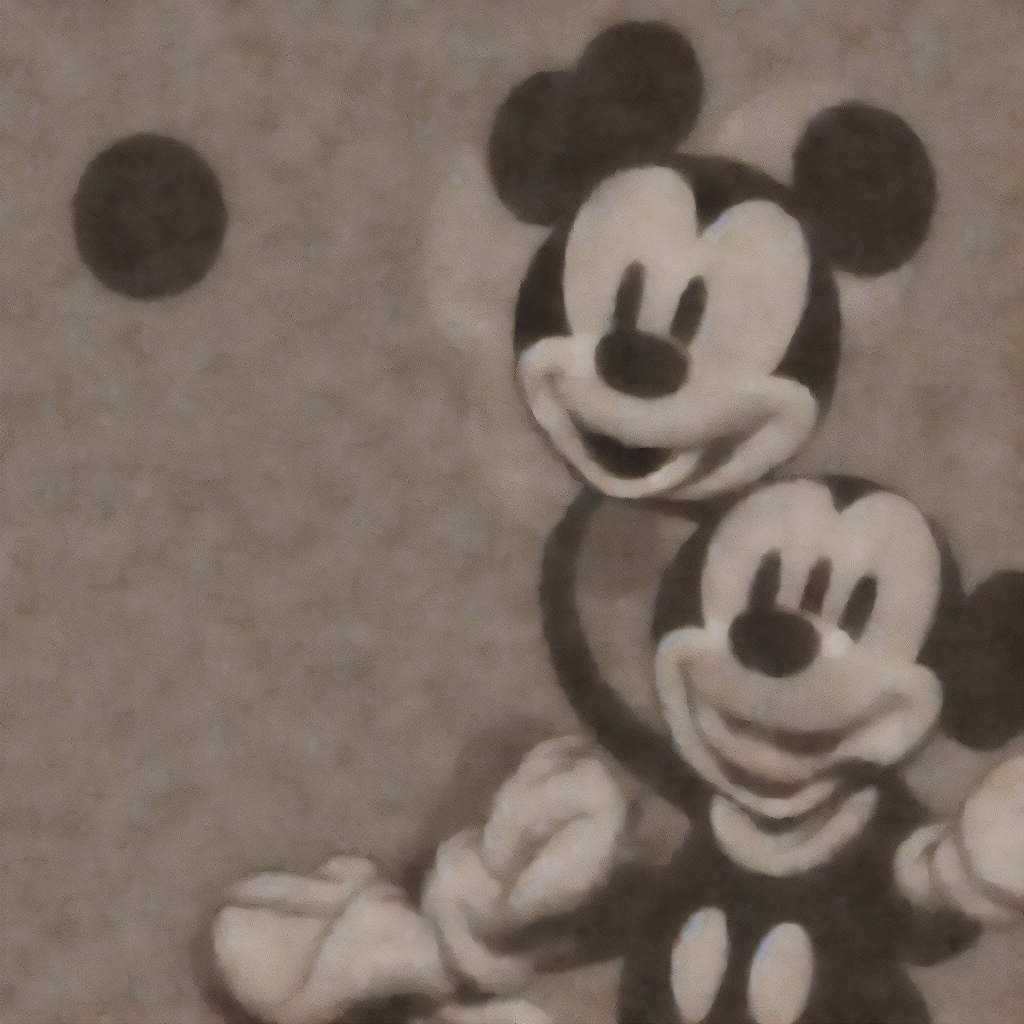} &
    \includegraphics[width=.09\textwidth, valign=c]{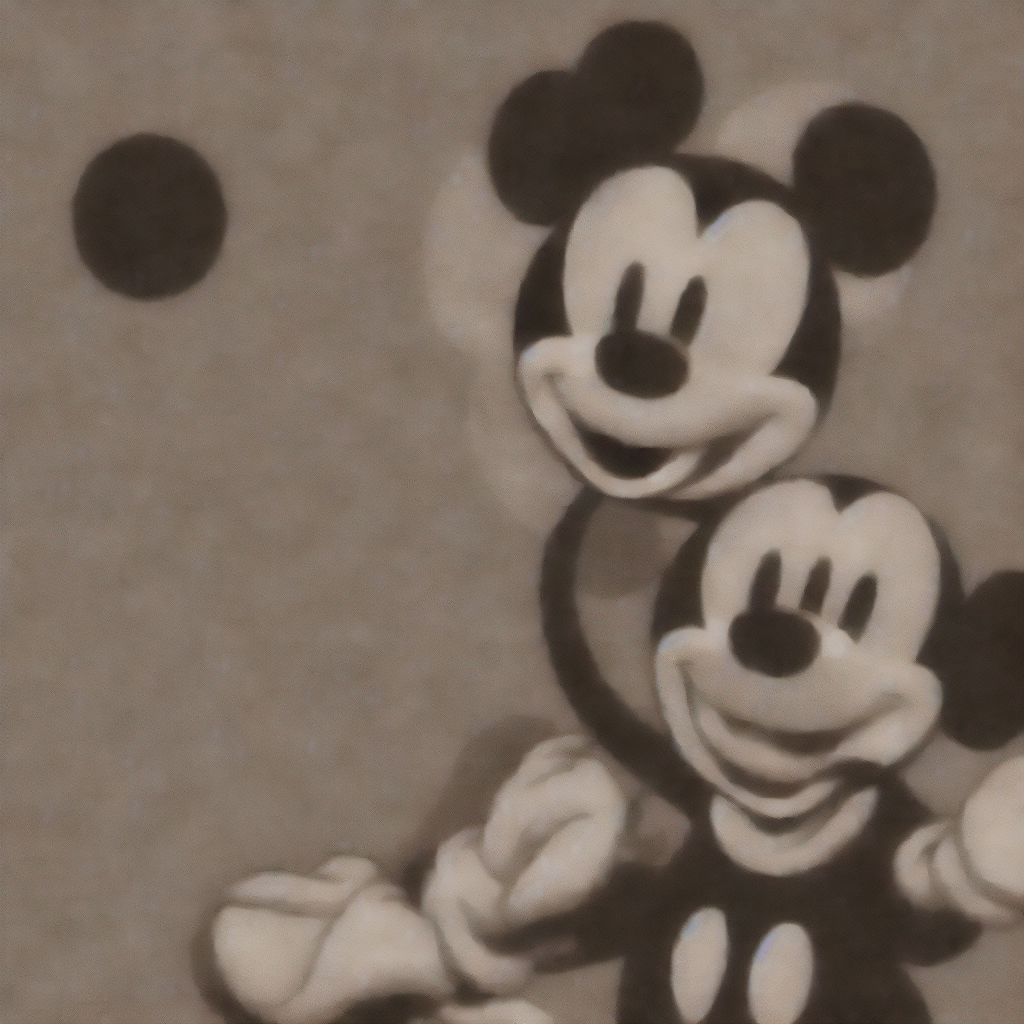} &
    \includegraphics[width=.09\textwidth, valign=c]{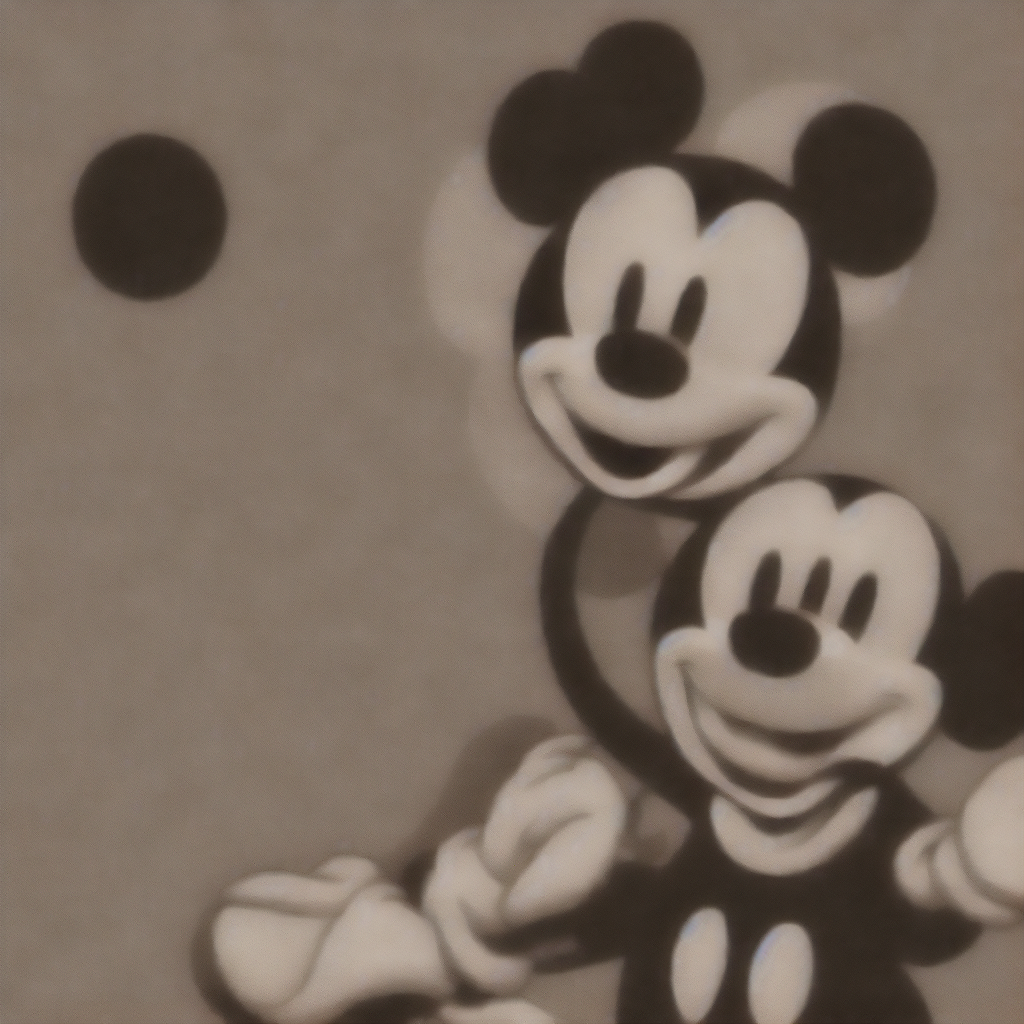} &
    \includegraphics[width=.09\textwidth, valign=c]{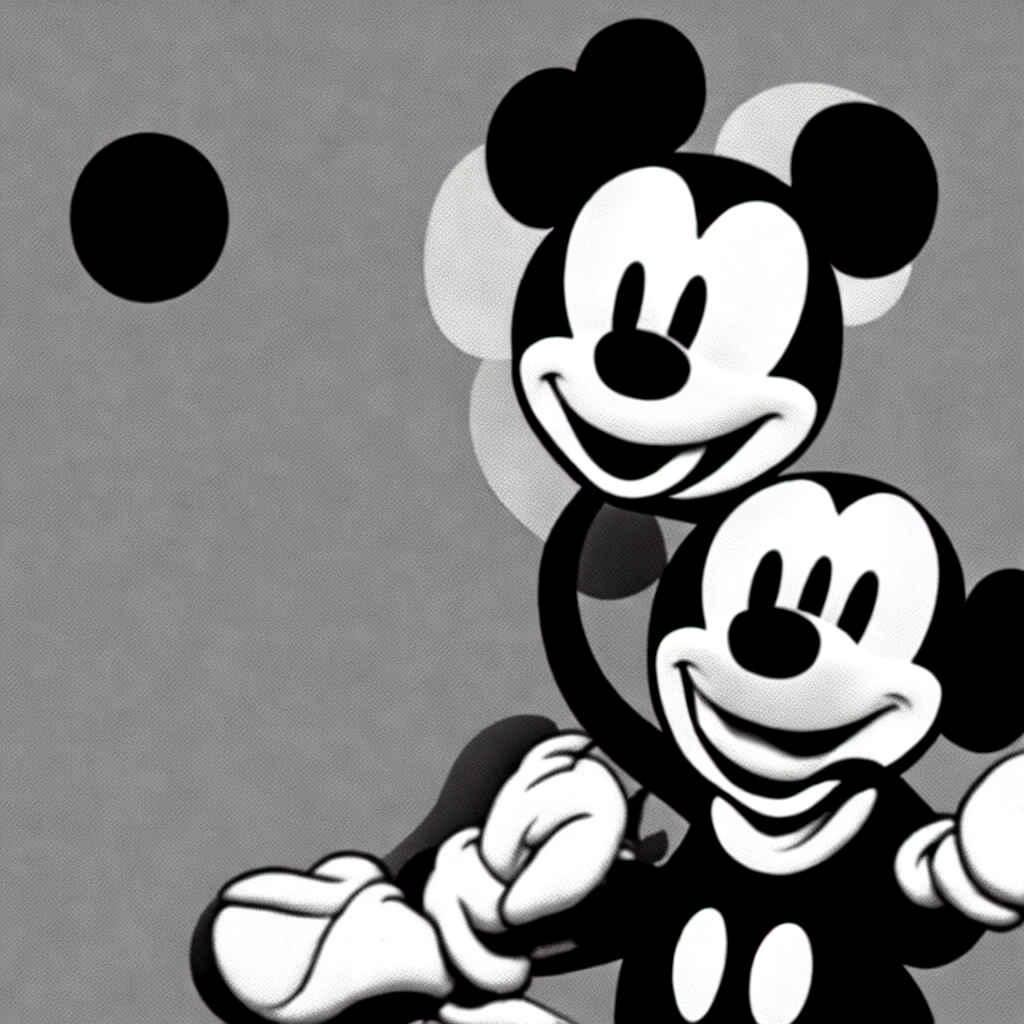}  \\
    \addlinespace[2mm] %
    
    \begin{tabular}[c]{@{}c@{}}$x$-pred \\ $D(\boldsymbol{x}_\theta(\boldsymbol{z}_t, t))$\end{tabular} & 
    \includegraphics[width=.09\textwidth, valign=c]{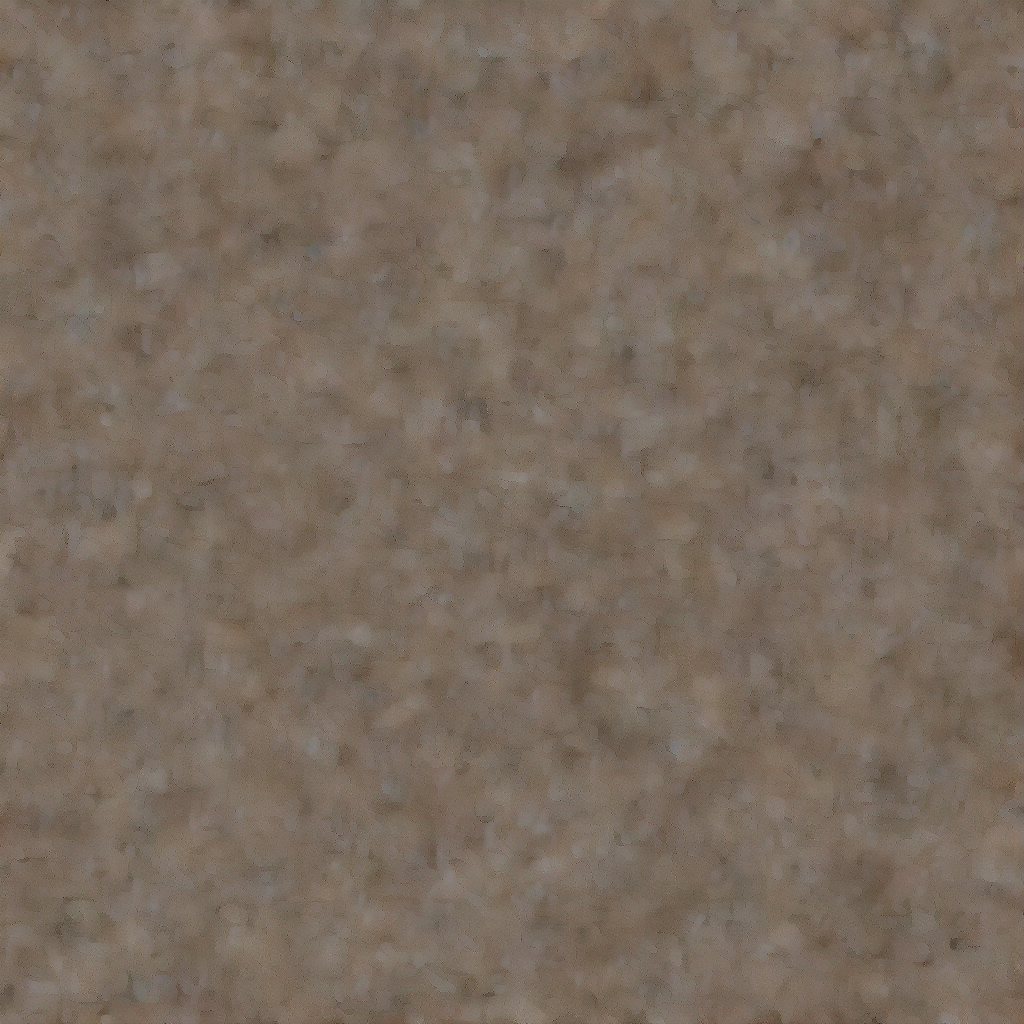} &
    \includegraphics[width=.09\textwidth, valign=c]{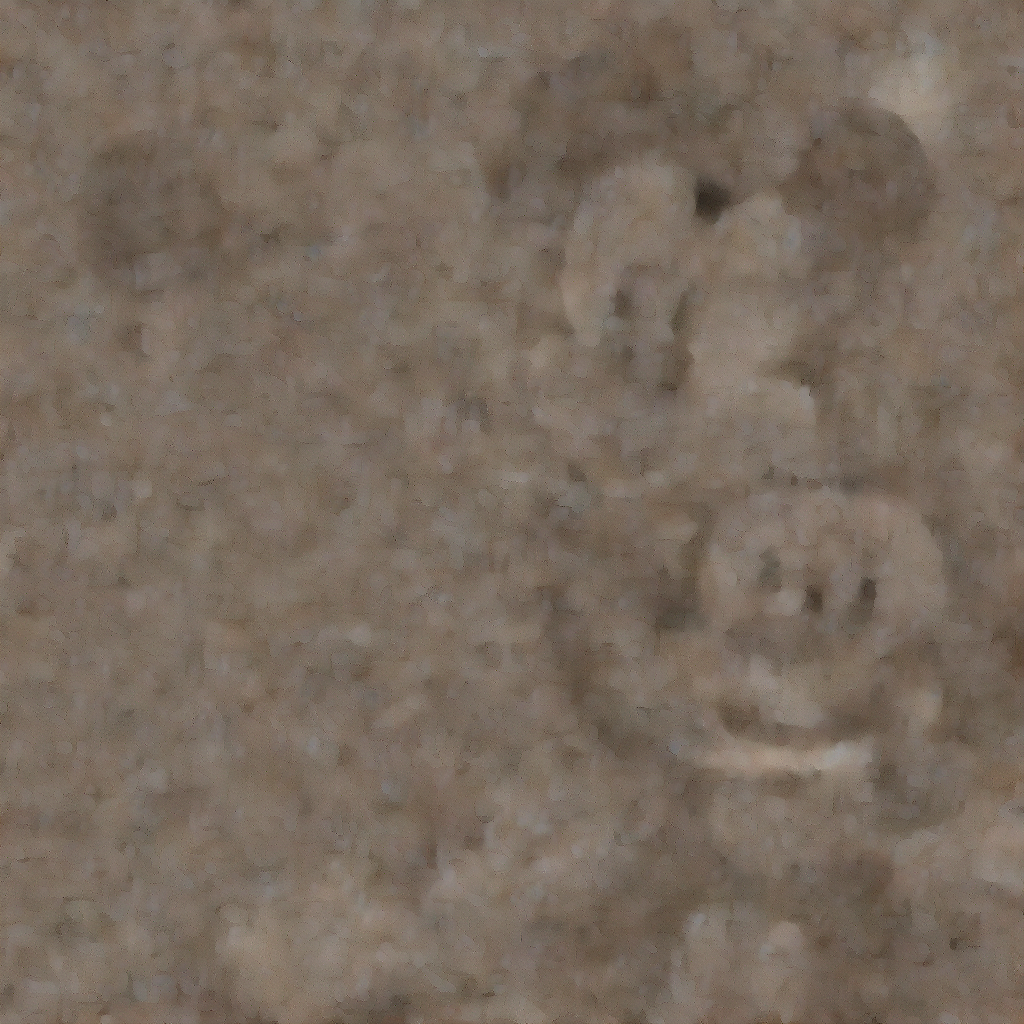} &
    \includegraphics[width=.09\textwidth, valign=c]{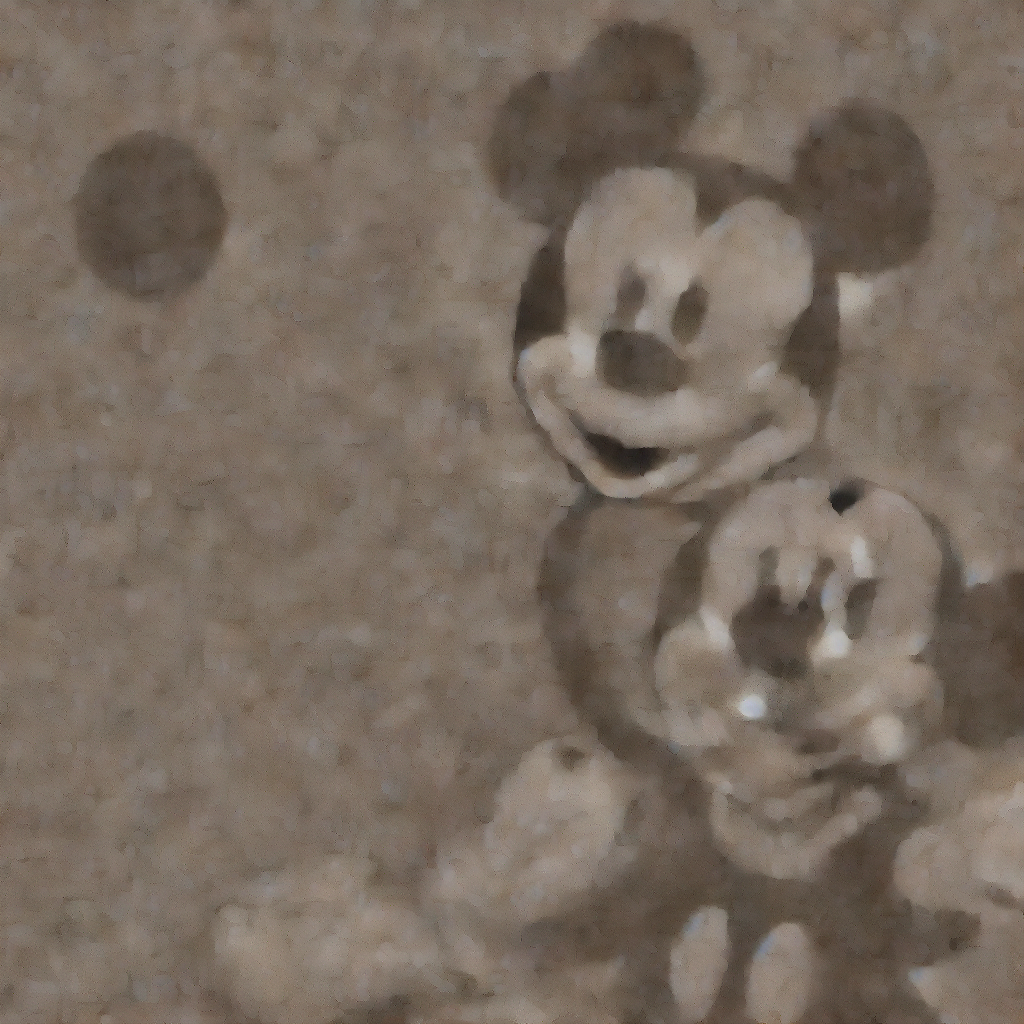} &
    \includegraphics[width=.09\textwidth, valign=c]{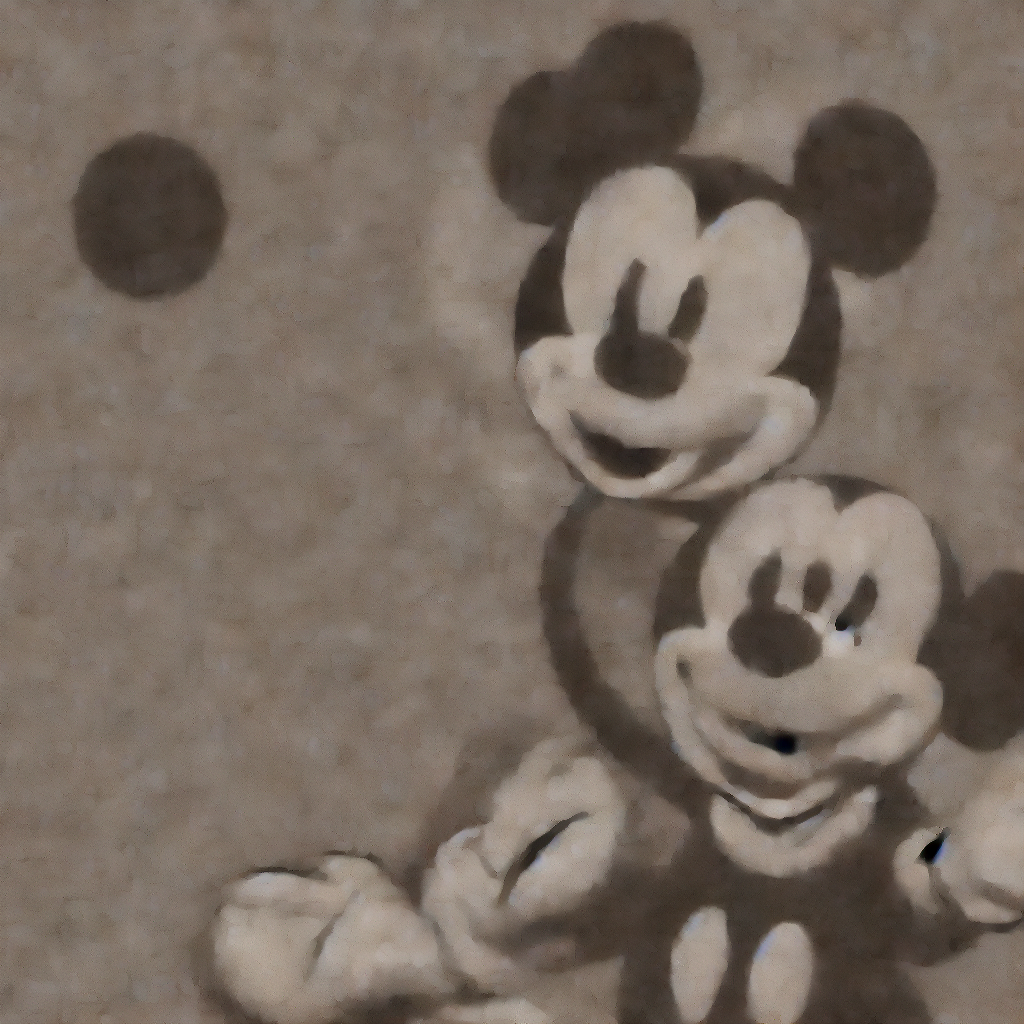} &
    \includegraphics[width=.09\textwidth, valign=c]{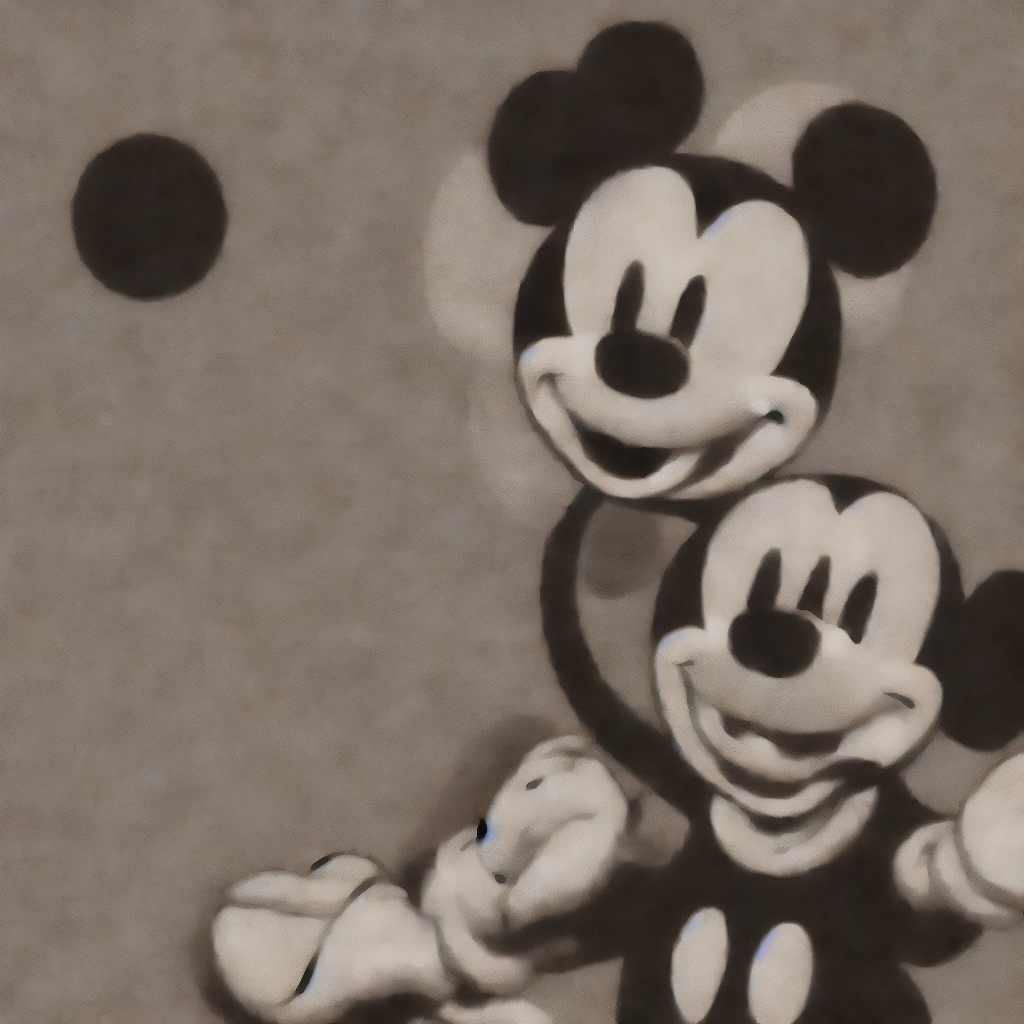} &
    \includegraphics[width=.09\textwidth, valign=c]{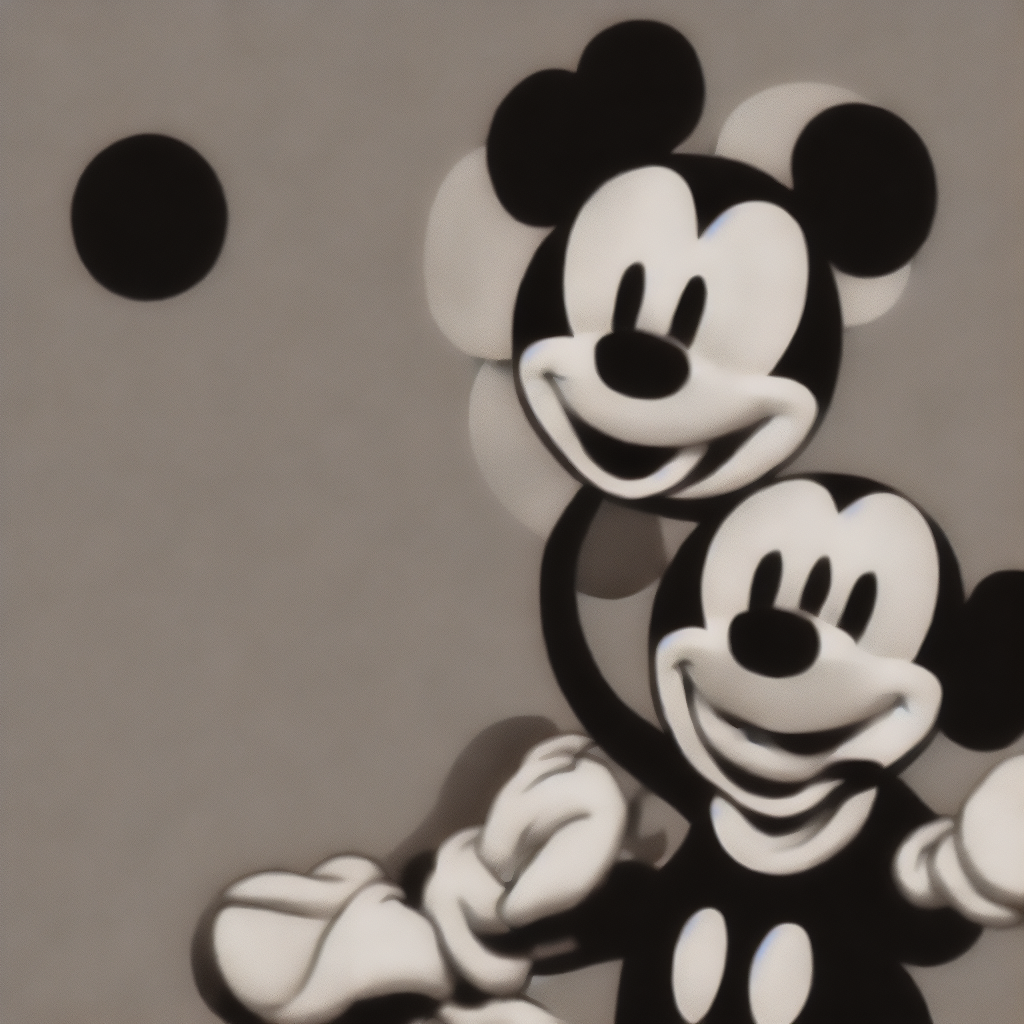} &
    \includegraphics[width=.09\textwidth, valign=c]{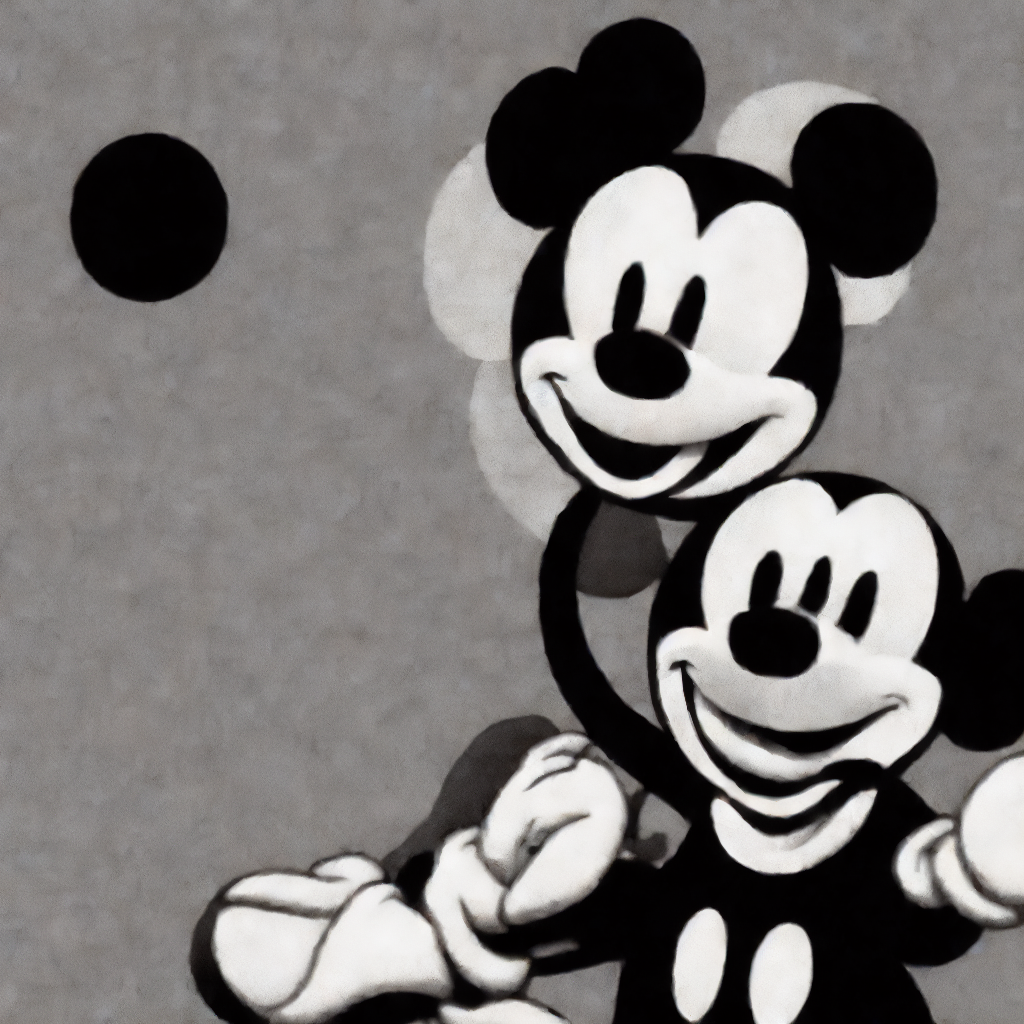} &
    \includegraphics[width=.09\textwidth, valign=c]{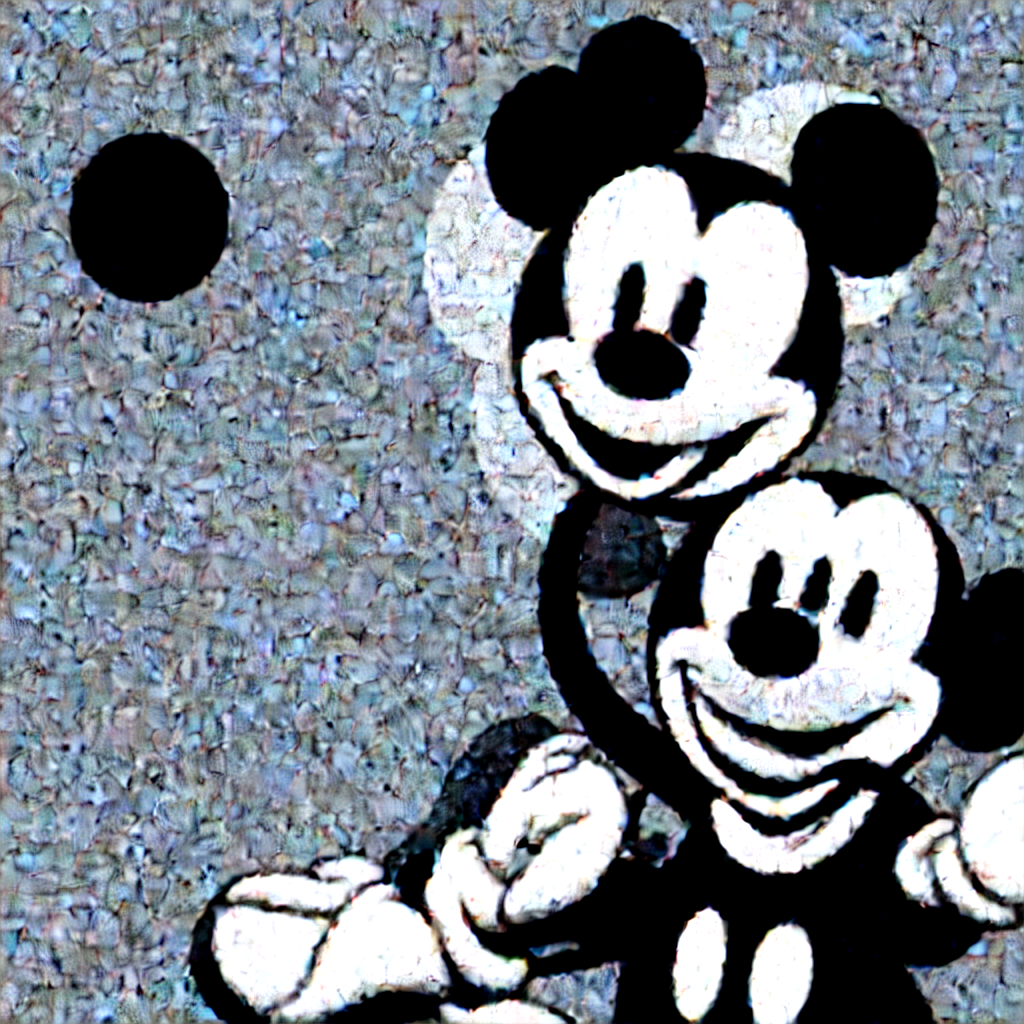} &
    \includegraphics[width=.09\textwidth, valign=c]{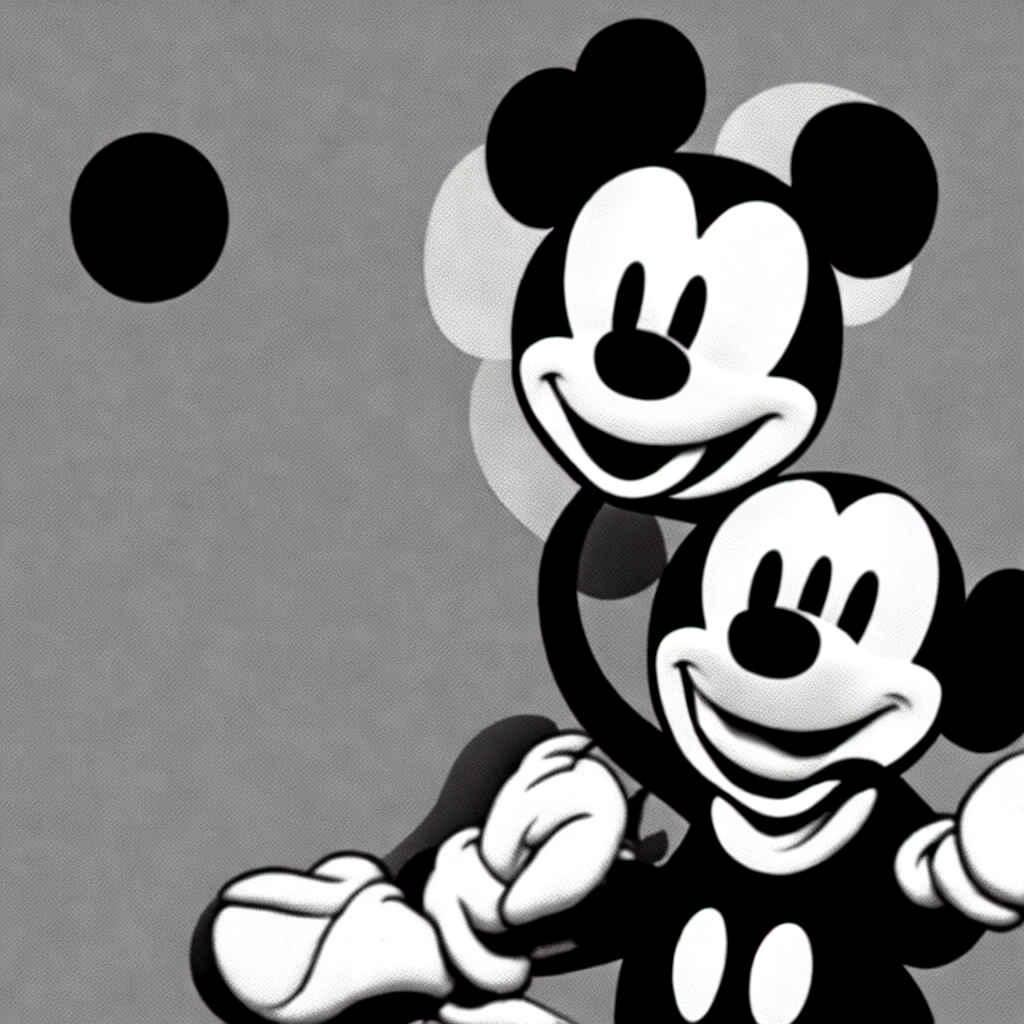}  \\
    \bottomrule
  \end{tabular}}
  \caption{Visualized intermediate latents of SD1.4, comparing states with and without the $x$-pred transformation across 40 generation timesteps.}
  \label{fig:sd_ablation}
\end{figure}

Fig.~\ref{fig:sd_ablation} shows the qualitative results of the ablation. 
The effectiveness of $x$-pred transformation appears from $T=20$ to $T=30$, which is the mid term as same as their result of Qwen-Image.

\subsection{Comparison with proprietary model}

\begin{table}[htbp]
\setlength{\tabcolsep}{2pt}
\caption{ROC-AUC and PR-AUC scores of violation binary classification for CPDM and HUB, and the time by the average score of single generation. 
The score of EDGE-Shield shown in this table is calculated at step 1 for Z-Image, at step 25 for Qwen-Image, respectively.
The experimental setup follows the setting of \textit{Effectiveness} in Sec.~5.1.
}\label{tbl:efficiency}
\begin{tabular}{@{}lllllll@{}}
\toprule
Method      & ROC-AUC & PR-AUC & Time (s) & ROC-AUC & PR-AUC & Time (s) \\ \midrule
gpt-4o-mini & 0.696   & 0.705  & 25.656   & 0.691   & 0.714  & 32.423   \\ \midrule
Ours w/ CLIP               & 0.827 & 0.846 & 0.404 & 0.830 & 0.844 & 11.621 \\
Ours w/ SigLIP             & 0.846 & 0.875 & \textbf{0.402} & 0.843 & 0.862 & 11.619 \\
Ours w/ SigLIP2            & 0.835 & 0.869 & 0.407 & 0.833 & 0.854 & \textbf{11.617} \\
Ours w/ Q3VLEmbed  & \textbf{0.857} & \textbf{0.898} & 0.454 & \textbf{0.844} & \textbf{0.883} & 12.107 \\ \bottomrule
\end{tabular}
\end{table}

We compare EDGE-Shield with gpt-4o-mini~\cite{openai2024gpt4ocard}.
Since the proprietary model cannot output the direct logits, we indicates the proprietary models to output the score by the prompt.
The other experimental setup follows the setting of \textit{Effectiveness} in Sec.~5.1 of the main draft.

\subsection{Qualitative comparison between EDGE-Shield and the other baselines}

\begin{table}[h]
\centering
\caption{Qualitative comparison on the HUB dataset. 
While \textbf{Reject} refers that content filter classifies the image as violative ones, Accept indicates that the generated image is classified as non-violative ones.
For EDGE-Shield, we use CLIP as an image encoder and use the threshold value as 0.7.
}\label{tab:qualitative}
\renewcommand{\arraystretch}{1.0}
\small
\begin{tabularx}{\textwidth}{p{3.5cm} *{2}{>{\centering\arraybackslash}X}} 
\toprule
\textbf{Reference} & \textbf{Lady Gaga} & \textbf{Emma Watson} \\ 
\midrule
\textbf{Generated Image}  
& \raisebox{-.5\height}{\includegraphics[width=0.5\linewidth]{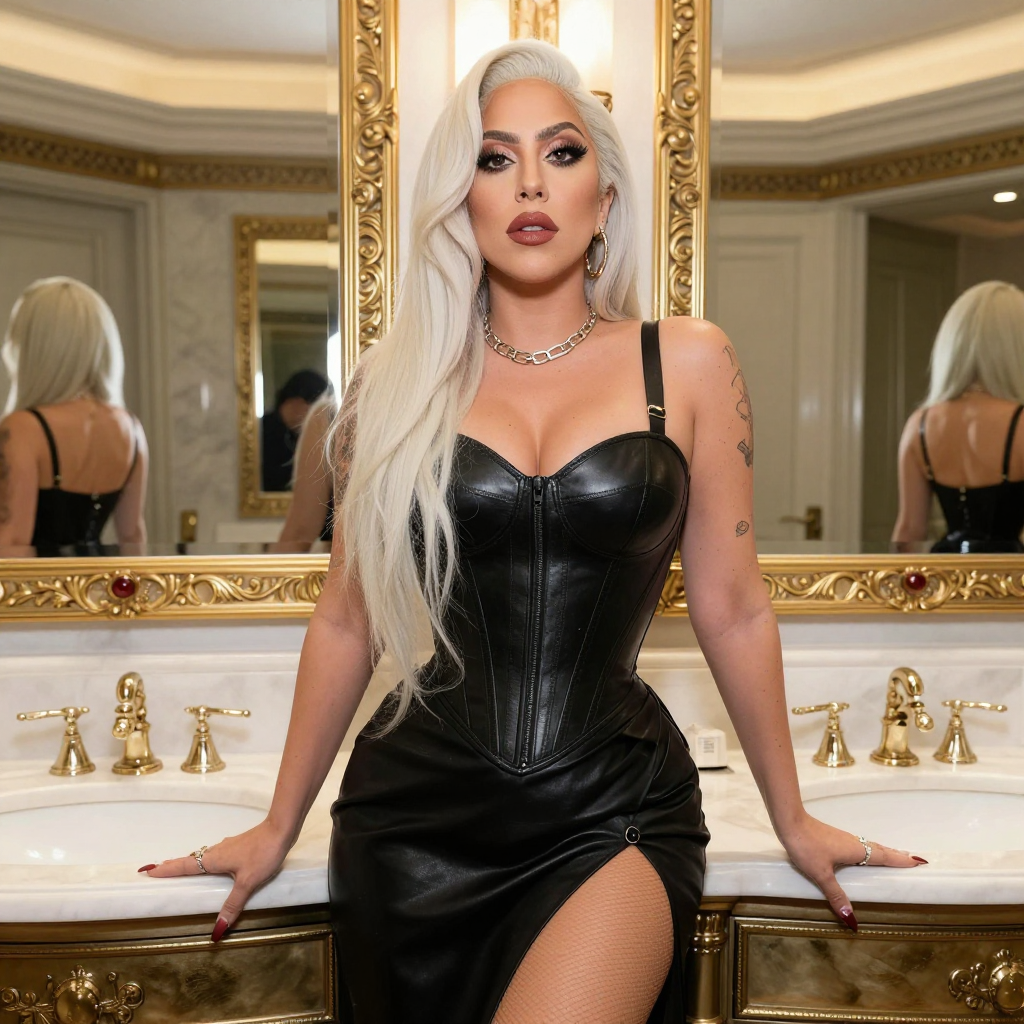}} 
& \raisebox{-.5\height}{\includegraphics[width=0.5\linewidth]{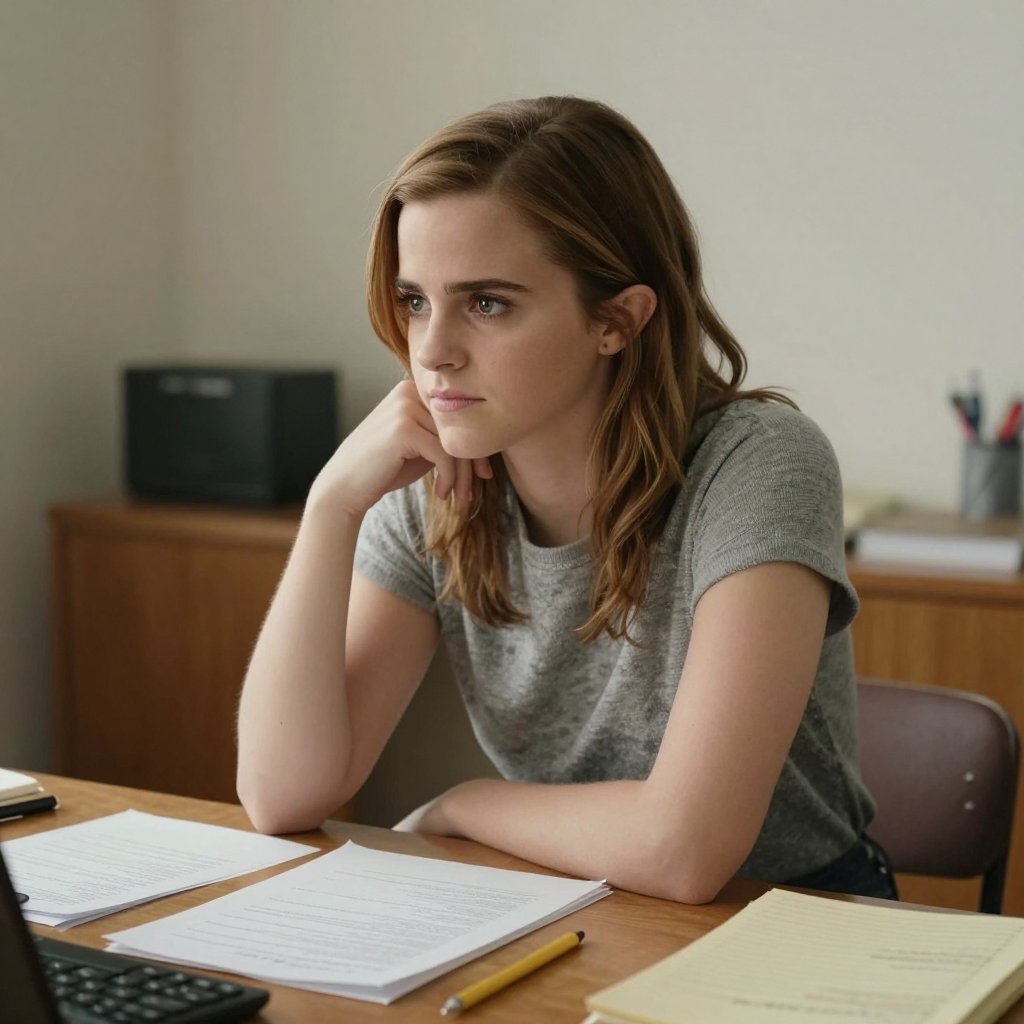}} \\ \midrule
\renewcommand{\arraystretch}{1.0} 
LLaVaGuard~\cite{helff2025llavaguard}       & \scriptsize{\textbf{Reject}} & \scriptsize{Accept}  \\
LLaVa-NEXT~\cite{liu2024llavanext}       & \scriptsize{\textbf{Reject}} & \scriptsize{\textbf{Reject}} \\
InternVL3.5~\cite{wang2025internvl35advancingopensourcemultimodal}      & \scriptsize{\textbf{Reject}} & \scriptsize{Accept}  \\
Qwen3-VL~\cite{bai2025qwen3vltechnicalreport}      & \scriptsize{\textbf{Reject}} & \scriptsize{\textbf{Reject}} \\ 
\midrule
\textbf{ours}    & \scriptsize{\textbf{Reject}} & \scriptsize{\textbf{Reject}} \\
\bottomrule
\end{tabularx}
\end{table}

Table~\ref{tab:qualitative} shows the qualitative comparison. 
Our approach correctly banned the content of the specified reference.

\section{Implementation details}
We will publish our code on acceptance.
We show the list of image encoders, VLMs, and T@I models we used in the experiment.
We utilize openai/clip-vit-base-patch32 for CLIP, google/siglip-base-patch16-224 for SigLIP, google/siglip2-base-patch16-224 for SigLIP2, and Qwen/Qwen3-VL-Embedding-2B for Qwen3VL-Embedding, all of them is on huggingface.
For VLMs including LLaVaGuard, LLaVa-NEXT, InternVL3.5, Qwen3-VL, we also use the models on huggingface.
For T2I models, we employ Z-Image-Turbo and Qwen-Image on huggingface.
All experiments are conducted by using 2 NVIDIA H100 GPUs.

\begin{table}[t]
\centering
\caption{Comparison of Straight (S) and Roundabout (R) prompting methods.}
\label{tbl:input-filter-comparison-only-prompt}
\small
\begin{tabularx}{\textwidth}{l|c|X}
\toprule
\textbf{Content Name} & \textbf{Style} & \textbf{Prompt} \\
\midrule
\multirow{2}{*}{Winnie the Pooh} & S & Winnie the Pooh, famous cartoon character \\
                                 & R & A lovable golden bear with a red shirt in cartoon style, known for his love of honey and his adventures in a forest. \\
\midrule
\multirow{2}{*}{Yosemite Sam}    & S & Yosemite Sam, cartoon character, cowboy, angry, Looney Tunes. \\
                                 & R & A short-tempered cowboy character with a large mustache in cartoon style, known for his explosive personality and wild west antics. \\
\midrule
\multirow{2}{*}{Foghorn Leghorn} & S & Foghorn Leghorn, cartoon character, rooster, Southern accent. \\
                                 & R & A large rooster with a booming Southern accent in cartoon style, known for his loud personality and comedic misadventures. \\
\midrule
\multirow{2}{*}{Pepe Le Pew}     & S & Pepe Le Pew, cartoon character, skunk, French accent, romantic. \\
                                 & R & A romantic skunk with a French accent in cartoon style, constantly pursuing a female cat despite her reluctance. \\
\bottomrule
\end{tabularx}
\end{table}

\end{document}